\pdfoutput=1

\documentclass[11pt]{article}

\usepackage[]{acl}
\usepackage{xspace}
\newcommand*{\yoruba}{Yor\`ub\'a\xspace}

\usepackage{times}

\usepackage{times}
\usepackage{latexsym}
\usepackage[utf8]{inputenc}
\usepackage{csquotes}

\usepackage[T4,T1]{fontenc}
\usepackage[verbose]{newunicodechar}

\usepackage{amssymb}
\usepackage{amsmath}

\usepackage{microtype}
\usepackage{booktabs}
\usepackage{multicol}
\usepackage{multirow}
\usepackage{cleveref}
\usepackage{caption}
\usepackage{subcaption}
\usepackage{longtable}

\usepackage{pgfplots}
\usetikzlibrary{matrix}
\usepgfplotslibrary{groupplots}
\pgfplotsset{compat=newest}
\usepackage{pgfplotstable}

\usepackage{listings}

\usepackage{pgf}
\usepackage{collcell}

\usepackage{inconsolata}

\usepackage{adjustbox}
\usepackage{babel}
\usetikzlibrary{babel, fit}

\usepackage{xurl}

\usepackage{newfloat}

\DeclareFloatingEnvironment[
    fileext=lop,
    listname={List of Prompt Templates},
    name={Prompt Template},
    placement=tbhp,
]{prompttemplate}

\DeclareCaptionSubType{prompttemplate}
\newcommand{\datasetname}{AfriScience-MT\xspace}
\newcommand{\datasetnamebf}{\textbf{AfriScience-MT}\xspace}

\DeclareFloatingEnvironment[
    fileext=lop,
    listname={List of Prompt Examples},
    name={Prompt Example},
    placement=tbhp,
]{promptexample}
\DeclareCaptionSubType{promptexample}
\usepackage{enumitem}
\setlist{itemsep=1pt, leftmargin=1.5em}

\usepackage[most,skins]{tcolorbox}
\usepackage{xcolor}
\usepackage{xspace}

\usepackage{pdflscape}
\usepackage{graphicx}
\usepackage{tikz}

\definecolor{metricsbg}{HTML}{E3F2FD}

\newtcbox{\evalbox}[2][]{%
  colback=#2,
  colframe=black!10,
  left=2mm, right=2mm, top=1mm, bottom=1mm,
  boxrule=0pt, arc=2mm, outer arc=2mm,
  #1
}

\def\footurl#1{\footnote{\url{#1}}}

\definecolor{spcNLLB}{HTML}{3B67C4}
\definecolor{spcGemma}{HTML}{2CA25F}
\definecolor{spcGPT}{HTML}{D24A4A}
\definecolor{spcFrame}{HTML}{888888}

\newtcolorbox{spLangGroup}{%
  enhanced, colback=white, colframe=spcFrame!50,
  boxrule=0.4pt, arc=2pt,
  left=2pt, right=2pt, top=2pt, bottom=2pt,
  before skip=2pt, after skip=2pt,
}

\newcommand{\spPanel}[4]{%
  \begin{minipage}[t]{0.48\linewidth}\centering
    \includegraphics[width=0.82\linewidth]%
                    {figures/spider_per_pair/#3/#1-#2.pdf}\\[-2pt]
    {\footnotesize (#4)~\texttt{#1$\to$#2}}
  \end{minipage}%
}

\newcommand{\spLangGrp}[5]{%
  \begin{minipage}[t]{0.49\linewidth}
    \begin{spLangGroup}
      \spPanel{eng}{#2}{#5}{#3}\hfill\spPanel{#2}{eng}{#5}{#4}
    \end{spLangGroup}
  \end{minipage}%
}

\newcommand{\spGridChrF}{%
  \spLangGrp{Amharic}{amh}{m}{n}{chrf}\hfill\spLangGrp{Hausa}{hau}{o}{p}{chrf}\par
  \vspace{2pt}%
  \spLangGrp{Luganda}{lug}{q}{r}{chrf}\hfill\spLangGrp{N.~Sotho}{nso}{s}{t}{chrf}\par
  \vspace{2pt}%
  \spLangGrp{Yor\`ub\'a}{yor}{u}{v}{chrf}\hfill\spLangGrp{isiZulu}{zul}{w}{x}{chrf}%
}
\newcommand{\spGridCOMET}{%
  \spLangGrp{Amharic}{amh}{a}{b}{comet}\hfill\spLangGrp{Hausa}{hau}{c}{d}{comet}\par
  \vspace{2pt}%
  \spLangGrp{Luganda}{lug}{e}{f}{comet}\hfill\spLangGrp{N.~Sotho}{nso}{g}{h}{comet}\par
  \vspace{2pt}%
  \spLangGrp{Yor\`ub\'a}{yor}{i}{j}{comet}\hfill\spLangGrp{isiZulu}{zul}{k}{l}{comet}%
}
\title{AfriScience-MT: Towards Decolonizing Science in Africa through Text Translation}

\author{
Idris Abdulmumin$^1$, Tajuddeen Gwadabe$^2$, Shamsuddeen Hassan Muhammad$^3$, \\
\bf David Ifeoluwa Adelani$^{4,5}$, Nomonde Khalo$^6$, Ibrahim Said Ahmad$^7$, Abiodun Modupe$^1$, \\
\bf Anina Mumm$^8$, Sibusiso Biyela$^9$, Michelle Rabie$^{10}$, Johanna Havemann$^{11}$, Marek Rei$^3$, \\
\bf Jade Abbott$^{12}$, Vukosi Marivate$^{1,12}$ \\[2mm]
\footnotesize $^1$Data Science for Social Impact, University of Pretoria, $^2$Masakhane Research Foundation, $^3$Imperial College London,\\
\footnotesize $^4$Mila, McGill University, $^5$Canada CIFAR AI Chair, $^6$University of Cape Town, $^7$University of Wisconsin - Stevens Point,\\
\footnotesize $^8$Independent Consultant, $^9$University of South Africa, $^{10}$Independent Researcher, $^{11}$Access 2 Perspectives, $^{12}$Lelapa AI\\[1mm]
\footnotesize\texttt{Contact: idris.abdulmumin@up.ac.za, tajgwadabe@gmail.com}
}

\begin{document}
\maketitle

\begin{abstract}

The dominance of colonial languages in African education and scientific communication limits how hundreds of millions of speakers of African languages access and produce scientific knowledge. A core obstacle is the lack of established scientific terminology in these languages. We introduce \textbf{\datasetname}, a parallel corpus covering six African languages (Amharic, Hausa, Luganda, Northern Sotho, Yorùbá, and isiZulu) across 11 scientific domains. Professional translators, working with expert science communicators, translated plain-language summaries of scientific papers into each target language and 
created new terms where none existed. We benchmark machine translation systems and large language models in zero-shot, few-shot, and fine-tuned settings. Our results show that closed-source models outperform all open-source models at both the sentence and document levels: GPT-5.4 and Gemini-3.1-Flash-Lite lead with average sentence-level COMET scores of 68.3 and 68.0, respectively, and tie at an average document-level COMET of 48.3. Among open systems, fine-tuned NLLB-1.3B reaches 67.3 at the sentence level, and TranslateGemma-12B reaches 44.0 at the document level with 1-shot in-context learning. We release \datasetname to support benchmarking and document-level scientific MT for African languages.
\end{abstract}

\section{Introduction}

\begin{displayquote}
    \textit{Without literacy in the languages of the masses, science and technology cannot be culturally-owned by Africans} -- \cite{Ngugi1986}
\end{displayquote}


In many African countries, colonial languages such as English, French, and Portuguese serve as the official languages of education and scientific communication \cite{Bamgbose2000}, excluding millions of indigenous-language speakers from accessing scientific knowledge. Translation has emerged as a decolonial practice to bridge this gap \cite{mazrui1998power}, with documented success in fields such as medicine and agriculture \cite{Bamgbose2000,alabi2025afridocmtdocumentlevelmtcorpus}. However, the lack of standardized scientific terminology in African languages impedes accurate translation of scientific concepts. For example, \citet{nekoto-etal-2020-participatory} reported difficulties translating COVID-19 surveys due to the absence of established scientific terminology in African languages.

To bridge this gap, we introduce \datasetname, a parallel corpus developed with expert science communicators and professional translators that aims to decolonize science in Africa by translating scientific articles into six African languages from West (Hausa, \yoruba), East (Amharic, Luganda), and Southern (Northern Sotho, isiZulu) Africa. The science communicators first develop summaries of 230 open-access papers,
and the professional translators then translate these summaries while co-developing scientific terminology in these languages.

We describe the \datasetname dataset and the terminology creation process and benchmark translation quality across several settings, including sequence-to-sequence and decoder-only LLMs (open- and closed-source) under zero-shot, in-context learning, and fine-tuning. We find that a fine-tuned NLLB, an encoder-decoder model, lands within 1 COMET of GPT-5.4 and Gemini-3.1-Flash-Lite at the sentence level and surpasses every fine-tuned open-weight LLM. For document-level translation, the newer proprietary models lead, and the best open-weight LLMs overtake the older GPT-4o and Gemini-1.5-Flash models. We also find that domain-specific data is essential and that artificial domain balancing hurts more than it helps. \textbf{We release \datasetname}\footnote{Link redacted for anonymity.} and the co-developed bilingual scientific glossaries, which we believe will help accelerate the decolonization of science in African languages.

\section{Related Works}
\paragraph{African datasets for machine translation:} 
Building parallel corpora for African languages has long been a central challenge in low-resource MT, with early efforts relying on web mining and automatic alignment, e.g., ParaCrawl \cite{banon-etal-2020-paracrawl} and CCAligned \cite{el-kishky-etal-2020-ccaligned}. These resources advanced African MT, though noisy alignments and other quality issues \cite{kreutzer-etal-2022-quality}.  \citet{abdulmumin-etal-2024-correcting} identified inconsistencies and inaccuracies in the FLORES evaluation benchmark that can compromise the integrity of downstream evaluation.

Recent efforts emphasize manual translation for higher quality. \citet{adelani-etal-2022-thousand} introduced MAFAND-MT, a manually curated parallel corpus covering 21 African languages in the news domain. \citet{elmadany-etal-2024-toucan} introduced Toucan with the AfroLingu-MT benchmark covering 43 African languages across 156 language pairs, enabling many-to-many translation between African languages. Similarly, \citet{alabi2025afridocmtdocumentlevelmtcorpus} developed AfriDoc-MT, a domain specific document-level corpus covering five African languages (Amharic, Hausa, Swahili, Yorùbá, and Zulu) in health and IT news domains with over 10,000 parallel sentences per language pair. Community efforts like Masakhane \cite{nekoto-etal-2020-participatory} and shared tasks at WMT \cite{adelani-etal-2022-findings} have further advanced the field. Large-scale multilingual models have also improved coverage, with NLLB-200 \cite{NLLB_Team2024-sm} supporting 55 African languages.

\paragraph{Scientific Translation in African Languages:}

Most existing MT resources for African languages focus on general-purpose or religious texts \cite{akhbardeh-etal-2021-findings,kreutzer-etal-2022-quality}. To date, no publicly available scientific translation corpora for African languages that cover fundamental STEM disciplines. This gap has tangible consequences: during COVID-19, rural healthcare workers struggled with untranslated medical guidelines \cite{nekoto-etal-2020-participatory}, and STEM students face comprehension challenges when instruction relies solely on colonial languages \cite{prah2018challenge}. We address this by introducing the AfriScience-MT dataset, the first multilingual corpus for scientific translation across eleven fundamental scientific disciplines in African languages, including co-developed terminology for concepts lacking standardized translations.

\section{AfriScience-MT Dataset}

\subsection{Languages and Data Source}

The \datasetname dataset covers six African languages representing major regions and language families of sub-Saharan Africa: Hausa and Yoruba (West Africa, Niger-Congo), Luganda (East Africa, Niger-Congo), Amharic (East Africa, Afro-Asiatic), and Zulu and Northern Sotho (Southern Africa, Niger-Congo). These languages are spoken by over 200 million native speakers across 15 countries ( \citeauthor{eberhard2025ethnologue}, \citeyear{eberhard2025ethnologue}). 

We selected 230 open-access English-language papers from three platforms: AfricArXiv, Zenodo, and Lens.org,\footnote{AfricArxiv (https://africarxiv.ubuntunet.net); Lens.org (https://www.lens.org); Zenodo (https://zenodo.org)} spanning 11 scientific domains: Agriculture, Biochemistry, Biology, Chemistry, Computer Science, Engineering, Geography, Health, Indigenous Knowledge, Sociology, and Statistics. Two principles guided paper selection.

\paragraph{Topic relevance} We prioritized papers that introduce new scientific terminology and support undergraduate science education and communication, allocating 70\% of selections to core natural sciences (biology, physics, medicine, and chemistry) to address terminology gaps in basic STEM concepts.

\paragraph{Author demographics} We applied decolonial principles by prioritizing papers with African first authors, ensuring diversity of author nationalities across African regions, and tracking gender representation.

\begin{figure*}[t]
  \centering
  \includegraphics[width=\textwidth]{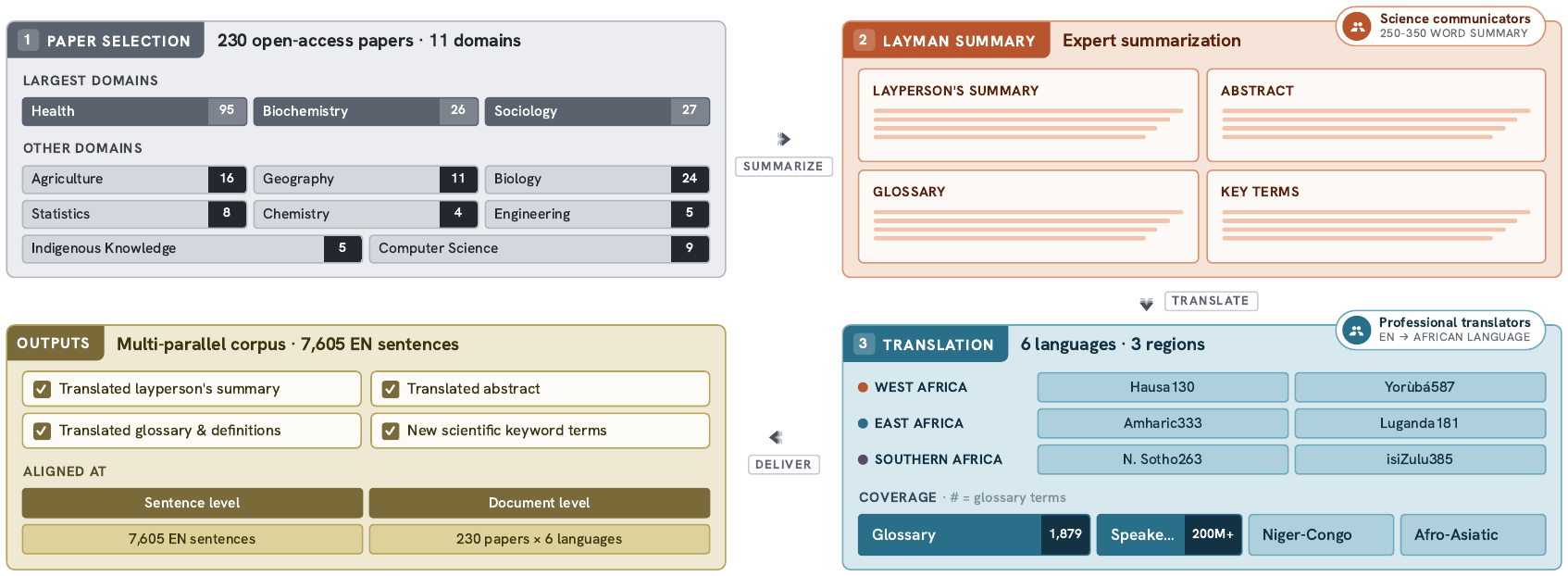}
  \caption{\textbf{Pipeline for constructing the AfriScience MT corpus}: (1) paper selection across 11 domains, (2) layperson summarization by science communicators, and (3) translation into 6 African languages by professional translators, producing a multi-parallel corpus of 7,605 aligned English sentences.}
  \label{fig:data-creation-framework}
\end{figure*}


\subsection{Scientific Translation Methodology}

We initially proposed full-text translation of scientific papers and conducted a collaborative workshop to train professional translators in processing technical academic content. However, post-training evaluation revealed that scientific translation required significantly more time than general-domain translation, with professional translators reporting up to a $10\times$ increase in processing time. To address this, we adopted a two-stage process: expert summarization followed by professional translation (\Cref{fig:data-creation-framework}).

\paragraph{Expert summarization} A domain-expert science communicator first produced a lay summary (250--350 words) of each paper, preserving the key contributions while simplifying technical content. Each summary was accompanied by a glossary of scientific terms; the per-language distribution is shown in \Cref{tab:data_splits_stats}.
\paragraph{Professional translation} A professional translator then translated the lay summary, abstract, terminology, and definitions from English into the target African language. Finally, we aligned all translations across languages to ensure multi-parallel corpus quality.

\subsection{Quality Control}

We implemented multiple quality control measures across all stages of dataset creation. We selected peer-reviewed, open-access scientific publications as source texts and validated each lay summary 
for completeness and clarity. For each target language, the lead translator reviewed all translations to verify accuracy and maintain terminological consistency. Finally, we verified sentence- and document-level alignment between each translation and the source.

\subsection{Dataset Statistics}
\label{subsec:data_stats}

The \datasetname corpus comprises 230 scientific papers and 7,605 English sentences, each translated into the six African target languages (\Cref{tab:data_splits_stats}). The 11 scientific domains are not balanced: Health dominates with 2,605 sentences in the training split alone (cf. \Cref{tab:balance_domain_eng_xxx_dual}), followed by Sociology, Biochemistry, and Biology (each 600--700 sentences). Further, token counts vary substantially across target languages, reflecting morphological and orthographic differences: Northern Sotho is the most token-rich at 256,464 tokens, Amharic and isiZulu are the most compact at $\approx$139k, and the English source has 179,517 tokens.
\Cref{tab:data_splits_stats} reports the bilingual scientific glossaries co-developed during translation and range from just over 100 to about 600 entries per language. The largest glossary is for \yoruba.

\begin{table}[t!]
    \centering
    \resizebox{\columnwidth}{!}{
        \begin{tabular}{llrrrr}
        \toprule
           & & & train & dev & test \\
        \midrule
        \multicolumn{3}{l}{\# papers} & 177 & 25 & 28 \\
        \midrule
        \midrule
        \multicolumn{6}{l}{\# sents [docs] per domain} \\
        \midrule
            & \multicolumn{2}{l}{Agriculture} & 388 [12] & 78 [2] & 67 [2] \\
            & \multicolumn{2}{l}{Biochemistry} & 651 [20] & 96 [3] & 112 [3] \\
            & \multicolumn{2}{l}{Biology} & 632 [19] & 74 [2] & 101 [3] \\
            & \multicolumn{2}{l}{Chemistry} & 46 [2] & 29 [1] & 42 [1] \\
            & \multicolumn{2}{l}{Computer Science} & 213 [7] & 33 [1] & 27 [1] \\
            & \multicolumn{2}{l}{Engineering} & 93 [3] & 20 [1] & 37 [1] \\
            & \multicolumn{2}{l}{Geography} & 218 [8] & 29 [1] & 85 [2] \\
            & \multicolumn{2}{l}{Health} & 2,605 [76] & 313 [9] & 324 [10] \\
            & \multicolumn{2}{l}{Indigenous Knowledge} & 92 [3] & 42 [1] & 49 [1] \\
            & \multicolumn{2}{l}{Sociology} & 670 [21] & 91 [3] & 103 [3] \\
            & \multicolumn{2}{l}{Statistics} & 184 [6] & 38 [1] & 23 [1] \\
        \midrule
            & \multicolumn{2}{l}{\textbf{total}} & 5,792 & 843 & 970 \\
        \midrule
        \midrule
        \multicolumn{6}{l}{tokens stats} \\
        \midrule
     \multicolumn{2}{l}{lang} & // gloss. & train & dev & test \\
        \midrule
        &     Amharic & 333 & 105,993 & 15,061 & 18,126 \\
        &     English & -- & 136,638 & 19,347 & 23,532 \\
        &     Hausa & 130 & 168,352 & 23,717 & 28,549 \\
        &     Luganda & 181 & 155,303 & 22,098 & 26,247 \\
        &     Northern Sotho & 263 & 195,665 & 27,405 & 33,394 \\
        &     Yoruba & 587 & 173,352 & 24,287 & 30,103 \\
        &     isiZulu & 385 & 106,476 & 15,087 & 17,947 \\
        \bottomrule
        \end{tabular}
    }
    \caption{\datasetname data statistics and splits for machine translation experiments}
    \label{tab:data_splits_stats}
\end{table}


\section{Experimental Setup}

To establish baselines on \datasetname, we run machine translation experiments across all six languages. Our experiments focus on English$\to$Target, reflecting the primary use case of translating scientific content from English into African languages. We also evaluate Target$\to$English to assess bidirectional capability.
We evaluate performance along four axes:

\begin{itemize}

    \item \textbf{Directionality:} English$\to$Target (primary) and Target$\to$English, plus a zero-shot African$\leftrightarrow$African (n-to-n) probe.

    \item \textbf{Models:} seq2seq and decoder-only LLMs (open- and closed-source) under zero-shot, in-context learning (ICL), and fine-tuning.

    \item \textbf{Training scope:} bilingual vs. multilingual.

    \item \textbf{Granularity:} sentence- vs. document-level translation.

\end{itemize}



\Cref{tab:data_splits_stats} shows the 80/10/10 dataset split for training, validation, and testing. Splits are made at the \emph{document} level within each domain, so that every split contains documents from every domain and no within-paper sentences leak across splits.

\paragraph{Evaluation Metrics} We report SSA-COMET \cite{li-etal-2025-ssa}, the African-language-tuned variant of COMET \cite{rei-etal-2020-comet}, as our primary metric and chrF \cite{popovic-2015-chrf} as a surface-level secondary metric; we refer to SSA-COMET simply as COMET. We surface chrF in the body only when it changes a ranking or exposes a degeneration signal (a large COMET--chrF gap indicates fluent but off-target generation; \Cref{tab:prompt_engineering_open}, \Cref{fig:multilingual_nllb}); full per-pair chrF for every table is in Appendix~\ref{app:chrf}, and the rationale for both using metrics is provided in Appendix~\ref{app:metric}. For document-level evaluation, we additionally use two LLM judges (GPT-5.5 and Gemini-3.1-Flash-Lite) for accuracy, fluency, and lexical/grammatical cohesion, following \citet{sun-llm-evaluator-2025} and \citet{alabi2025afridocmtdocumentlevelmtcorpus}; prompts are in Templates~\ref{prompt:fluency_eval}--\ref{prompt:cohesion_eval}.

\begin{table*}[t]
\centering
\small
\renewcommand{\arraystretch}{1.05}
\begin{tabular}{lrrrrrrrrrrrrr}
\toprule
\textbf{Model} & \multicolumn{6}{c}{\textbf{eng$\to$}} & \multicolumn{6}{c}{\textbf{$\to$eng}} & \\
\cmidrule(lr){2-7}\cmidrule(lr){8-13}
   & \texttt{amh} & \texttt{hau} & \texttt{lug} & \texttt{nso} & \texttt{yor} & \texttt{zul} & \texttt{amh} & \texttt{hau} & \texttt{lug} & \texttt{nso} & \texttt{yor} & \texttt{zul} & \textbf{Avg} \\
\midrule
\multicolumn{14}{l}{\textit{Zero-shot}} \\
\midrule
M2M100\textunderscore{}1.2B & 31.7 & 36.1 & 31.6 & 32.8 & 42.1 & 43.9 & 53.6 & 55.6 & 42.0 & 40.1 & 42.9 & 58.4 & 42.6 \\
M2M100\textunderscore{}418M & 23.5 & 27.9 & 32.2 & 30.6 & 29.1 & 33.0 & 46.8 & 47.0 & 37.9 & 39.6 & 37.7 & 51.3 & 36.4 \\
NLLB\textunderscore{}1.3B & \textbf{62.8} & \textbf{61.9} & \textbf{67.7} & \textbf{70.7} & \textbf{61.9} & \textbf{66.0} & \textbf{67.8} & \textbf{67.3} & \textbf{65.0} & \textbf{68.3} & \textbf{63.3} & \textbf{67.8} & \textbf{65.9} \\
NLLB\textunderscore{}600M & 62.2 & 60.2 & 65.7 & 70.2 & 59.3 & 65.5 & 67.2 & 66.6 & 63.8 & 67.2 & 62.0 & 67.2 & 64.8 \\
\midrule
AfriqueLlama-8B & 33.9 & 35.5 & \emph{42.2} & 38.2 & \emph{40.4} & 36.8 & 36.1 & 35.1 & 40.9 & 39.8 & 39.4 & 35.8 & 37.8 \\
AfriqueQwen-8B & 33.6 & 35.8 & 36.5 & 33.9 & 37.8 & 32.7 & 28.7 & 31.2 & 37.1 & 32.7 & 33.9 & 30.8 & 33.7 \\
Gemma2-9B & \emph{35.7} & \emph{36.2} & 40.3 & \emph{40.4} & 37.2 & \emph{36.8} & 41.1 & 39.5 & \emph{42.3} & \emph{42.7} & 39.9 & \emph{40.5} & \emph{39.4} \\
Llama3-8B & 21.6 & 24.6 & 35.9 & 34.0 & 29.2 & 27.1 & \emph{42.3} & 41.3 & 41.6 & 41.0 & 40.6 & 39.5 & 34.9 \\
Tiny-Aya-E & 27.1 & 28.1 & 37.9 & 36.6 & 37.9 & 33.3 & 39.1 & 37.4 & 41.0 & 38.0 & 41.8 & 38.6 & 36.4 \\
Tiny-Aya-G & 26.8 & 28.4 & 36.7 & 35.9 & 37.8 & 33.2 & 40.8 & 37.4 & 41.6 & 36.9 & 41.6 & 39.0 & 36.3 \\
TranslateGemma-12B & 29.5 & 33.7 & 37.5 & 37.8 & 37.9 & 33.6 & 40.7 & \emph{42.8} & 40.9 & 41.9 & \emph{45.6} & 30.8 & 37.7 \\
\midrule\midrule
\multicolumn{14}{l}{\textit{Fine-tuned}} \\
\midrule
M2M100\textunderscore{}1.2B & 59.7 & 59.8 & 66.0 & 68.0 & 64.5 & 61.9 & 60.0 & 63.8 & 61.7 & 65.0 & 59.6 & 61.1 & 62.6 \\
M2M100\textunderscore{}418M & 58.6 & 58.8 & 65.2 & 66.8 & 64.4 & 59.7 & 61.9 & 63.0 & 61.2 & 63.4 & 60.0 & 62.5 & 62.1 \\
NLLB\textunderscore{}1.3B & \textbf{66.0} & \textbf{63.3} & \textbf{70.0} & \textbf{70.7} & \textbf{67.3} & \textbf{66.0} & \textbf{68.6} & \textbf{68.0} & \textbf{65.9} & \textbf{69.0} & \textbf{64.8} & \textbf{68.2} & \textbf{67.3} \\
NLLB\textunderscore{}600M & 65.1 & 62.9 & 69.5 & 70.3 & 66.8 & 65.7 & 68.0 & 67.3 & 65.1 & 68.3 & 63.9 & 67.8 & 66.7 \\
\midrule
AfriqueLlama-8B & 49.6 & 51.5 & 58.7 & 58.7 & 55.5 & 51.6 & 60.9 & 53.5 & 54.7 & 58.7 & 55.7 & 56.4 & 55.5 \\
AfriqueQwen-8B & 23.8 & 37.2 & 45.1 & 52.1 & 21.2 & 28.7 & 36.6 & 28.2 & 22.8 & 25.8 & 41.3 & 26.5 & 32.4 \\
Gemma2-9B & 54.0 & 50.5 & 62.7 & 62.8 & 58.9 & 48.4 & 60.4 & 60.6 & 58.4 & 61.1 & 56.3 & 60.4 & 57.9 \\
Llama3-8B & 42.4 & 51.3 & 61.1 & 60.6 & 56.8 & 50.3 & 58.1 & 60.2 & 57.7 & 60.4 & 55.6 & 59.0 & 56.1 \\
Tiny-Aya-E & 54.3 & 54.3 & 59.8 & 59.2 & 57.1 & 55.7 & 57.8 & 61.0 & 56.5 & 58.7 & 50.2 & 60.5 & 57.1 \\
Tiny-Aya-G & \emph{61.9} & \emph{61.6} & \emph{65.7} & \emph{65.3} & \emph{65.2} & \emph{63.1} & \emph{67.3} & \emph{67.2} & \emph{64.3} & \emph{66.4} & \emph{63.8} & \emph{67.2} & \emph{64.9} \\
\bottomrule
\end{tabular}
\caption{\textbf{Sentence-level zero-shot vs. fine-tuned performance (COMET)}. The four seq2seq models (M2M100-418M, NLLB-600M, M2M100-1.2B, NLLB-1.3B) and six open-source LLMs appear in both major groups; LLMs in the \textit{Fine-tuned} group use per-pair LoRA $r{=}64$. \textbf{Bold}: best performance across \textit{zero-shot} or \textit{Fine-tuned}; \emph{italic}: best performance within seq2seq or LLM model group.}
\label{tab:main_zs_ft_comet}
\end{table*}

\paragraph{Fine-tuned baselines}
We fine-tune four seq2seq models: M2M100-418M and 1.2B \cite{fan-et-al-2021-beyond}, NLLB-600M and 1.3B \cite{nllbteam2022language}; and six open-weight LLMs: Llama3-8B \cite{grattafiori2024llama3herdmodels}, Gemma2-9B \cite{team2024gemma}, AfriqueLlama-8B and AfriqueQwen-8B \cite{yu2026afriquellmdatamixingmodel}, and the two Cohere Aya variants Tiny-Aya-Earth and Tiny-Aya-Global \cite{salamanca2026tinyayabridgingscale}. LLM fine-tuning uses QLoRA in 4-bit NF4 with ranks $r{\in}\{4, 64\}$. Additionally, we used $r{\in}\{8, 16, 32, 128\}$ for Gemma2-9B and Llama3-8B in the multilingual-vs-bilingual analysis (\Cref{tab:multilingual_vs_bilingual}). 
We additionally evaluate TranslateGemma-12B \cite{gemmatranslate2026} only in zero-shot and ICL. Unless a size is given, NLLB and M2M100 refer to the larger NLLB-1.3B and M2M100-1.2B.

\paragraph{Zero- and few-shot evaluation} We evaluate all eleven LLMs (seven open; four closed: GPT-4o \cite{openai2024gpt4ocard}, Gemini-1.5-Flash \cite{team2024gemini}, GPT-5.4 \cite{singh2026openaigpt5card}, Gemini-3.1-Flash-Lite \cite{gemini3}) in zero-shot and ICL. Unless a version is given, GPT and Gemini denote the newer GPT-5.4 and Gemini-3.1-Flash-Lite. At the sentence level, we use one shared configuration for all models -- template5, 10-shot, $T{=}0$, with randomly sampled exemplars; at the document level, where inputs are long, we cap exemplars at one and report zero-shot and 1-shot ICL at $T{=}0$. Configuration justification, the semantic-vs-random retrieval and temperature ablations, and decoding controls are provided in Appendix~\ref{app:translation_generation}.

\begin{table*}[t]
\centering
\small
\renewcommand{\arraystretch}{1.05}
\begin{tabular}{lrrrrrrrrrrrrr}
\toprule
\textbf{Model} & \multicolumn{6}{c}{\textbf{eng$\to$}} & \multicolumn{6}{c}{\textbf{$\to$eng}} & \\
\cmidrule(lr){2-7}\cmidrule(lr){8-13}
   & \texttt{amh} & \texttt{hau} & \texttt{lug} & \texttt{nso} & \texttt{yor} & \texttt{zul} & \texttt{amh} & \texttt{hau} & \texttt{lug} & \texttt{nso} & \texttt{yor} & \texttt{zul} & \textbf{Avg} \\
\midrule
\multicolumn{14}{l}{\textit{Zero-shot}} \\
\midrule
Gemini-1.5 & 57.9 & 55.8 & 59.4 & 62.2 & \textbf{57.5} & 53.9 & 62.6 & 61.4 & 58.1 & 59.8 & 54.8 & 60.8 & 58.7 \\
Gemini-3.1 & 55.6 & 55.1 & 58.7 & 53.0 & 53.0 & 52.2 & 58.7 & 61.0 & 58.5 & 60.2 & 57.9 & 60.8 & 57.0 \\
GPT-4o & 39.4 & 56.8 & 60.7 & 62.6 & 54.6 & 58.3 & 60.5 & 62.3 & 59.3 & 62.0 & 57.8 & 62.1 & 58.0 \\
GPT-5.4 & \textbf{61.7} & \textbf{58.4} & \textbf{64.3} & \textbf{66.7} & 55.9 & \textbf{63.2} & \textbf{69.1} & \textbf{69.0} & \textbf{67.1} & \textbf{69.5} & \textbf{66.2} & \textbf{68.8} & \textbf{65.0} \\
\midrule
Best open (see \Cref{tab:main_zs_ft_comet}) & \emph{35.7} & \emph{36.2} & \emph{42.2} & \emph{40.4} & \emph{40.4} & \emph{36.8} & \emph{42.3} & \emph{42.8} & \emph{42.3} & \emph{42.7} & \emph{45.6} & \emph{40.5} & \emph{40.6} \\
\midrule\midrule
\multicolumn{14}{l}{\textit{template5, 10-shot ICL}} \\
\midrule
Gemini-1.5 & 64.8 & 62.4 & 66.8 & 68.4 & 64.7 & 62.2 & 68.7 & 67.9 & 65.6 & 67.5 & 64.5 & 67.6 & 65.9 \\
Gemini-3.1 & \textbf{67.8} & \textbf{64.8} & 70.3 & 70.1 & 67.8 & 65.9 & 69.3 & 68.7 & 66.8 & 69.1 & 66.4 & 68.4 & 68.0 \\
GPT-4o & 52.7 & 62.2 & 68.0 & 68.3 & 65.4 & 64.6 & 67.2 & 68.4 & 66.3 & 68.4 & 65.6 & 68.3 & 65.4 \\
GPT-5.4 & 66.6 & 64.8 & \textbf{70.6} & \textbf{70.7} & \textbf{68.0} & \textbf{66.2} & \textbf{69.7} & \textbf{69.1} & \textbf{67.5} & \textbf{69.9} & \textbf{67.4} & \textbf{68.7} & \textbf{68.3} \\
\midrule
AfriqueLlama-8B & 50.1 & 49.0 & 48.7 & 49.1 & 49.8 & 48.5 & 36.6 & 50.4 & 53.6 & 52.4 & 47.1 & 50.4 & 48.8 \\
AfriqueQwen-8B & 43.8 & 42.9 & 41.2 & 44.0 & 47.8 & 43.5 & 42.8 & 41.3 & 43.0 & 43.5 & 44.4 & 41.0 & 43.3 \\
Gemma2-9B & 33.7 & 50.4 & 37.5 & 40.4 & 35.1 & 40.0 & 64.1 & \emph{66.3} & \emph{61.3} & \emph{63.0} & 60.4 & 65.1 & 51.4 \\
Llama3-8B & 25.9 & 36.1 & 44.2 & 40.2 & 38.5 & 32.8 & 56.5 & 63.4 & 56.9 & 54.8 & 57.2 & 56.9 & 46.9 \\
Tiny-Aya-E & 55.6 & \emph{56.7} & 36.8 & 31.7 & 61.4 & 59.0 & \emph{66.3} & 66.0 & 50.3 & 50.4 & \emph{63.4} & \emph{66.2} & \emph{55.3} \\
Tiny-Aya-G & \emph{55.8} & 56.1 & 35.9 & 30.6 & \emph{61.7} & \emph{59.2} & 66.2 & 66.0 & 50.6 & 50.2 & 63.3 & 66.1 & 55.1 \\
TranslateGemma-12B & 53.6 & 52.4 & \emph{50.1} & \emph{50.3} & 51.0 & 50.6 & 41.2 & 39.2 & 45.8 & 43.6 & 31.6 & 39.4 & 45.7 \\
\bottomrule
\end{tabular}
\caption{\textbf{Open vs. closed-source LLMs} at sentence-level in-context learning (COMET), all rows at the uniform \texttt{template5, 10-shot} configuration (see \Cref{tab:template_shot_ablation_comet} for the template/shot ablation that selected this config). \textit{Best open} in the zero-shot block is the per-cell maximum across the seven open LLMs in \Cref{tab:main_zs_ft_comet}. AfriqueLlama-8B, AfriqueQwen-8B, and TranslateGemma-12B degenerate at this prompt (visible as COMET~$\gg$~ChrF in their rows); see \Cref{tab:prompt_engineering_open} for their vanilla-prompt baseline. \textbf{Bold}: best performance across \textit{zero-shot} and \textit{ICL}; \emph{italic}: best performance within open or closed model group.}
\label{tab:open_vs_closed_icl_comet}
\end{table*}

\section{Results}
\label{sec:results}

\subsection{Sentence-Level Performance}
\label{subsec:sentence_results}

\paragraph{Effectiveness of supervised fine-tuning:}
\Cref{tab:main_zs_ft_comet} reports sentence-level COMET for both zero-shot and fine-tuned models across the 12 directional pairs. Fine-tuning proves beneficial across every model family. The strongest system overall is NLLB-1.3B fine-tuned on AfriScience-MT, averaging \textbf{67.3 COMET} across all 12 pairs and reaching 70.7 on English$\to$Northern Sotho. The smaller NLLB-600M follows at 66.7 avg -- only 0.6 COMET behind despite being less than half the size, a strong efficiency result for African MT. The two M2M100 baselines fall further behind (62.6 / 62.1), confirming that NLLB's massively multilingual pre-training does the heavy lifting. Among the seven open LLMs (per-pair LoRA $r{=}64$), Tiny-Aya-Global leads at 64.9 avg, but its lead is suspect: in Target$\to$English its COMET is unusually high while its chrF is well below the comparable open LLMs. This indicates a potentially fluent but off-target output that COMET rewards, whereas chrF does not (\Cref{tab:prompt_engineering_open}). We therefore treat Gemma2-9B (57.9 avg) as the better-behaved best open reference for the rest of the paper.
Appendix~\ref{app:lora_rank_and_scope} reports a LoRA rank ablation ($r{=}4$ vs $r{=}64$, six open LLMs) and a bilingual vs. multilingual training-scope ablation across NLLB, Llama, and Gemma families.

\paragraph{Directional asymmetry}
Across both seq2seq and LLMs, Target$\to$English consistently outperforms English$\to$Target, by 1 to 3 COMET on average for the strongest models and by much wider margins for weaker open LLMs. This reflects both the models' stronger English representations and the inherent difficulty of generating morphologically complex African target languages and is consistent across our zero-shot, ICL, and fine-tuned settings.

\subsection{Zero- and Few-Shot LLM Prompting}
\label{subsec:icl_results}

\Cref{tab:open_vs_closed_icl_comet} presents all eleven LLMs under our shared ICL configuration (template~5, 10-shot, $T{=}0$). GPT and Gemini (GPT-5.4 and Gemini-3.1-Flash-Lite) lead the sentence-level ICL average at 68.3 and 68.0, with GPT-4o at 65.4 and Gemini-1.5-Flash at 65.9. The remaining gap from several open LLMs to the leading closed models looks small in the COMET column but is partly misleading: three open LLMs, AfriqueLlama-8B, AfriqueQwen-8B, and TranslateGemma-12B, \textbf{degenerate} at this prompt, with COMET~$\gg$~chrF in the companion \Cref{tab:prompt_engineering_open}, also indicating fluent but off-target output that COMET tolerates and chrF penalizes. The two clean open models at this prompt, Llama3-8B and Gemma2-9B-IT, come within a few COMET of GPT-4o and Gemini-1.5-Flash and beat them on several Target$\to$English pairs. Crucially, every open LLM is dominated by fine-tuned NLLB-1.3B (\Cref{tab:main_zs_ft_comet}; per-domain breakdown in \Cref{fig:spider_per_pair}), so prompting is not a substitute for fine-tuning when domain-specific data is available.

For the open LLMs, almost all of the vanilla-to-templated gain is attributable to the prompt template rather than the shot count (\Cref{app:open_lift_decomposition}); for the three degenerate models above, the same template move drives a large negative $\Delta$ chrF that is invisible to COMET. On the closed models, neither retrieval strategy (semantic vs. random) nor decoding temperature ($T{=}0$ vs. $T{=}0.6$) matters much at the adopted configuration: prompt template choice dominates both (Appendix~\ref{app:closed_api_ablations}, \Cref{fig:gpt_gemini_eng_xxx,fig:gpt_gemini_xxx_eng}).

\subsection{Cross-Domain Data Transfer}
\label{subsec:cross_domain_results}

\Cref{tab:mafand_nllb} evaluates whether out-of-domain African-language data (MAFAND-MT, primarily news) can substitute for or augment AfriScience-MT for scientific translation, on the four MAFAND-supported pairs. Training NLLB-1.3B on AfriScience-MT alone outperforms training on MAFAND-MT alone by \textbf{5--10 COMET on average} in both directions, with a substantial gap of 7.1 COMET (62.8 vs. 55.7) on English$\to$Hausa. Combining the two corpora (\texttt{+MF}) does \emph{not} help and, in some pairs, slightly hurts the AfriScience-MT-only score: the in-domain signal is strong enough that mixing in news-domain data dilutes rather than augments it. The chrF in the table tells the same story slightly more strongly. This provides the clearest evidence in the paper that AfriScience-MT cannot be substituted for a generic African-language news corpus.

\begin{table}[t]
    \centering
    \resizebox{\columnwidth}{!}{
    \begin{tabular}{llrrrrrrrr}
    \toprule
        \multirow{2}{*}{\textbf{metric}} & \multirow{2}{*}{\textbf{data}} & \multicolumn{4}{l}{\textbf{\texttt{eng$\rightarrow$xxx}}} & \multicolumn{4}{l}{\textbf{\texttt{xxx$\rightarrow$eng}}} \\
        \cmidrule(lr){3-6}\cmidrule(lr){7-10}
        & & \textbf{\texttt{hau}} & \textbf{\texttt{lug}} & \textbf{\texttt{yor}} & \textbf{\texttt{zul}} & \textbf{\texttt{hau}} & \textbf{\texttt{lug}} & \textbf{\texttt{yor}} & \textbf{\texttt{zul}} \\
    \midrule
        \multirow{3}{*}{\textit{COMET}} & MF & 55.7 & 62.4 & 57.6 & 59.1 & 61.0 & 57.6 & 55.4 & 61.6 \\
         & AS-MT & \textbf{62.8} & \textbf{69.6} & \textbf{67.0} & \textbf{65.6} & \textbf{67.7} & \textbf{65.7} & \textbf{64.5} & \textbf{68.2} \\
         & ~~~+MF & 56.7 & 65.9 & 60.9 & 59.1 & 61.7 & 58.9 & 56.9 & 61.9 \\
    \midrule
        \multirow{3}{*}{\textit{ChrF}} & MF & 54.5 & 40.5 & 15.1 & 62.6 & 57.2 & 48.9 & 46.3 & 65.8 \\
         & AS-MT & 66.3 & \textbf{50.8} & 17.5 & 65.6 & 63.3 & 54.3 & 51.5 & 68.4 \\
         & ~~~+MF & \textbf{67.4} & 50.1 & \textbf{17.6} & \textbf{66.6} & \textbf{63.7} & \textbf{55.1} & \textbf{52.6} & \textbf{68.6} \\
    \bottomrule
    \end{tabular}
    }
    \caption{Cross-domain transfer for NLLB-1.3B fine-tuned on AfriScience-MT (\texttt{AS-MT}), MAFAND-MT (\texttt{MF}), or their union (\texttt{+MF}); COMET and ChrF on the four MAFAND-supported language pairs in both directions.}
    \label{tab:mafand_nllb}
\end{table}

\begin{table*}[t]
\centering
\small
\renewcommand{\arraystretch}{1.05}
\begin{tabular}{lrrrrrrrrrrrrr}
\toprule
\textbf{Model} & \multicolumn{6}{c}{\textbf{eng$\to$}} & \multicolumn{6}{c}{\textbf{$\to$eng}} & \\
\cmidrule(lr){2-7}\cmidrule(lr){8-13}
   & \texttt{amh} & \texttt{hau} & \texttt{lug} & \texttt{nso} & \texttt{yor} & \texttt{zul} & \texttt{amh} & \texttt{hau} & \texttt{lug} & \texttt{nso} & \texttt{yor} & \texttt{zul} & \textbf{Avg} \\
\midrule
\multicolumn{14}{l}{\textit{Zero-shot}} \\
\midrule
Gemini-1.5 & 27.2 & 19.2 & 37.3 & 32.7 & 31.2 & 25.4 & 39.0 & 39.0 & 40.6 & 41.3 & 41.6 & 39.8 & 34.5 \\
GPT-4o & 16.9 & 18.8 & 33.5 & 33.7 & 27.0 & 25.3 & 37.6 & 38.1 & 39.4 & 38.5 & 38.0 & 38.7 & 32.1 \\
Gemini-3.1 & \textbf{43.6} & \textbf{39.8} & 52.5 & \textbf{51.5} & \textbf{48.9} & 35.0 & \textbf{50.4} & \textbf{51.2} & \textbf{51.6} & 51.4 & \textbf{51.5} & \textbf{52.0} & \textbf{48.3} \\
GPT-5.4 & 40.0 & 39.1 & \textbf{53.6} & 50.4 & 46.7 & \textbf{42.5} & 50.3 & 51.1 & 51.2 & \textbf{51.7} & 50.8 & 51.6 & 48.3 \\
\midrule
AfriqueLlama-8B & \emph{37.3} & \emph{39.2} & \emph{40.2} & \emph{46.7} & \emph{44.4} & \emph{42.3} & 32.2 & 36.9 & 41.6 & 39.5 & 43.6 & 39.5 & 40.3 \\
AfriqueQwen-8B & 24.1 & 31.0 & 34.5 & 40.0 & 30.5 & 27.7 & 31.0 & 29.0 & 34.6 & 33.7 & 35.7 & 28.4 & 31.7 \\
Gemma2-9B & 28.4 & 29.1 & 37.1 & 35.2 & 32.0 & 29.8 & 34.5 & 34.2 & 39.9 & 37.7 & 38.8 & 35.3 & 34.3 \\
Llama3-8B & 24.7 & 28.4 & 35.9 & 36.9 & 31.0 & 32.0 & 36.9 & \emph{44.4} & 44.2 & \emph{48.0} & 40.9 & 38.6 & 36.8 \\
Tiny-Aya-E & 18.6 & 33.6 & 34.4 & 39.4 & 39.8 & 34.8 & 46.9 & 38.0 & 44.6 & 42.7 & \emph{49.1} & 43.9 & 38.8 \\
Tiny-Aya-G & 27.6 & 32.1 & 38.1 & 35.8 & 41.7 & 35.8 & \emph{48.2} & 43.0 & 48.0 & 41.2 & 48.1 & \emph{47.2} & \emph{40.6} \\
TranslateGemma-12B & 31.6 & 31.7 & 35.6 & 40.4 & 39.0 & 39.5 & 44.5 & 41.4 & \emph{49.8} & 43.9 & 44.7 & 43.6 & 40.5 \\
\midrule\midrule
\multicolumn{14}{l}{\textit{1-shot}} \\
\midrule
Gemini-1.5 & 27.2 & 17.5 & 38.9 & 35.1 & 30.8 & 24.1 & 38.0 & 38.3 & 39.3 & 38.3 & 39.7 & 39.4 & 33.9 \\
Gemini-3.1 & \emph{41.3} & \textbf{39.4} & 51.6 & 48.3 & 44.2 & \textbf{44.9} & 51.2 & \textbf{51.2} & \textbf{51.8} & \textbf{52.2} & \textbf{51.4} & \textbf{51.9} & \textbf{48.3} \\
GPT-4o & 18.7 & 18.7 & 38.6 & 34.6 & 22.5 & 26.7 & 37.4 & 38.2 & 39.3 & 38.4 & 37.8 & 38.1 & 32.4 \\
GPT-5.4 & 39.4 & 36.8 & \textbf{52.9} & \textbf{50.3} & \textbf{46.7} & 42.5 & \textbf{51.5} & 50.7 & 51.0 & 51.3 & 51.4 & 51.0 & 48.0 \\
\midrule
AfriqueLlama-8B & \textbf{41.4} & \emph{35.4} & 40.4 & \emph{42.2} & 41.9 & \emph{42.4} & 32.2 & 40.8 & 48.9 & 45.8 & 50.3 & 42.9 & 42.0 \\
AfriqueQwen-8B & 24.7 & 27.1 & 30.5 & 29.2 & 34.4 & 31.1 & 30.0 & 27.3 & 36.7 & 36.8 & 39.7 & 33.3 & 31.7 \\
Gemma2-9B & 33.9 & 29.9 & 35.9 & 33.8 & 31.3 & 29.3 & 32.0 & 35.8 & 38.4 & 39.1 & 38.6 & 35.5 & 34.5 \\
Llama3-8B & 23.2 & 27.7 & 38.3 & 35.3 & 33.1 & 32.8 & 42.3 & 43.3 & 43.2 & 41.9 & 42.0 & 42.7 & 37.2 \\
Tiny-Aya-E & 34.7 & 32.3 & 35.8 & 34.4 & 34.7 & 32.1 & 46.5 & 31.9 & 40.1 & 37.9 & 42.8 & 34.9 & 36.5 \\
Tiny-Aya-G & 34.5 & 30.3 & 35.7 & 33.5 & 37.0 & 32.5 & 44.5 & 35.2 & 40.7 & 42.5 & 43.9 & 35.1 & 37.1 \\
TranslateGemma-12B & 36.9 & 30.9 & \emph{40.5} & 41.0 & \emph{43.7} & 36.4 & \emph{50.9} & \emph{49.1} & \emph{50.0} & \emph{47.5} & \emph{51.1} & \emph{49.8} & \emph{44.0} \\
\bottomrule
\end{tabular}
\caption{\textbf{Document-level zero-shot vs. 1-shot in-context learning (COMET)}. Two major groups, each with closed- and open-LLM models. \textbf{Bold}: per-language best performance across \textit{Zero-shot} and \textit{1-shot} setups; \emph{italic}: per-setup best performance (closed or open).}
\label{tab:doc_comet}
\end{table*}

At constant training volume, sampling-weight balancing within AfriScience-MT does not help: the natural distribution outperforms uniform sampling ($\tau{=}\infty$) by up to 8.3 chrF on English$\to$Hausa, with data-rich domains hurt and only the smallest few helped (\Cref{tab:balance_domain_eng_xxx_dual} and \Cref{fig:balance_combined}; full ablation in Appendix~\ref{app:domain_balance}).

\subsection{Document Level Performance}
\label{subsec:doc_results}

\Cref{tab:doc_comet} reports document-level COMET for all eleven LLMs at zero-shot and 1-shot ICL. GPT and Gemini lead the document-level average, both reaching 48.3 COMET, with the best open LLM (TranslateGemma-12B) at 44.0 and AfriqueLlama-8B at 42.0. GPT-4o and Gemini-1.5-Flash sit at 32--35, below both TranslateGemma-12B and AfriqueLlama-8B in both directions and both shot counts. The two-judge breakdown across accuracy, fluency, and lexical/grammatical mistakes, and the per-direction error-type distributions, are in Appendix~\ref{app:doc_llm_judge} and \Cref{fig:doc_error_dist}. Notably, the best open-weight model (TranslateGemma-12B) lands within 4.3 COMET of GPT-5.4 and Gemini-3.1-Flash-Lite and well above the older GPT-4o and Gemini-1.5-Flash, showing that competitive document-level scientific translation is attainable with a fully self-hostable model.

\section{Discussion}
\label{sec:discussion}

\paragraph{In-domain data, not scale, is decisive}
A 1.3B-parameter fine-tuned encoder-decoder model matches GPT-5.4 and Gemini-3.1-Flash-Lite to within 1 COMET at the sentence level and outperforms every fine-tuned open-weight LLM and the older GPT-4o and Gemini-1.5-Flash (\Cref{tab:main_zs_ft_comet,tab:open_vs_closed_icl_comet}), despite being orders of magnitude smaller and trained on far less text. We attribute this to the nature of the signal rather than its quantity: scientific translation turns on terminology, register, and rhetorical structure that general-purpose pretraining underrepresents and that a small amount of in-domain parallel data supplies directly. The same reading explains why breadth does not substitute for specificity: adding general-domain news (MAFAND) does not help, and flattening the domain distribution actively hurts (\Cref{tab:mafand_nllb,tab:balance_domain_eng_xxx_dual}), because news data carries neither the terminology nor the register the task rewards. For African scientific translation, the decisive resource is in-domain data, not model scale.

\paragraph{The right tool depends on granularity}
The sentence- and document-level results point to different choices. For sentence-level scientific translation, the fine-tuned seq2seq model is the strongest open option and is competitive with GPT and Gemini, even outperforming their older versions. Contrastingly, the open vs. closed gap narrows for document-level translation, with TranslateGemma-12B and AfriqueLlama-8B overtaking the older GPT-4o and Gemini-1.5-Flash (\Cref{tab:doc_comet}). Two possible explanations for this shift are: (i) a long source supply sufficient in-context signal that a general-purpose model lacks otherwise. This signifies that the model's broad competence matters less when the input is in itself informative, and (ii) prompt templating carries most of the open-LLM gain at the sentence level but has little room to help at the scale of document translation, where the closed models mostly produced flat performances across the translation pairs (\Cref{tab:doc_template_shot_ablation_comet}). The practical implication is that granularity, and not a single best system, should drive model selection for translation.

\paragraph{An evaluation lesson}
Our main results survived the choice of metrics, but that was not automatic. SSA-COMET, like other learned metrics, can reward texts that are fluent but off-target: three open models produced high COMET at the templated prompt while their chrF collapses (\Cref{tab:prompt_engineering_open}), and a multilingual NLLB checkpoint shows COMET and chrF disagreeing on which languages are hardest (\Cref{fig:multilingual_nllb}). Reporting a surface metric alongside the semantic one turned these failures into a visible signal (a large COMET--chrF gap) rather than a silent inflation of scores. We therefore treat the two metrics as complementary by design and recommend the pairing for low-resource MT evaluation, where references are few and degeneration is easy to miss.

\paragraph{Terminology is the dominant failure mode}
Across both directions and all four closed models, mistranslated content (\emph{Wrong Translation}) accounts for 77--80\% of the document-level accuracy errors flagged by our judges, far outweighing omissions (13--17\%) and hallucinated additions (5--7\%; \Cref{fig:doc_error_dist}). For scientific text, a mistranslation is most often a specialized term rendered with a generic or incorrect equivalent. Lexical mistakes remain a substantial, persistent error dimension even for the newer models (\Cref{tab:doc_evaluation}). This is precisely the gap our co-developed bilingual glossaries target (\Cref{tab:data_splits_stats}), which are largest for \yoruba (587 entries). Getting specialized terminology right, rather than avoiding omissions or hallucinations, is therefore what most separates strong from weak scientific translation in these languages.

\section{Conclusion}
\label{sec:conclusion}

We introduced \datasetnamebf, a parallel corpus for scientific translation covering six African languages (Amharic, Hausa, Luganda, Northern Sotho, Yor\`ub\'a, isiZulu) across 11 scientific domains, comprising 230 papers and 7,605 parallel sentences co-developed with science communicators and professional translators. Alongside the corpus, we will release the bilingual glossaries built during translation and the full set of predictions, evaluations, and trained adapter weights.

Our main empirical result is that a fine-tuned NLLB-1.3B model reaches an average of 67.3 SSA-COMET across the 12 directional pairs, outperforming GPT-4o (65.4) and Gemini-1.5-Flash (65.9) and every fine-tuned open-weight LLM at sentence level (\Cref{tab:main_zs_ft_comet,tab:open_vs_closed_icl_comet}). GPT and Gemini (GPT-5.4, Gemini-3.1-Flash-Lite) marginally outperform it, by 0.7--1.0 COMET (68.3 and 68.0). The bilingual fine-tuning advantage holds for both seq2seq and Gemma LoRA families; Llama-3.1-8B-Instruct LoRA is the exception where bilingual and multilingual training tie. At document level, GPT and Gemini lead at 48.3 average COMET, ahead of the best open-weight LLM (TranslateGemma-12B at 44.0) and of GPT-4o and Gemini-1.5-Flash (32--35). Domain-specific data is essential: substituting general-domain MAFAND-MT underperforms by 5--10 COMET, and forced domain balance hurts more than it helps. The directional asymmetry favouring Target$\to$English over English$\to$Target persists across setups. We discuss accessibility implications in Appendix~\ref{app:accessibility}.

We provide a replicable template for decolonizing science access in Africa by demonstrating that high-quality scientific translation into African languages is achievable with modest, openly available resources. Our open-source dataset, terminology glossaries, and trained models enable researchers, educators, and policymakers to expand access to scientific knowledge in indigenous African languages. Future work will (i) expand to full-text translation rather than expert summarization, (ii) develop morphology-aware models to close the persistent English$\to$Target gap, (iii) establish pan-African terminology standardization with community input, and (iv) extend coverage to additional languages and scientific domains.

\section*{Limitations}

\paragraph{Coverage} AfriScience-MT covers six African languages spoken by about 200M people but excludes the vast majority of the more than 2,000 languages spoken on the continent. The 11 scientific domains span our existing partner translators' interests rather than a representative sample of African scientific output; Chemistry ($N{=}46$ test sentences), Indigenous Knowledge ($N{=}92$), and Engineering ($N{=}93$) are under-represented relative to Health ($N{=}2,605$). Results may not generalize to other languages, underrepresented domains, or resource regimes where the domain-specific signal is weaker than ours.


\paragraph{Evaluation} We rely on SSA-COMET, chrF, and two LLM judges (GPT-5.5 and Gemini-3.1-Flash-Lite) rather than systematic human MQM by native bilingual domain experts. Three factors limit the risk this introduces. First, SSA-COMET is fine-tuned on human MT-quality judgments for sub-Saharan African languages and is the strongest learned metric reported for them, and we pair it with chrF precisely to expose the fluent-but-off-target failures a semantic metric can reward (\Cref{tab:prompt_engineering_open}). Second, our main rankings hold under both metrics, so they are not metric artefacts. Third, the corpus itself was produced and verified by professional translators with per-language lead review, so human quality control sits in the data even where it is absent from system scoring. We additionally report inter-judge agreement (\Cref{tab:judge_agreement}) and treat absolute error counts as judge-dependent. Systematic expert evaluation of system outputs remains future work.

\paragraph{Modelling} We compare a fixed set of M2M100, NLLB, and open-weight LLM families and do not study retrieval-augmented translation, post-editing, or chain-of-thought prompting; our LoRA sweep covers ranks ${4,8,16,32,64,128}$ but fixes the remaining adapter choices ($\alpha{=}r$, 4-bit NF4, one target-module set). These choices keep the comparison controlled and matched across systems, which is what our claims require: our conclusions concern the data and the deployment setting -- in-domain data is decisive, and the best system depends on granularity -- not a specific fine-tuning recipe, and the rank sweep already covers the main capacity axis. Stronger per-system engineering would likely raise absolute scores but is orthogonal to these conclusions.

\paragraph{Closed-model coverage} Our closed baselines are two families at two generations each (GPT-4o, GPT-5.4; Gemini-1.5-Flash, Gemini-3.1-Flash-Lite); we did not evaluate other proprietary systems or higher-capacity tiers. We therefore make no claim that these are the strongest LLMs available; all comparisons are limited to the named models. Because our central result is that a small fine-tuned model is competitive with much larger proprietary ones, broader closed-model coverage would strengthen rather than overturn the argument.


\section*{Ethical Considerations}

This work engages with the decolonization of scientific knowledge in Africa. Translation is a pragmatic bridge but not a substitute for the broader project of valuing African languages in research, supporting African researchers, and recognising indigenous knowledge systems as sources of scientific authority. Terminology development decisions affect how concepts are understood and integrated into knowledge systems, and our translator-led approach may not reflect every community's preferences; we present the glossaries as contributions to ongoing conversations rather than authoritative standards.

Our trained models are research prototypes and are unsuitable for high-stakes deployment (medical, educational, or legal) without additional validation. Translation errors could cause patient harm, teach incorrect concepts, or affect rights. Users should validate quality with native speakers before deployment and engage African language communities as partners, not data sources.

Professional translators were fairly compensated. Compensation terms were discussed transparently and mutually agreed upon after translators were informed about the nature and intended use of the work. We acknowledge that this compensation is a one-time payment, not an ongoing benefit. AfriScience-MT will be openly released; the openness brings risks of commercial exploitation, low-quality re-deployment, and surveillance. Users are encouraged to respect community preferences, acknowledge translator labour, and explore equitable benefit-sharing in future work. Training over 100 models for this paper consumed considerable computational resources with associated environmental cost. Finally, we commit to maintaining the released artefacts for at least five years.

\bibliography{anthology,custom}

\appendix


\section{Data processing}
\label{app:data_processing}

\subsection{Translation cleaning}
We applied a rigorous data-cleaning procedure to ensure translation quality. 
First, we removed untranslated sentences: cases where English text appeared unchanged in the target-language columns or vice versa (typically section names such as \textit{Abstract} or \textit{Introduction}).  Second, we filtered out single-word sentences, which were often section headers (e.g., ``Title,'' ``Abstract,'' ``Introduction'') that provided little or no useful training signal even when translated.  For sentence-level translation, we deduplicated English sentences while retaining their first occurrence and corresponding translations across all target languages to prevent the model from memorizing repeated content.  At the document level, we retained such repetitions to preserve the natural flow and coherence of the document structure.

\subsection{Glossary normalization}
Scientific terminology glossaries underwent preprocessing to facilitate consistent terminology matching during translation. We removed extraneous whitespace and converted all entries to lowercase, enabling case-insensitive matching between glossary terms and their occurrences in translated texts. This normalization proved essential for accurately identifying and evaluating scientific terminology usage across languages.

\subsection{Data splits}
All models used identical 80/10/10 splits for training, validation, and test sets. Critically, splitting was performed at the \textit{domain level} rather than randomly: we allocated complete papers (with all their sentences) to each split while ensuring every split contained at least one document from each of the 11 scientific domains. This domain-stratified approach maintains domain diversity across splits and prevents domain-specific overfitting, enabling robust evaluation of cross-domain generalization. Table~\ref{tab:data_splits_stats} provides detailed statistics for each split, showing sentence and document counts per domain.

\section{Fine-tuning details}
\subsection{NLLB multilingual model}
We fine-tuned the \texttt{NLLB-200-distilled-600M} model using Fairseq with multilingual training across all 12 language-pair directions simultaneously (\texttt{eng}$\rightarrow$\texttt{xxx} and \texttt{xxx}$\rightarrow$\texttt{eng}). The model was initialized from a pretrained checkpoint (\texttt{facebook/nllb-200-distilled-600M}) and fine-tuned using the default configuration\footurl{https://github.com/facebookresearch/fairseq/tree/main/examples/multilingual} provided in \citet{tang-etal-2021-multilingual}. The default configuration uses a temperature sampling ($\tau=1.5$), source-side encoder language tokens with decoder language tokens, label smoothing (0.2), Adam optimizer ($\beta_1=0.9$, $\beta_2=0.98$, $\epsilon=1e-06$), inverse square root learning rate schedule with initial LR of $3e-05$, 2,500 warmup updates, maximum 40,000 updates, dropout (0.3), attention dropout (0.1), weight decay (0.0), maximum tokens per batch (1,024), and gradient accumulation (update-freq=2). We saved checkpoints every 5,000 updates, retaining the last 10 checkpoints, and used seed 222 for reproducibility. The training employed encoder and decoder layer normalization before sublayers with embedding layer normalization.

\subsection{Bilingual M2M100 and NLLB models} 
We finetuned M2M100 (418M and 1.2B) and NLLB (600M and 1.3B) models using the Hugging Face Transformers library with finetuning script provided in \citet{adelani-etal-2022-thousand}, using default configuration. We maintained consistent hyperparameters across models: batch size per device (16), gradient accumulation steps (2), learning rate ($5e-05$), 10 training epochs with early stopping (patience=3), AdamW optimizer, linear learning rate scheduler with 500 warmup steps, and evaluation at each epoch. Models used fp16 mixed precision training and were saved based on best validation loss.

\subsection{Open-source LLMs}
In addition to the rank $r{=}64$ fine-tuned models, we additionally train $r{=}4$ for the rank ablation in \Cref{tab:lora_comet}. LoRA targets the standard \texttt{q,k,v,o} and \texttt{gate,up,down} projections. The training uses max sequence length 2048, batch size 4 per device, gradient accumulation 4, AdamW 8-bit, linear schedule with 5 warm-up steps, learning rate $2{\times}10^{-4}$, weight decay 0.01, max 10 epochs with patience-2 early stopping, bf16 (fp16 fallback), seed 3407. Instruction-tuned models use their native chat template and we mask the loss to the assistant span.

\section{Evaluation}
\label{app:evaluation}

All models were evaluated using a unified external evaluation script rather than relying on built-in evaluation metrics from training frameworks. During preliminary experiments, we observed systematic discrepancies between training-framework evaluations and standardized metric implementations, particularly in \citet{adelani-etal-2022-thousand}. These inconsistencies could misreport translation quality and render cross-model comparisons unreliable. By decoupling evaluation from training, we ensured consistent and comparable metrics across all experimental conditions.\footnote{This error was also reported by other users on the Hugging Face forums: \url{https://discuss.huggingface.co/t/inconsistent-bleu-score-between-test-metrics-test-bleu-and-written-to-file-test-metric-predictions/6352}}

\subsection{Translation generation}
\label{app:translation_generation}
\paragraph{Fine-tuned seq2seq models} After training completion, we generated predictions for all test sets using each model's best checkpoint (selected by validation loss). Predictions were decoded using beam search with a beam size of 5 and a maximum length of 200 tokens.

\paragraph{Fine-tuned LLMs} For fine-tuned open-weight LLMs, we used vLLM with greedy decoding, a maximum of 512 new tokens, repetition penalty 1.08, and the model's chat-template stop tokens. We added the repetition penalty and explicit stop tokens after the Tiny-Aya-Global degeneration.

\paragraph{Zero- and few-shot LLMs} For closed models (GPT-4o, Gemini-1.5-Flash, GPT-5.4, Gemini-3.1-Flash-Lite), we accessed the APIs with temperature set to 0 for deterministic outputs and maximum tokens set to 512. For the seven open-weight LLMs (Llama3-8B, Gemma2-9B-IT, AfriqueLlama-8B, AfriqueQwen-8B, Tiny-Aya-Earth, Tiny-Aya-Global, TranslateGemma-12B), we used greedy decoding with temperature 0 and a maximum generation length of 512 tokens. Few-shot examples (when applicable) were selected from the training set either \textit{randomly} or using \textit{semantic similarity search} based on OpenAI embeddings \cite{neelakantan2022text}. All examples were formatted consistently across models using the prompt templates described in Templates~\ref{prompt:sent_templates} and \ref{prompt:doc_templates}.

\paragraph{Document-level translation} For document-level experiments, we increased maximum generation length to 1024 tokens for LLMs to accommodate longer translations. Seq2seq models maintained beam search with a beam size of 5.

\paragraph{Decoding controls and manual inspection}
All fine-tuned LLM inference runs greedy decoding with the model's chat-template end-of-turn tokens as explicit stop conditions and a mild repetition penalty (1.08). We adopted these controls after a routine spot-check caught one adapter (Tiny-Aya-Global) generating sentence-doubled outputs (\texttt{<sent>.<sent>}) on the six $\rightarrow$English pairs. The duplication halved COMET while leaving chrF essentially flat: on \texttt{lug}$\to$\texttt{eng}, for instance, COMET fell from $63.9$ to $21.9$, but chrF moved only from $47.0$ to $44.9$. N-gram overlap survives duplication, so a surface-form metric did not flag the failure. The lesson is broader than one bug: we recommend pairing scalar metrics with a quick manual look at a sample of generations and treating large inter-metric (COMET vs. chrF here) gaps as a cheap degeneration signal in their own right. We document the same pattern at inference for three open LLMs in \Cref{tab:prompt_engineering_open}.

\paragraph{ICL setup}
All seven open LLMs and four closed models (GPT-4o, Gemini-1.5-Flash, and the newer GPT-5.4 and Gemini-3.1-Flash-Lite) are evaluated in zero-shot and ICL. At the sentence level we adopt a single configuration across all eleven LLMs -- template~5, 10-shot, $T{=}0$ -- for a fair open-vs-closed comparison; we justify each component below. At the document level, where source inputs are long, we cap in-context examples at one and report zero-shot and 1-shot ICL with a vanilla prompt at $T{=}0$ (\Cref{tab:doc_comet}); the document-level template/shot ablation (templates~1--3, \{0,1\}-shot) is in \Cref{tab:doc_template_shot_ablation_comet}.

We ran inference on the open LLMs using vLLM \cite{kwon2023efficient} with greedy decoding ($T{=}0$) and a uniform set of anti-degeneration controls -- repetition penalty 1.08, the chat-template's native stop tokens, and standard EOS handling -- which removed the worst-case fluency-without-meaning loops seen in early runs without measurably hurting the well-behaved models. We selected ICL exemplars from the training split by random sampling or semantic similarity over OpenAI embeddings \cite{neelakantan2022text}; the random-retrieval baseline is the vanilla 5-shot column in \Cref{tab:prompt_engineering_open}. For the closed models we additionally run retrieval and temperature ablations (\Cref{tab:closed_ablations_comet}). Templates and a sample 5-shot prompt are in Templates~\ref{prompt:sent_templates},~\ref{prompt:doc_templates} and Example~\ref{prompt:prompt_example}.

\paragraph{Configuration justification}
We justify the three components of the sentence-level configuration below.
\begin{description}
\item[Why template~5:] Across the five prompt templates we trialled, template~5 ties or leads on the majority of directional pairs for GPT-4o and Gemini-1.5-Flash (\Cref{fig:gpt_gemini_eng_xxx,fig:gpt_gemini_xxx_eng}) and is within 0.5 COMET of every per-row maximum across the six reference models in \Cref{tab:template_shot_ablation_comet}; templates~3 and~5 dominate the alternatives (1--3 COMET spread) over templates~1, 2, 4, making template~5 both the best and the most reliable choice across providers.
\item[Why 10-shot:] At fixed template~5, moving from 5- to 10-shot buys at most 1.4 COMET (on Llama3-8B) and at most 0.2 COMET for the four closed models; 10$\to$20-shot buys at most 0.2 COMET on any reference model, below the 1-COMET threshold we set for a higher shot count.
\item[Why $T{=}0$:] Replaying each closed model's main configuration at $T{=}0.6$ moves the average by at most $\pm 0.7$ COMET (\Cref{tab:closed_ablations_comet}); we default to $T{=}0$ (greedy) for reproducibility.
\end{description}

\subsection{Metrics}
\label{app:metric}

\paragraph{Reporting Policy}
We report a hierarchy of three automatic metrics with the following policy:
\begin{enumerate}
    \item \textbf{SSA-COMET} \cite{li-etal-2025-ssa} is the primary metric in every body table. It is the African-language-tuned variant of COMET and is the only metric that operates at the \emph{semantic} level for these languages. Bold/italic highlights and the textual claims in the body track SSA-COMET unless explicitly stated.
    \item \textbf{chrF} \cite{popovic-2015-chrf} is reported side-by-side with SSA-COMET when (a) the COMET vs. chrF ranking diverges on a finding worth surfacing, or (b) the divergence itself is the finding (degeneration -- COMET~$\gg$~chrF). Otherwise, we omit chrF from the body to keep tables compact.
    \item \textbf{BLEU} is computed for every prediction set and will be released alongside the code and predictions, but is not reported in the paper.
\end{enumerate}

\begin{figure*}[t]
  \centering

  \begin{subfigure}[t]{0.49\textwidth}
    \centering
    \includegraphics[width=\linewidth]%
                    {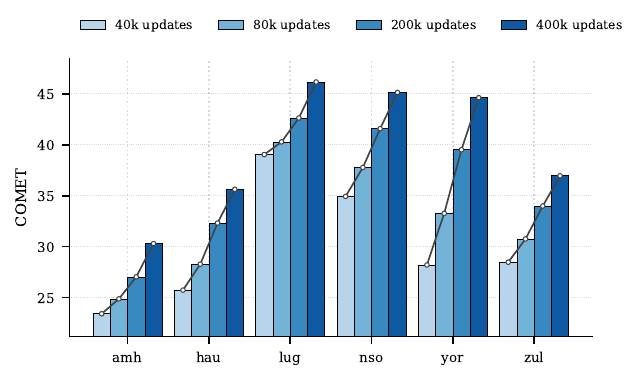}
    \caption{\texttt{eng$\to$xxx} (COMET)}
    \label{fig:mnllb_eng_xxx_comet}
  \end{subfigure}\hfill
  \begin{subfigure}[t]{0.49\textwidth}
    \centering
    \includegraphics[width=\linewidth]{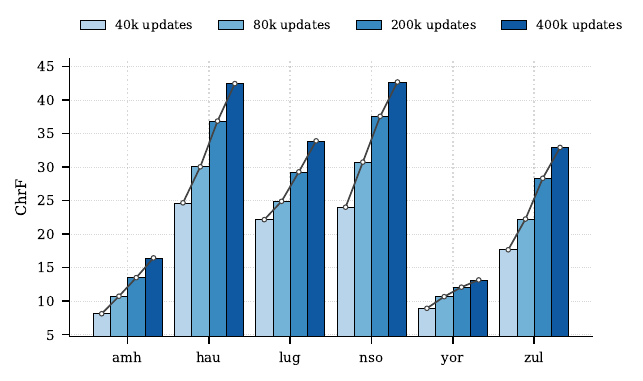}
    \caption{\texttt{eng$\to$xxx} (ChrF)}
    \label{fig:mnllb_eng_xxx_chrf}
  \end{subfigure}

  \vspace{4pt}

  \begin{subfigure}[t]{0.49\textwidth}
    \centering
    \includegraphics[width=\linewidth]{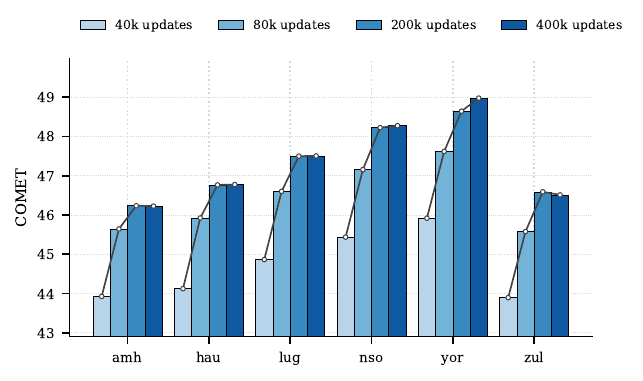}
    \caption{\texttt{xxx$\to$eng} (COMET)}
    \label{fig:mnllb_xxx_eng_comet}
  \end{subfigure}\hfill
  \begin{subfigure}[t]{0.49\textwidth}
    \centering
    \includegraphics[width=\linewidth]%
                    {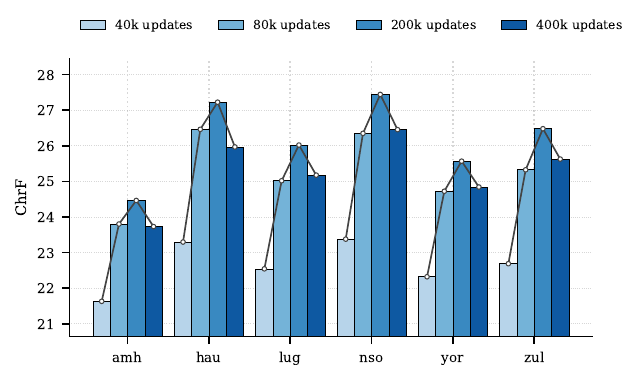}
    \caption{\texttt{xxx$\to$eng} (ChrF)}
    \label{fig:mnllb_xxx_eng_chrf}
  \end{subfigure}

  \caption{Multilingual NLLB-600M performance as a function of training updates (40k, 80k, 200k, 400k); each language is plotted as its own bar group with an in-group trend line. COMET and ChrF disagree substantially on which languages are hardest.}

  \label{fig:multilingual_nllb}
\end{figure*}

This policy follows the lessons drawn from the metric divergence on the multilingual NLLB-600M experiment (\Cref{fig:multilingual_nllb}, where COMET and chrF disagree on which languages are hardest) and on the open-LLM degeneration diagnosed via large COMET~$-$~chrF gaps in \Cref{tab:prompt_engineering_open}.

\paragraph{Why SSA-COMET and chrF} Learned metrics such as COMET \cite{rei-etal-2020-comet} correlate substantially better with human judgment than surface-overlap metrics \cite{kocmi-etal-2021-ship}, but COMET is trained largely on European-language judgments and underperforms on African languages, which motivated African-tuned variants. SSA-COMET \cite{li-etal-2025-ssa} is fine-tuned on human MT-quality annotations spanning 14 African language pairs and is the strongest learned metric reported for sub-Saharan African MT, surpassing earlier African-tuned metrics and rivaling much larger LLM judges; we adopt it as our primary, semantic-level metric. We retain chrF \cite{popovic-2015-chrf} as a surface check: character $n$-gram overlap is tokenization-independent and correlates better than BLEU for morphologically rich languages, and a surface metric complements a neural one, since a large COMET--chrF gap exposes fluent but off-target generation that COMET tolerates. BLEU is computed for every prediction set and released with the code, but not reported, being the least reliable of the three here.

\paragraph{Computation}
All predictions and references were processed through our standardized evaluation script, which computes chrF \citep{popovic-2015-chrf} and SSA-COMET \citep{li-etal-2025-ssa} scores. For sentence-level evaluation, we computed metrics on individual sentence pairs. For document-level evaluation, we computed metrics on complete document translations (concatenated sentences). This single-script approach eliminates implementation-specific variations and ensures that performance differences reflect true model capabilities rather than evaluation inconsistencies.

\paragraph{LLM-as-judge setup}
For document-level translation quality assessment, we additionally employed two LLM judges, GPT-5.5 and Gemini-3.1-Flash-Lite, with temperature 0 and consistent prompts across all models and language pairs, following the methodology of \citet{sun-llm-evaluator-2025} and \citet{alabi2025afridocmtdocumentlevelmtcorpus}. Each judge scored four dimensions: fluency (1--5, higher is better), accuracy errors, lexical mistakes, and grammatical mistakes (counts, lower is better). The scores reported in \Cref{tab:doc_evaluation} are per-judge means; inter-judge agreement is analysed in \Cref{app:doc_llm_judge}. The evaluation prompts are presented in Templates~\ref{prompt:fluency_eval}--\ref{prompt:cohesion_eval}.

\subsection{Closed-model ablation setup}
\label{app:closed_ablations_notes}

The closed-model ablation is reported on English$\to$Target only (\Cref{tab:closed_ablations_comet}). The four blocks are constructed so vanilla and templated numbers are never compared inside the same block.

\paragraph{Reported, vanilla} Vanilla zero-shot vs. vanilla 5-shot at $T{=}0$. Vanilla 5-shot is available only for GPT-5.4 and Gemini-3.1-Flash-Lite; GPT-4o and Gemini-1.5-Flash appear as zero-shot only.

\paragraph{Reported, templated} Templated zero-shot vs. templated 10-shot at $T{=}0$. Templated zero-shot is the per-pair mean of the three 0-shot paper templates (1, 2, 4). Templated 10-shot is each model's best configuration: template~5 for Gemini-1.5-Flash, GPT-5.4, and Gemini-3.1-Flash-Lite; template~3 for GPT-4o.

\paragraph{Retrieval ablation} The same templated 10-shot run as in the previous block, contrasted against semantic-similarity 10-shot exemplars at $T{=}0$.

\paragraph{Temperature ablation} $T{=}0$ vs. $T{=}0.6$ at the two templated configurations comparable across all four closed models: templated zero-shot (mean of templates 1, 2, 4 at 0-shot) and each model's best templated 10-shot. Vanilla 0-shot at $T{=}0.6$ is excluded for comparability.

\paragraph{Highlighting} \textbf{Bold} marks the per-column maximum within a block. \emph{Italic} marks the per-language best across the rows of one model version (e.g. ICL vs. zero-shot within the same prompt-style class for that model); italic is suppressed when only one row in the model's group has data for that column.


\section{Fine-tuning: LoRA rank, bilingual vs. multilingual}
\label{app:lora_rank_and_scope}

\subsection{LoRA rank}
\Cref{tab:lora_comet} compares LoRA $r{=}4$ against $r{=}64$ for the six fine-tuned open LLMs. At $r{=}4$, AfriqueLlama-8B is the strongest model on Amharic$\to$English (67.6 COMET, the table-wide maximum); its African pre-training already provides a starting point that low-rank adapters quickly tune. At $r{=}64$, Tiny-Aya-G dominates 9 of 12 results, showing that larger adapters unlock its capacity for the other languages. AfriqueQwen-8B trails the other open LLMs at both ranks, suggesting that LoRA fine-tuning alone is not enough to bring it up to the same level on this task.

\subsection{Bilingual vs. multilingual fine-tuning}
\Cref{tab:multilingual_vs_bilingual} compares bilingual fine-tuning against multilingual for three model families. For NLLB, bilingual fine-tuning is stronger than the (short, early-stage) multilingual NLLB-600M checkpoints (67.2 vs 39.8 avg COMET at the final checkpoint). For Gemma-2-9B LoRA, bilingual leads multilingual by about 3.7 COMET or 5 chrF at the best matching rank. \emph{Llama-3.1-8B is the exception among the three families we tested:} bilingual and multilingual LoRA end up essentially tied at about 57 COMET across matching ranks. The $r{=}64$ Llama bilingual row is also an outlier in the opposite direction -- its chrF is higher than neighbouring ranks but with lower COMET, suggesting that it is matching surface character patterns of the reference without preserving meaning, and it should be cited cautiously. Overall, the bilingual advantage is robust on seq2seq and Gemma, with Llama as a caveat that we discuss in \Cref{sec:discussion}.

\subsection{Domain balance and the "less balance is more" result}
\label{app:domain_balance}
\begin{figure*}[t]
  \centering
  \begin{subfigure}[t]{0.49\textwidth}
    \centering
    \includegraphics[width=\linewidth]{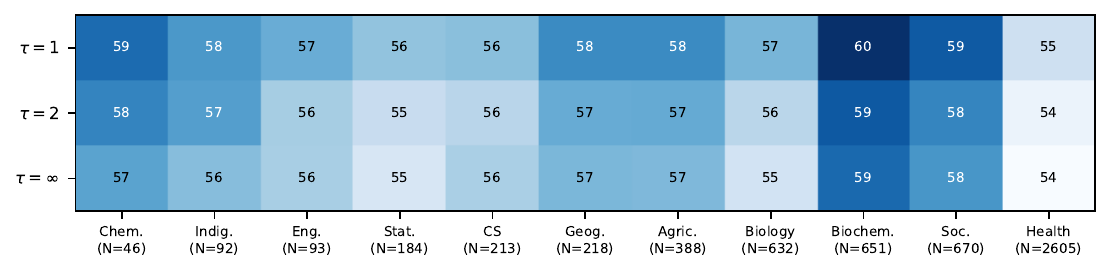}
    \caption{Per-domain chrF for $\tau{\in}\{1, 2, \infty\}$ on eng$\to$xxx, averaged over the six target languages. Each row is one sampling regime; each column is one domain. Darker shading indicates higher chrF.}
    \label{fig:balance_heatmap}
  \end{subfigure}\hfill
  \begin{subfigure}[t]{0.49\textwidth}
    \centering
    \includegraphics[width=\linewidth]{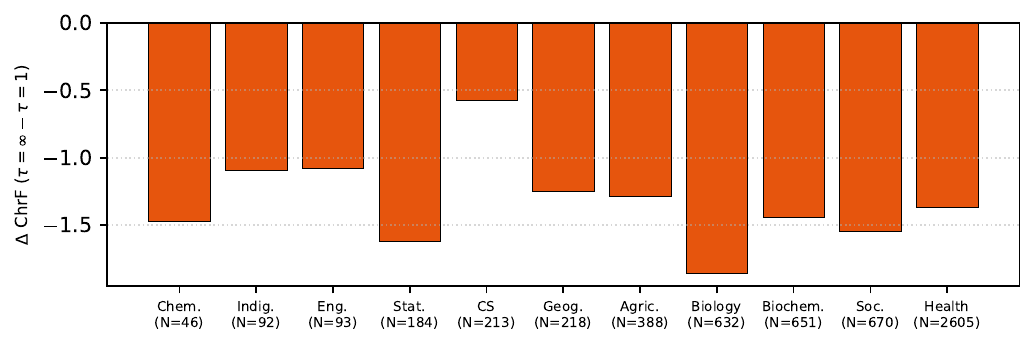}
    \caption{Per-domain $\Delta$ chrF $(\tau{=}\infty - \tau{=}1)$, ordered by training-set domain size (ascending). Negative bars indicate domains where uniform sampling underperforms the natural distribution; positive bars indicate domains where it helps.}
    \label{fig:balance_delta_by_size}
  \end{subfigure}
  \caption{Domain-balance ablation visualised: the heatmap (left) shows that uniform sampling ($\tau{=}\infty$) is broadly worse than natural sampling ($\tau{=}1$) across most (regime, domain) combinations; the size-ordered $\Delta$ chrF view (right) shows that the uniform regime helps only the smallest few domains and hurts the data-rich ones, with the largest single-domain drop on Health ($N{=}2{,}605$). Companion to
  \Cref{tab:balance_domain_eng_xxx_dual}.}
  \label{fig:balance_combined}
\end{figure*}

Furthermore, \Cref{tab:balance_domain_eng_xxx_dual} ablates the \emph{within-AfriScience-MT} sampling distribution at constant total training volume (5,792 sentences). Three setups were compared: $\tau{=}1$ (natural domain proportions), $\tau{=}2$ (sqrt smoothing), and $\tau{=}\infty$ (uniform across all 11 domains). Sampling-weight balancing keeps every training sentence available and only changes how often each is drawn, which lets us isolate the effect of \emph{distribution} from the effect of \emph{data volume} -- unlike hard subsetting, which would also discard usable training signal from the larger domains. We found that forced uniformity hurts: at the per-domain level, the deltas relative to natural sampling are predominantly red (down) on medium and large domains and only scattered blue (up) on the smallest ones (visualized in \Cref{fig:balance_heatmap}). The largest single-domain drop is 8.3 chrF on English$\to$Hausa. \Cref{fig:balance_delta_by_size} re-orders the per-domain $\Delta$ chrF $(\tau{=}\infty - \tau{=}1)$ by training-set size, showing that the uniform regime helps only the smallest few domains and hurts the data-rich ones. The takeaway is that fine-tuning benefits more from naturally distributed, abundant data -- where the largest domains supply disproportionate signal -- than from artificial domain balance, even when the artificial distribution "protects" the smallest domains. Both chrF and COMET agree on this ranking, with COMET showing a slightly weaker effect.

\section{LLM ICL}
\label{app:open_lift_decomposition}

\subsection{Prompt template vs. shot count}
\Cref{tab:prompt_engineering_open} separates the \emph{prompt-template} effect from the \emph{shot-count} effect on the open LLMs by reporting vanilla 0-shot, vanilla 5-shot, templated 5-shot, and templated 10-shot side-by-side. For the two clean models, almost the entire vanilla-to-templated gain (about 5--10 COMET) is attributable to the template, not to the shot count: $\Delta$ from 5-shot to 10-shot at fixed template~5 is at most 1.5 COMET / 2.4 chrF. For the three degenerate open models, templating produces a large positive $\Delta$ COMET paired with a large negative $\Delta$ chrF ($-21.7$, $-5.9$, $-23.9$ respectively), confirming the degeneration. The vanilla baseline is therefore the more honest reading for those three models, and we recommend treating their templated numbers as upper bounds on adequacy rather than faithful translations.

TranslateGemma-12B is a telling case: as a translation-specialized model it is sensitive to prompt format, and the templated 10-shot prompt pushes it off-distribution into fluent off-target output (chrF drops from 32.9 at the vanilla prompt to 12.2 when templated). The same model is the strongest open system at the document level (\Cref{tab:doc_comet}), where the simpler vanilla 0/1-shot prompt keeps it faithful, so its weak sentence-level ICL numbers reflect prompt-format mismatch rather than an inability to translate.

\subsection{Older vs. newer closed models}
\label{app:closed_evolution}
\Cref{tab:closed_evolution} compares each closed-model family (Gemini, GPT) between its older and newer releases at the two main configurations: sentence-level ICL (template~5, 10-shot, $T{=}0$) and document-level vanilla 0-shot. Both families improve in both directions on COMET; the document-level gain (+13--16 COMET) is substantially larger than the sentence-level gain (+2--3 COMET). chrF moves in the same direction, with the largest gain concentrated on English$\to$Target at document level (+9--20 chrF).

\subsection{Closed-model ablations}
\label{app:closed_api_ablations}
\begin{figure*}[htbp]
    \centering

    \begin{subfigure}[b]{\textwidth}
       \includegraphics[width=\textwidth]{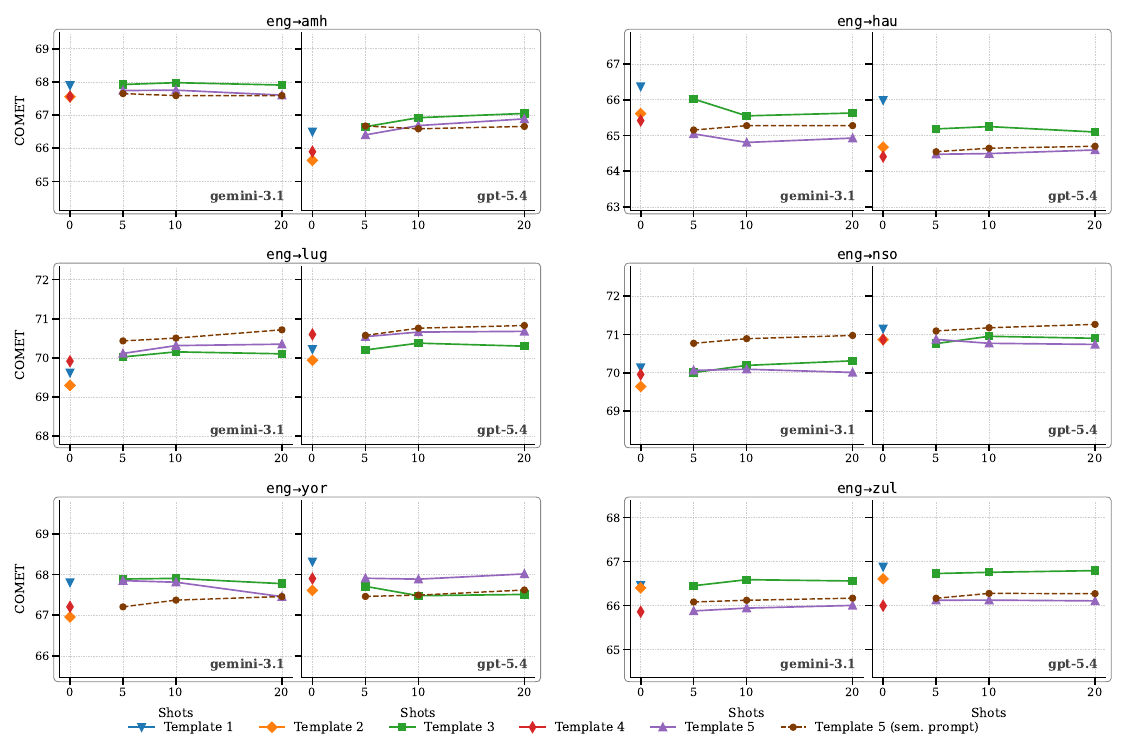}
       \caption{COMET}
       \label{fig:gpt_gemini_eng_xxx_comet}
    \end{subfigure}

    \vspace{0.5cm}

    \begin{subfigure}[b]{\textwidth}
       \includegraphics[width=\textwidth]{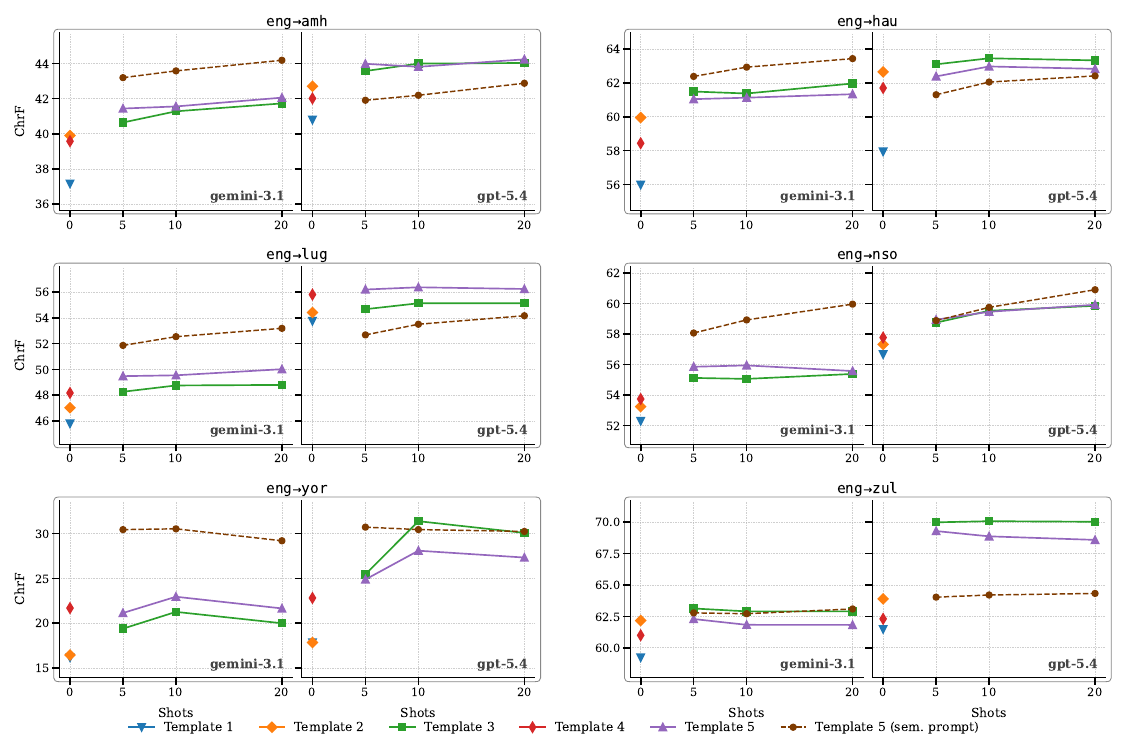}
       \caption{ChrF}
       \label{fig:gpt_gemini_eng_xxx_chrf}
    \end{subfigure}

    \caption{GPT-4o vs. Gemini-1.5-Flash on \texttt{eng$\to$xxx}, across five prompt templates and shot counts $\in\{0,5,10,20\}$}
    \label{fig:gpt_gemini_eng_xxx}
\end{figure*}

\begin{figure*}[htbp]
    \centering

    \begin{subfigure}[b]{\textwidth}
       \includegraphics[width=\textwidth]{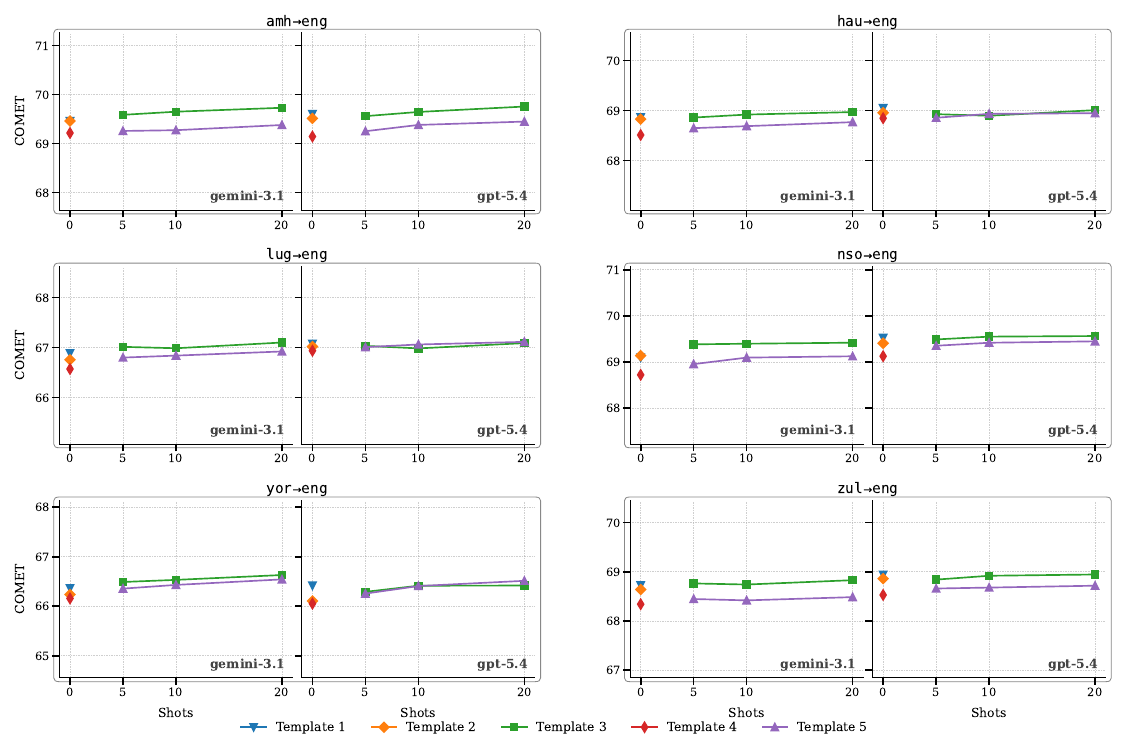}
       \caption{COMET}
       \label{fig:gpt_gemini_xxx_eng_comet}
    \end{subfigure}

    \vspace{0.5cm}

    \begin{subfigure}[b]{\textwidth}
       \includegraphics[width=\textwidth]{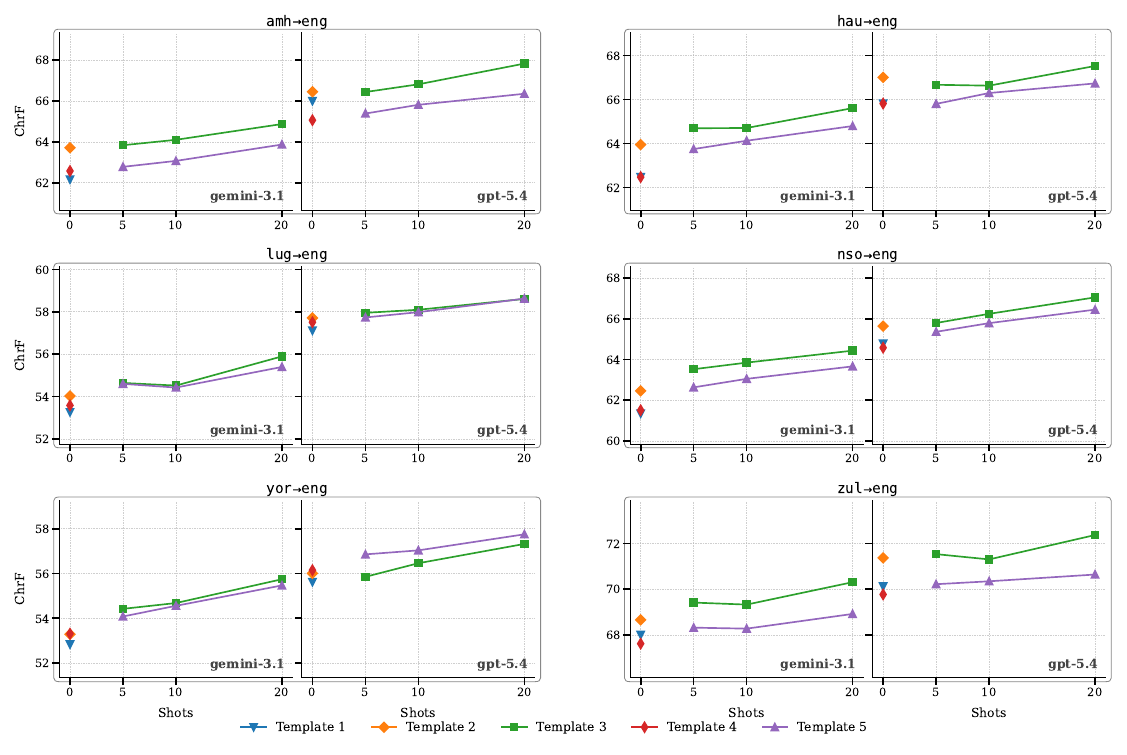}
       \caption{ChrF}
       \label{fig:gpt_gemini_xxx_eng_chrf}
    \end{subfigure}

    \caption{GPT-4o vs. Gemini-1.5-Flash on \texttt{xxx$\to$eng}, across five prompt templates and shot counts $\in\{0,5,10,20\}$}
    \label{fig:gpt_gemini_xxx_eng}
\end{figure*}

\Cref{tab:closed_ablations_comet} shows that at the adopted templated 10-shot configuration, neither retrieval strategy (template-fixed semantic examples vs. random) nor decoding temperature ($T{=}0$ vs. $T{=}0.6$) shifts the average by more than 1 COMET for any of the four closed models. The closed-model rows of \Cref{tab:template_shot_ablation_comet} give the full template$\times$shot grid at $T{=}0$; for GPT-4o and Gemini-1.5-Flash, \Cref{fig:gpt_gemini_eng_xxx,fig:gpt_gemini_xxx_eng} visualise the same grid. The template choice contributes a 1--3 COMET spread, which motivated adopting template~5 in the main results. For closed models, prompt template choice dominates both retrieval and temperature.

\subsection{Full chrF results}
\label{app:chrf}
For transparency, we report the chrF counterpart of every COMET table in the paper, so that every COMET-based result can be cross-checked against the surface metric. At the sentence level, \Cref{tab:main_zs_ft_chrf} mirrors the zero-shot and fine-tuned comparison of \Cref{tab:main_zs_ft_comet}, \Cref{tab:open_vs_closed_icl_chrf} the open-vs-closed ICL comparison of \Cref{tab:open_vs_closed_icl_comet}, and \Cref{tab:lora_chrf} the LoRA rank ablation of \Cref{tab:lora_comet}. At the document level, \Cref{tab:doc_chrf} mirrors \Cref{tab:doc_comet}. For the closed-model configuration ablations, \Cref{tab:closed_ablations_chrf}, \Cref{tab:template_shot_ablation_chrf}, and \Cref{tab:doc_template_shot_ablation_chrf} mirror \Cref{tab:closed_ablations_comet}, \Cref{tab:template_shot_ablation_comet}, and \Cref{tab:doc_template_shot_ablation_comet}.

\section{Document-level LLM-judge results}
\label{app:doc_llm_judge}
\Cref{tab:doc_evaluation} reports the per-judge mean of two LLM judges (GPT-5.5 and Gemini-3.1-Flash-Lite) for the four closed models across six document-level configurations and four dimensions: fluency (1--5, higher is better), accuracy errors, lexical mistakes, and grammatical mistakes (counts, lower is better). On English$\to$Target, GPT-5.4 and Gemini-3.1-Flash-Lite more than halve the per-judge accuracy-error count of GPT-4o and Gemini-1.5-Flash, to 1.5 from 3.1--3.4 errors, and similarly cut lexical mistakes to 1.5 from 2.2--2.4. Fluency follows the same ordering: the two newer models average 3.8/5 against 2.6/5. On Target$\to$English, fluency saturates near 4.0/5 across all four models, accuracy errors drop relative to English$\to$Target, and the two newer models retain a clear lead on accuracy. The directional asymmetry holds across all four models, with English$\to$Target consistently harder.

\begin{figure*}[t]
    \centering

    \begin{subfigure}[t]{\textwidth}
        \centering
        \includegraphics[width=\textwidth, clip, trim=0 25 0 0]{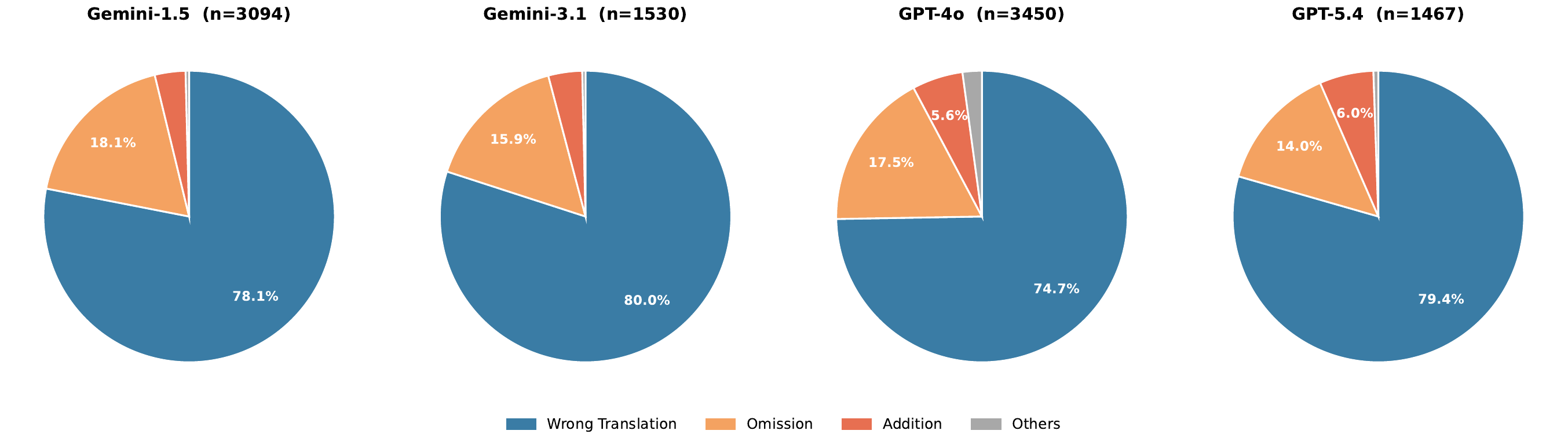}
        \caption{eng $\rightarrow$ xxx}
        \label{fig:doc_error_dist_a}
    \end{subfigure}
    \hfill

    \begin{subfigure}[t]{\textwidth}
        \centering
        \includegraphics[width=\textwidth, clip, trim=0 25 0 0]{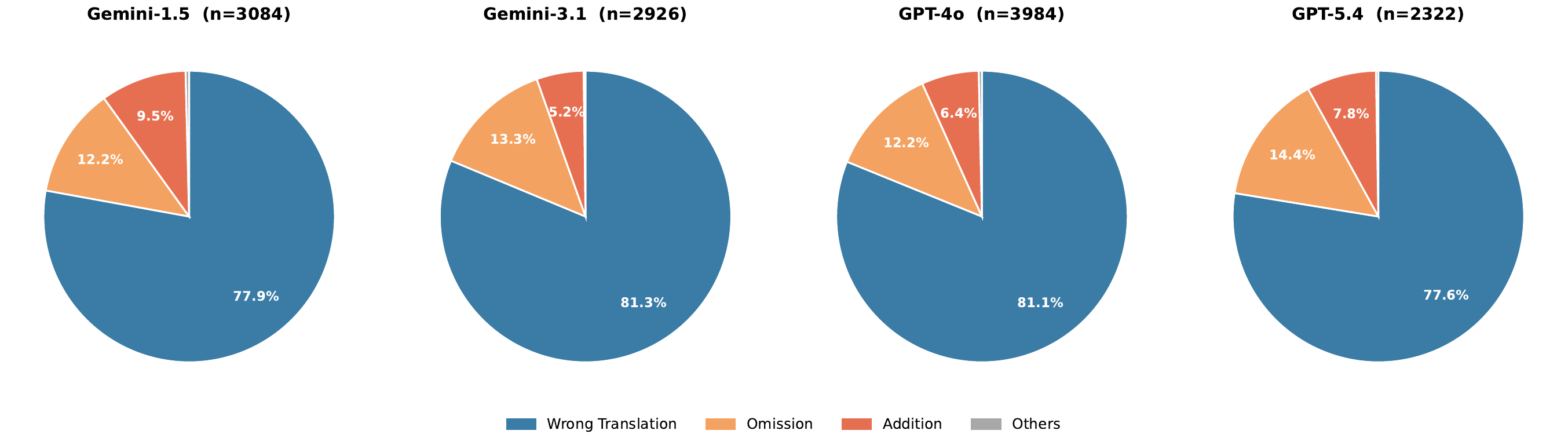}
        \caption{xxx $\rightarrow$ eng}
        \label{fig:doc_error_dist_b}
    \end{subfigure}

    \caption{Document-level accuracy error distribution for the four closed models, broken down by error category (Wrong Translation, Omission, Addition, Others). Each pie aggregates errors across the six template configurations common to the older and newer models and the six language pairs in that direction; categories are summed across two LLM judges (GPT-5.5 and Gemini-3.1-Flash-Lite). The $n$ in each pie title is the mean number of accuracy errors per judge. \textbf{Wrong Translation} (mistranslated content) and \textbf{Omission} (missing content) account for the bulk of errors in every model and direction; \textbf{Addition} (hallucinated content not in the source) is consistently small. The newer pair (GPT-5.4, Gemini-3.1-Flash-Lite) more than halves the per-judge mean error count of the older pair on \texttt{eng}$\to$\texttt{xxx}.}
    \label{fig:doc_error_dist}
\end{figure*}

\Cref{fig:doc_error_dist} aggregates accuracy errors by category (Wrong Translation, Omission, Addition, Others) across both judges, both directions, and the six configurations. \emph{Wrong Translation} and \emph{Omission} account for the bulk of errors in every model and direction; \emph{Addition} is consistently small. This reduction in accuracy errors is concentrated in Wrong Translation and Omission on English$\to$Target.

\paragraph{Inter-judge agreement}
The two judges agree closely on fluency (Pearson 0.84 across 288 paired observations; \Cref{tab:judge_agreement}) but interpret the error-count scale differently: the Gemini-3.1-Flash-Lite judge reports 7--11 times more accuracy, lexical, and grammatical errors than the GPT-5.5 judge, and the two judges show near-zero or negative correlation on the count-based metrics. The qualitative finding that drives the main result -- GPT-5.4 and Gemini-3.1-Flash-Lite more than halve the per-judge accuracy-error count of GPT-4o and Gemini-1.5-Flash on English$\to$Target -- holds for both judges (GPT-5.5: a 25-fold reduction, 0.47 to 0.02; Gemini-3.1-Flash-Lite: a 2-fold reduction, 6.02 to 2.95). The lexical and grammatical-error reductions on the same direction hold for the GPT-5.5 judge but are smaller for the Gemini-3.1-Flash-Lite judge (33\% and 23\%). We therefore report the per-judge mean in \Cref{tab:doc_evaluation} for transparency and treat the absolute counts as judge-dependent.

\section{Implications for accessibility}
\label{app:accessibility}
A fine-tuned NLLB-1.3B can be run on a single mid-range GPU, retrained inexpensively, hosted by African institutions, and audited end-to-end. By contrast, the closed models that lead at sentence level are charged per token and cannot be deployed offline. NLLB-1.3B is within 1 COMET of GPT-5.4 in 8 of 12 directional pairs, ties on \texttt{eng}$\to$\texttt{nso}, and outperforms Gemini-3.1-Flash-Lite on \texttt{eng}$\to$\texttt{nso} and \texttt{eng}$\to$\texttt{zul}; it also dominates every fine-tuned open-weight LLM. The case for open, fine-tuned seq2seq models for African scientific translation is therefore also a case for accessibility: a model within reach of the leading proprietary systems is also the cheapest to run, the most auditable, and the most amenable to community ownership. Closed models remain useful as benchmarks and as a fallback for languages where parallel data does not yet exist, but they should not anchor the deployment story.

\begin{table*}[t]
\centering
\small
\renewcommand{\arraystretch}{1.05}
\begin{tabular}{lllrrrrrrr}
\toprule
\textbf{Model} & \textbf{Version} & \textbf{Setting} & \multicolumn{6}{c}{\textbf{eng$\to$}} & \\
\cmidrule(lr){4-9}
   &    &    & \texttt{amh} & \texttt{hau} & \texttt{lug} & \texttt{nso} & \texttt{yor} & \texttt{zul} & \textbf{Avg} \\
\midrule
\multicolumn{10}{l}{\textit{Reported, vanilla prompts ($T{=}0$): ZS vs ICL (5-shot)}} \\
\midrule
\multirow{3}{*}{Gemini} & \multirow{1}{*}{1.5} & ZS & 57.9 & 55.8 & 59.4 & 62.2 & 57.5 & 53.9 & 57.8 \\
\cmidrule(lr){2-10}
 & \multirow{2}{*}{3.1} & ZS & 55.6 & 55.1 & 58.7 & 53.0 & 53.0 & 52.2 & 54.6 \\
 &  & ICL (5-shot) & \emph{60.9} & \emph{56.0} & \emph{61.9} & \emph{53.5} & \emph{53.6} & \emph{53.7} & \emph{56.6} \\
\cmidrule(lr){1-10}
\multirow{3}{*}{GPT} & \multirow{1}{*}{4o} & ZS & 39.4 & 56.8 & 60.7 & 62.6 & 54.6 & 58.3 & 55.4 \\
\cmidrule(lr){2-10}
 & \multirow{2}{*}{5.4} & ZS & 60.1 & 58.4 & 64.3 & \textbf{\emph{66.7}} & 55.9 & 63.2 & 61.4 \\
 &  & ICL (5-shot) & \textbf{\emph{62.8}} & \textbf{\emph{62.5}} & \textbf{\emph{65.8}} & 65.6 & \textbf{\emph{61.1}} & \textbf{\emph{65.2}} & \textbf{\emph{63.8}} \\
\midrule\midrule
\multicolumn{10}{l}{\textit{Reported, templated prompts ($T{=}0$): ZS (avg \texttt{template\{1,2,4\}\textunderscore{}0shot}) vs ICL (best 10-shot)}} \\
\midrule
\multirow{4}{*}{Gemini} & \multirow{2}{*}{1.5} & ZS (avg \texttt{template\{1,2,4\}}) & 64.3 & 62.3 & 65.2 & 67.3 & 64.3 & 61.7 & 64.2 \\
 &  & ICL (10-shot, template5) & \emph{64.9} & \emph{62.5} & \emph{67.0} & \emph{68.5} & \emph{64.6} & \emph{62.2} & \emph{65.0} \\
\cmidrule(lr){2-10}
 & \multirow{2}{*}{3.1} & ZS (avg \texttt{template\{1,2,4\}}) & 67.7 & \textbf{\emph{65.8}} & 69.6 & 69.9 & 67.3 & \emph{66.2} & 67.8 \\
 &  & ICL (10-shot, template5) & \textbf{\emph{67.8}} & 64.8 & \emph{70.3} & \emph{70.1} & \emph{67.8} & 65.9 & \textbf{\emph{67.8}} \\
\cmidrule(lr){1-10}
\multirow{4}{*}{GPT} & \multirow{2}{*}{4o} & ZS (avg \texttt{template\{1,2,4\}}) & 51.5 & 63.3 & 67.0 & 67.8 & 62.6 & 65.3 & 62.9 \\
 &  & ICL (10-shot, template3) & \emph{53.9} & \emph{63.4} & \emph{68.3} & \emph{68.6} & \emph{65.0} & \emph{65.4} & \emph{64.1} \\
\cmidrule(lr){2-10}
 & \multirow{2}{*}{5.4} & ZS (avg \texttt{template\{1,2,4\}}) & 66.0 & \emph{65.0} & 70.3 & \textbf{\emph{71.0}} & \textbf{\emph{67.9}} & \textbf{\emph{66.5}} & \emph{67.8} \\
 &  & ICL (10-shot, template5) & \emph{66.7} & 64.5 & \textbf{\emph{70.7}} & 70.8 & 67.9 & 66.1 & 67.8 \\
\midrule\midrule
\multicolumn{10}{l}{\textit{Retrieval ablation: templated vs. semantic-similarity, $T{=}0$}} \\
\midrule
\multirow{4}{*}{Gemini} & \multirow{2}{*}{1.5} & 10-shot (templated, template5) & 64.9 & 62.5 & 67.0 & 68.5 & 64.6 & 62.2 & 65.0 \\
 &  & 10-shot (semantic similarity) & \emph{65.0} & \emph{62.9} & \emph{68.4} & \emph{69.5} & \emph{65.8} & \emph{63.5} & \emph{65.8} \\
\cmidrule(lr){2-10}
 & \multirow{2}{*}{3.1} & 10-shot (templated, template5) & \textbf{\emph{67.8}} & 64.8 & 70.3 & 70.1 & \emph{67.8} & 65.9 & 67.8 \\
 &  & 10-shot (semantic similarity) & 67.6 & \textbf{\emph{65.3}} & \emph{70.5} & \emph{70.9} & 67.4 & \emph{66.1} & \textbf{\emph{68.0}} \\
\cmidrule(lr){1-10}
\multirow{4}{*}{GPT} & \multirow{2}{*}{4o} & 10-shot (templated, template3) & 53.9 & \emph{63.4} & 68.3 & 68.6 & 65.0 & \emph{65.4} & 64.1 \\
 &  & 10-shot (semantic similarity) & \emph{57.2} & 63.4 & \emph{69.2} & \emph{69.8} & \emph{65.1} & 65.3 & \emph{65.0} \\
\cmidrule(lr){2-10}
 & \multirow{2}{*}{5.4} & 10-shot (templated, template5) & \emph{66.7} & 64.5 & 70.7 & 70.8 & \textbf{\emph{67.9}} & 66.1 & 67.8 \\
 &  & 10-shot (semantic similarity) & 66.6 & \emph{64.6} & \textbf{\emph{70.8}} & \textbf{\emph{71.2}} & 67.5 & \textbf{\emph{66.3}} & \emph{67.8} \\
\midrule\midrule
\multicolumn{10}{l}{\textit{Temperature ablation ($T{=}0$ vs $T{=}0.6$, all templated)}} \\
\midrule
\multirow{8}{*}{Gemini} & \multirow{4}{*}{1.5} & ZS ($T{=}0$) & 64.3 & 62.3 & 65.2 & 67.3 & 64.3 & 61.7 & 64.2 \\
 &  & ZS ($T{=}0.6$) & 64.2 & 62.3 & 64.9 & 67.1 & 64.3 & 61.6 & 64.1 \\
 &  & 10-shot (template5) $T{=}0$ & \emph{64.9} & \emph{62.5} & \emph{67.0} & \emph{68.5} & 64.6 & \emph{62.2} & \emph{65.0} \\
 &  & 10-shot (template5) $T{=}0.6$ & 64.8 & 62.4 & 66.8 & 68.4 & \emph{64.7} & 62.2 & 64.9 \\
\cmidrule(lr){2-10}
 & \multirow{4}{*}{3.1} & ZS ($T{=}0$) & 67.7 & \textbf{\emph{65.8}} & 69.6 & 69.9 & 67.3 & \emph{66.2} & 67.8 \\
 &  & ZS ($T{=}0.6$) & 67.4 & 65.5 & 69.1 & 69.6 & 67.1 & 66.0 & 67.4 \\
 &  & 10-shot (template5) $T{=}0$ & \textbf{\emph{67.8}} & 64.8 & \emph{70.3} & \emph{70.1} & \emph{67.8} & 65.9 & \emph{67.8} \\
 &  & 10-shot (template5) $T{=}0.6$ & 67.4 & 64.7 & 70.1 & 69.9 & 67.5 & 65.7 & 67.6 \\
\cmidrule(lr){1-10}
\multirow{8}{*}{GPT} & \multirow{4}{*}{4o} & ZS ($T{=}0$) & 51.5 & 63.3 & 67.0 & 67.8 & 62.6 & 65.3 & 62.9 \\
 &  & ZS ($T{=}0.6$) & 50.2 & 62.9 & 66.3 & 67.2 & 62.4 & 64.9 & 62.3 \\
 &  & 10-shot (template3) $T{=}0$ & \emph{53.9} & \emph{63.4} & \emph{68.3} & \emph{68.6} & \emph{65.0} & \emph{65.4} & \emph{64.1} \\
 &  & 10-shot (template3) $T{=}0.6$ & 51.9 & 62.8 & 67.8 & 68.1 & 64.6 & 65.2 & 63.4 \\
\cmidrule(lr){2-10}
 & \multirow{4}{*}{5.4} & ZS ($T{=}0$) & 66.0 & 65.0 & 70.3 & 71.0 & 67.9 & \textbf{\emph{66.5}} & 67.8 \\
 &  & ZS ($T{=}0.6$) & 66.0 & \emph{65.2} & 70.2 & \textbf{\emph{71.0}} & 67.9 & 66.5 & \textbf{\emph{67.8}} \\
 &  & 10-shot (template5) $T{=}0$ & \emph{66.7} & 64.5 & \textbf{\emph{70.7}} & 70.8 & 67.9 & 66.1 & 67.8 \\
 &  & 10-shot (template5) $T{=}0.6$ & 66.7 & 64.4 & 70.6 & 70.7 & \textbf{\emph{68.0}} & 66.0 & 67.7 \\
\bottomrule
\end{tabular}
\caption{Closed-model ablations on \texttt{eng}$\to$\texttt{xxx} (COMET) for the four closed models (older: GPT-4o, Gemini-1.5-Flash; newer: GPT-5.4, Gemini-3.1-Flash-Lite). Four blocks: vanilla ZS vs ICL; templated ZS vs ICL; retrieval (templated vs semantic 10-shot); and temperature ($T{=}0$ vs $T{=}0.6$, all templated). \textbf{Bold}: per-block column maximum. \emph{Italic}: per-version best across that model's rows. See \Cref{app:closed_ablations_notes} for the full setup and the rationale behind each block.}
\label{tab:closed_ablations_comet}
\end{table*}

\begin{table*}[t]
\centering
\setlength{\tabcolsep}{4pt}
\renewcommand{\arraystretch}{1.05}
\begin{tabular}{llrrrrrr}
\toprule
\textbf{Model} & \textbf{Version} & \multicolumn{3}{c}{\textbf{chrF}} & \multicolumn{3}{c}{\textbf{COMET}} \\
\cmidrule(lr){3-5}\cmidrule(lr){6-8}
 &  & \texttt{eng$\to$} & \texttt{$\to$eng} & \textbf{Avg} & \texttt{eng$\to$} & \texttt{$\to$eng} & \textbf{Avg} \\
\midrule
\multicolumn{8}{l}{\textit{Sentence-level ICL (template~5, 10-shot, $T{=}0$)}} \\
\midrule
\multirow{3}{*}{Gemini} & 1.5 & 44.0 & 55.5 & 49.7 & 64.9 & 67.0 & 65.9 \\
 & 3.1 & \textbf{48.8} & \textbf{61.2} & \textbf{55.0} & 67.8 & 68.1 & 68.0 \\
 & $\Delta$ & +4.8 & +5.8 & +5.3 & +2.9 & +1.2 & +2.0 \\
\cmidrule(lr){1-8}
\multirow{3}{*}{GPT} & 4o & 43.9 & 57.0 & 50.5 & 63.5 & 67.4 & 65.4 \\
 & 5.4 & 48.5 & 57.7 & 53.1 & \textbf{67.8} & \textbf{68.7} & \textbf{68.3} \\
 & $\Delta$ & +4.6 & +0.6 & +2.6 & +4.3 & +1.3 & +2.8 \\
\midrule\midrule
\multicolumn{8}{l}{\textit{Document-level (vanilla \texttt{doc\textunderscore{}0shot})}} \\
\midrule
\multirow{3}{*}{Gemini} & 1.5 & 27.1 & 66.4 & 46.8 & 28.8 & 40.2 & 34.5 \\
 & 3.1 & 47.1 & 67.7 & 57.4 & 45.2 & \textbf{51.4} & \textbf{48.3} \\
 & $\Delta$ & +20.0 & +1.3 & +10.6 & +16.4 & +11.2 & +13.8 \\
\cmidrule(lr){1-8}
\multirow{3}{*}{GPT} & 4o & 40.9 & 68.0 & 54.4 & 25.9 & 38.4 & 32.1 \\
 & 5.4 & \textbf{50.0} & \textbf{70.8} & \textbf{60.4} & \textbf{45.4} & 51.1 & 48.3 \\
 & $\Delta$ & +9.2 & +2.8 & +6.0 & +19.5 & +12.7 & +16.1 \\
\bottomrule
\end{tabular}
\caption{Closed models evolution, grouped by family (Gemini, GPT) with older $\to$ newer top-to-bottom and a $\Delta$ row reporting (newer $-$ older) per column. Two blocks: the main sentence-level configuration (\texttt{template5\textunderscore{}10shot}, $T{=}0$) and the main document-level configuration (vanilla \texttt{doc\textunderscore{}0shot}). Per-direction averages over the six target languages, plus an overall Avg. The $\Delta$ rows make the closed-model generation-over-generation improvement visible at a glance; both families improve on COMET in both directions, with the doc-level COMET gain (+13--16) much larger than the sentence-level gain (+2--3). \textbf{Bold}: column maximum within each block.}
\label{tab:closed_evolution}
\end{table*}

\begin{table*}[t]
\centering
\setlength{\tabcolsep}{4pt}
\renewcommand{\arraystretch}{1.05}
\begin{tabular}{lrrrrrrrrrr}
\toprule
\textbf{Model} & \multicolumn{5}{c}{\textbf{COMET}} & \multicolumn{5}{c}{\textbf{chrF}} \\
\cmidrule(lr){2-6}\cmidrule(lr){7-11}
 & V$_0$ & V$_5$ & T$_{10}$ & $\Delta_{V\to T}$ & $\Delta_{T_{s:~5\to 10}}$ & V$_0$ & V$_5$ & T$_{10}$ & $\Delta_{V\to T}$ & $\Delta_{T_{s:~5\to 10}}$ \\
\midrule
Gemini-3.1 & 57.0 & 60.3 & 68.0 & +7.6 & +0.0 & 44.1 & 54.5 & \textbf{55.0} & +0.5 & +0.3 \\
GPT-5.4 & \textbf{65.0} & \textbf{66.0} & \textbf{68.3} & +2.2 & +0.0 & \textbf{54.0} & \textbf{55.3} & 53.1 & -2.2 & +0.4 \\
\midrule
Gemma2-9B & 39.4 & 39.4 & 51.4 & +12.1 & +0.7 & 26.2 & 30.6 & 34.9 & \textbf{+4.4} & +1.1 \\
Llama3-8B & 34.9 & 35.2 & 46.9 & +11.8 & \textbf{+1.5} & 20.9 & 24.6 & 27.7 & +3.1 & \textbf{+2.4} \\
\cmidrule(lr){1-11}
AfriqueLlama-8B & 37.8 & 39.1 & 48.8 & +9.8 & -- & 25.0 & 29.0 & 7.3 & -21.7 & -- \\
AfriqueQwen-8B & 33.7 & 34.7 & 43.3 & +8.6 & -- & 11.3 & 17.9 & 12.0 & -5.9 & -- \\
Tiny-Aya-E & 36.4 & 41.6 & 55.3 & \textbf{+13.7} & -- & 30.9 & 35.9 & 34.0 & -1.9 & -- \\
Tiny-Aya-G & 36.3 & 46.6 & 55.1 & +8.6 & -- & 30.8 & 35.6 & 33.8 & -1.9 & -- \\
TranslateGemma-12B & 37.7 & 37.6 & 45.7 & +8.2 & -- & 32.9 & 36.1 & 12.2 & -23.9 & -- \\
\bottomrule
\end{tabular}
\caption{Vanilla vs. templated ICL on the two newer closed models (top: Gemini-3.1-Flash-Lite, GPT-5.4) and the seven open LLMs (bottom), bidirectional avg over the 12 \texttt{eng}$\leftrightarrow$\texttt{xxx} pairs at $T{=}0$, COMET and chrF side-by-side. V$_0$: vanilla 0-shot (\texttt{basic}); V$_5$: vanilla 5-shot (\texttt{few\textunderscore{}shot}, random retrieval); T$_{10}$: \texttt{template5\textunderscore{}10shot} (the locked body configuration); $\Delta_{V\to T} = $T$_{10} - $V$_5$ is the main gain from vanilla to templated; $\Delta_{T_{s:~5\to 10}} = $T$_{s=10} - $T$_{s=5}$ isolates the shot-count effect (template fixed at \texttt{template5}). Within each group, models with all columns populated are listed above the within-group divider (Gemma2-9B-IT, Llama3-8B for open; Gemini-3.1, GPT-5.4 for closed); the remaining open LLMs render ``--'' for T$_5$-derived entries. The COMET~$\gg$~chrF divergence on AfriqueLlama-8B, AfriqueQwen-8B, and TranslateGemma-12B exposes a degeneration signal: large positive $\Delta_{V\to T}$ COMET paired with large negative $\Delta_{V\to T}$ chrF ($-21.7$ / $-5.9$ / $-23.9$), indicating fluent off-target generations rather than real translations. \textbf{Bold}: column maximum across all 9 models.}
\label{tab:prompt_engineering_open}
\end{table*}

\onecolumn
\begingroup
\setlength{\tabcolsep}{4pt}
\renewcommand{\arraystretch}{1.05}
\begin{longtable}{llrrrrrrrrrrrrr}
\toprule
\textbf{Model} & \textbf{Config} & \multicolumn{6}{c}{\textbf{eng$\to$}} & \multicolumn{6}{c}{\textbf{$\to$eng}} & \\
\cmidrule(lr){3-8}\cmidrule(lr){9-14}
  &   & \textbf{\texttt{amh}} & \textbf{\texttt{hau}} & \textbf{\texttt{lug}} & \textbf{\texttt{nso}} & \textbf{\texttt{yor}} & \textbf{\texttt{zul}} & \textbf{\texttt{amh}} & \textbf{\texttt{hau}} & \textbf{\texttt{lug}} & \textbf{\texttt{nso}} & \textbf{\texttt{yor}} & \textbf{\texttt{zul}} & \textbf{Avg} \\
\midrule
\endfirsthead
\multicolumn{15}{l}{\textit{(continued)}} \\
\toprule
\textbf{Model} & \textbf{Config} & \multicolumn{6}{c}{\textbf{eng$\to$}} & \multicolumn{6}{c}{\textbf{$\to$eng}} & \\
\cmidrule(lr){3-8}\cmidrule(lr){9-14}
  &   & \textbf{\texttt{amh}} & \textbf{\texttt{hau}} & \textbf{\texttt{lug}} & \textbf{\texttt{nso}} & \textbf{\texttt{yor}} & \textbf{\texttt{zul}} & \textbf{\texttt{amh}} & \textbf{\texttt{hau}} & \textbf{\texttt{lug}} & \textbf{\texttt{nso}} & \textbf{\texttt{yor}} & \textbf{\texttt{zul}} & \textbf{Avg} \\
\midrule
\endhead
\midrule
\multicolumn{15}{r}{\textit{(continued on next page)}} \\
\endfoot
\bottomrule
\addlinespace[10pt]
\caption{Template/shot ablation at $T{=}0$ for the six reference models: two open (Llama3-8B, Gemma2-9B-IT) and four closed models (older: GPT-4o, Gemini-1.5-Flash; newer: GPT-5.4, Gemini-3.1-Flash-Lite) (COMET). Each row is a (template, shot-count) configuration (template number first, shot count as subscript $s$); the underlined row \texttt{5}$_{s=10}$ is the configuration used throughout the paper. \textbf{Bold}: per-language column maximum within each model block. The bottom \textbf{avg} row of each block reports the per-language average across configurations. Shot counts are capped at 20; the 10$\to$20-shot gain is negligible.} \label{tab:template_shot_ablation_comet} \\
\endlastfoot
\multirow{10}{*}{\rotatebox{90}{Gemini-1.5}} & 1 & 64.3 & 62.6 & 65.2 & 67.3 & 64.3 & 61.8 & 68.6 & 67.6 & 65.3 & 66.7 & 63.0 & 67.3 & 65.3 \\
 & 2 & 64.1 & 61.9 & 64.6 & 67.1 & 63.8 & 61.5 & 68.3 & 67.3 & 64.9 & 66.5 & 62.8 & 67.0 & 65.0 \\
 & 3$_{s=5}$ & 64.6 & 62.8 & 66.2 & 68.4 & 63.6 & 62.3 & 68.8 & 67.9 & 65.5 & 67.4 & 64.0 & 67.7 & 65.8 \\
 & 3$_{s=10}$ & 64.8 & \textbf{62.9} & 66.4 & \textbf{68.5} & 64.3 & 62.5 & 68.8 & 67.9 & 65.4 & 67.3 & 64.1 & 67.7 & 65.9 \\
 & 3$_{s=20}$ & 64.8 & 62.9 & 66.6 & 68.5 & 64.8 & \textbf{62.5} & \textbf{68.9} & 67.9 & 65.5 & 67.5 & 64.2 & \textbf{67.8} & \textbf{66.0} \\
 & 4 & 64.5 & 62.4 & 65.7 & 67.4 & 64.9 & 61.9 & 68.5 & 67.8 & 65.5 & 66.9 & 64.0 & 67.3 & 65.6 \\
 & 5$_{s=5}$ & 64.8 & 62.4 & 66.8 & 68.3 & 64.5 & 62.1 & 68.7 & 67.9 & \textbf{65.6} & 67.3 & 64.4 & 67.5 & 65.9 \\
 & \underline{5$_{s=10}$} & 64.9 & 62.5 & \textbf{67.0} & 68.5 & 64.6 & 62.2 & 68.7 & 67.9 & 65.6 & 67.5 & \textbf{64.5} & 67.6 & 66.0 \\
 & 5$_{s=20}$ & \textbf{64.9} & 62.5 & 66.9 & 68.5 & \textbf{65.0} & 62.2 & 68.8 & \textbf{68.0} & 65.6 & \textbf{67.5} & 64.5 & 67.6 & 66.0 \\
\cmidrule(l){2-15}
 & \textbf{avg} & 64.6 & 62.5 & 66.2 & 68.1 & 64.4 & 62.1 & 68.7 & 67.8 & 65.4 & 67.2 & 63.9 & 67.5 & 65.7 \\
\midrule
\multirow{10}{*}{\rotatebox{90}{GPT-4o}} & 1 & 51.5 & \textbf{63.9} & 66.9 & 67.9 & 62.0 & \textbf{65.5} & 67.0 & 68.4 & 66.2 & 68.4 & 65.4 & 68.4 & 65.1 \\
 & 2 & 51.1 & 63.2 & 66.6 & 67.5 & 61.7 & 65.5 & 67.2 & 68.4 & 66.3 & 68.4 & 65.4 & 68.4 & 65.0 \\
 & 3$_{s=5}$ & 53.0 & 63.6 & 67.9 & 68.4 & 65.0 & 65.4 & 67.4 & 68.7 & \textbf{66.6} & 68.8 & 65.6 & 68.5 & 65.7 \\
 & 3$_{s=10}$ & 53.9 & 63.4 & 68.3 & 68.6 & 65.0 & 65.4 & 67.5 & 68.7 & 66.6 & \textbf{68.8} & 65.7 & 68.6 & 65.9 \\
 & 3$_{s=20}$ & 54.4 & 63.4 & 68.3 & 68.8 & 65.5 & 65.4 & \textbf{67.6} & \textbf{68.7} & 66.6 & 68.8 & \textbf{65.7} & \textbf{68.6} & \textbf{66.0} \\
 & 4 & 52.1 & 62.7 & 67.5 & 67.9 & 63.9 & 64.8 & 66.7 & 68.1 & 66.0 & 68.1 & 65.3 & 68.0 & 65.1 \\
 & 5$_{s=5}$ & 53.6 & 62.9 & 68.4 & 68.3 & 65.6 & 64.7 & 67.1 & 68.3 & 66.3 & 68.4 & 65.6 & 68.2 & 65.6 \\
 & \underline{5$_{s=10}$} & \textbf{54.9} & 62.7 & 68.5 & 68.6 & 65.5 & 64.8 & 67.2 & 68.4 & 66.3 & 68.4 & 65.6 & 68.3 & 65.8 \\
 & 5$_{s=20}$ & 54.5 & 63.0 & \textbf{68.7} & \textbf{68.9} & \textbf{65.7} & 64.9 & 67.3 & 68.5 & 66.4 & 68.6 & 65.6 & 68.3 & 65.9 \\
\cmidrule(l){2-15}
 & \textbf{avg} & 53.2 & 63.2 & 67.9 & 68.3 & 64.4 & 65.2 & 67.2 & 68.5 & 66.4 & 68.5 & 65.6 & 68.4 & 65.6 \\
\midrule
\multirow{10}{*}{\rotatebox{90}{Gemini-3.1}} & 1 & 67.9 & \textbf{66.4} & 69.6 & 70.1 & 67.8 & 66.5 & 69.5 & 68.9 & 66.9 & 69.1 & 66.4 & 68.7 & 68.1 \\
 & 2 & 67.5 & 65.6 & 69.3 & 69.6 & 67.0 & 66.4 & 69.5 & 68.8 & 66.8 & 69.1 & 66.2 & 68.6 & 67.9 \\
 & 3$_{s=5}$ & 67.9 & 66.0 & 70.0 & 70.0 & 67.9 & 66.4 & 69.6 & 68.9 & 67.0 & 69.4 & 66.5 & 68.8 & 68.2 \\
 & 3$_{s=10}$ & \textbf{68.0} & 65.5 & 70.2 & 70.2 & \textbf{67.9} & \textbf{66.6} & 69.6 & 68.9 & 67.0 & 69.4 & 66.5 & 68.7 & 68.2 \\
 & 3$_{s=20}$ & 67.9 & 65.6 & 70.1 & \textbf{70.3} & 67.8 & 66.6 & \textbf{69.7} & \textbf{69.0} & \textbf{67.1} & \textbf{69.4} & \textbf{66.6} & \textbf{68.8} & \textbf{68.2} \\
 & 4 & 67.6 & 65.4 & 69.9 & 70.0 & 67.2 & 65.9 & 69.2 & 68.5 & 66.6 & 68.7 & 66.2 & 68.3 & 67.8 \\
 & 5$_{s=5}$ & 67.7 & 65.0 & 70.1 & 70.1 & 67.8 & 65.9 & 69.3 & 68.6 & 66.8 & 69.0 & 66.4 & 68.4 & 67.9 \\
 & \underline{5$_{s=10}$} & 67.8 & 64.8 & 70.3 & 70.1 & 67.8 & 65.9 & 69.3 & 68.7 & 66.8 & 69.1 & 66.4 & 68.4 & 68.0 \\
 & 5$_{s=20}$ & 67.6 & 64.9 & \textbf{70.4} & 70.0 & 67.5 & 66.0 & 69.4 & 68.8 & 66.9 & 69.1 & 66.5 & 68.5 & 68.0 \\
\cmidrule(l){2-15}
 & \textbf{avg} & 67.8 & 65.5 & 70.0 & 70.0 & 67.6 & 66.2 & 69.4 & 68.8 & 66.9 & 69.1 & 66.4 & 68.6 & 68.0 \\
\midrule
\multirow{10}{*}{\rotatebox{90}{GPT-5.4}} & 1 & \textbf{66.8} & \textbf{66.0} & 70.6 & \textbf{71.2} & \textbf{68.4} & 66.9 & 69.8 & 69.2 & 67.4 & 69.8 & 67.4 & 69.0 & \textbf{68.5} \\
 & 2 & 66.1 & 64.9 & 70.4 & 70.8 & 67.8 & 66.6 & 69.8 & 69.2 & 67.5 & 69.9 & 67.5 & 68.9 & 68.3 \\
 & 3$_{s=5}$ & 66.6 & 65.6 & 70.3 & 70.6 & 67.6 & 66.9 & 69.9 & 69.2 & \textbf{67.5} & 70.0 & \textbf{67.6} & 69.0 & 68.4 \\
 & 3$_{s=10}$ & 66.6 & 65.5 & 70.3 & 70.9 & 67.7 & 66.9 & 70.0 & 69.1 & 67.4 & \textbf{70.1} & 67.5 & 69.0 & 68.4 \\
 & 3$_{s=20}$ & 66.7 & 65.3 & 70.6 & 70.8 & 67.8 & \textbf{66.9} & \textbf{70.1} & 69.2 & 67.5 & 70.1 & 67.6 & \textbf{69.1} & 68.5 \\
 & 4 & 66.3 & 64.7 & \textbf{70.7} & 70.9 & 67.9 & 66.0 & 69.5 & 69.1 & 67.4 & 69.6 & 67.1 & 68.6 & 68.1 \\
 & 5$_{s=5}$ & 66.6 & 64.8 & 70.7 & 70.7 & 67.9 & 66.1 & 69.6 & 69.1 & 67.5 & 69.9 & 67.4 & 68.8 & 68.2 \\
 & \underline{5$_{s=10}$} & 66.6 & 64.8 & 70.6 & 70.7 & 68.0 & 66.2 & 69.7 & 69.1 & 67.5 & 69.9 & 67.4 & 68.7 & 68.3 \\
 & 5$_{s=20}$ & 66.7 & 64.6 & 70.5 & 70.9 & 68.0 & 66.3 & 69.8 & \textbf{69.2} & 67.5 & 69.9 & 67.6 & 68.8 & 68.3 \\
\cmidrule(l){2-15}
 & \textbf{avg} & 66.5 & 65.1 & 70.5 & 70.8 & 67.9 & 66.5 & 69.8 & 69.2 & 67.5 & 69.9 & 67.5 & 68.9 & 68.3 \\
\pagebreak
\multirow{10}{*}{\rotatebox{90}{Llama3-8B}} & 1 & 24.0 & 33.9 & 40.7 & 37.0 & 34.6 & 30.9 & 53.6 & 63.0 & 54.0 & 52.8 & 54.1 & 54.6 & 44.4 \\
 & 2 & 23.9 & 33.7 & 41.1 & 37.3 & 34.9 & 31.3 & 54.3 & 62.9 & 54.9 & 53.4 & 55.4 & 54.9 & 44.8 \\
 & 3$_{s=5}$ & 23.9 & 34.4 & 42.1 & 37.6 & 35.4 & 31.4 & 55.7 & 62.1 & 53.8 & 53.1 & 54.2 & 54.5 & 44.9 \\
 & 3$_{s=10}$ & 24.9 & 35.7 & 43.1 & 39.4 & 36.2 & 32.3 & 55.8 & \textbf{63.6} & 56.6 & 54.6 & 56.7 & \textbf{56.9} & 46.3 \\
 & 3$_{s=20}$ & 25.0 & \textbf{36.5} & 43.7 & 40.1 & 37.7 & 32.7 & 55.6 & 63.6 & 56.6 & 54.9 & 56.1 & 56.4 & 46.6 \\
 & 4 & 25.0 & 34.2 & 41.7 & 37.5 & 37.0 & 31.2 & 55.8 & 62.5 & 53.7 & 52.7 & 55.3 & 52.7 & 44.9 \\
 & 5$_{s=5}$ & 25.1 & 34.8 & 43.0 & 38.3 & 38.3 & 32.1 & 56.2 & 61.1 & 53.6 & 53.5 & 55.0 & 54.3 & 45.5 \\
 & \underline{5$_{s=10}$} & \textbf{25.9} & 36.1 & 44.2 & 40.2 & 38.5 & 32.8 & \textbf{56.5} & 63.4 & \textbf{56.9} & 54.8 & \textbf{57.2} & 56.9 & 46.9 \\
 & 5$_{s=20}$ & 25.8 & 36.4 & \textbf{44.7} & \textbf{40.4} & \textbf{40.1} & \textbf{33.5} & 56.4 & 63.3 & 56.5 & \textbf{55.2} & 56.8 & 56.7 & \textbf{47.1} \\
\cmidrule(l){2-15}
 & \textbf{avg} & 24.8 & 35.1 & 42.7 & 38.6 & 37.0 & 32.0 & 55.6 & 62.8 & 55.2 & 53.9 & 55.6 & 55.3 & 45.7 \\
\midrule
\multirow{10}{*}{\rotatebox{90}{Gemma2-9B}} & 1 & 32.3 & 48.5 & 35.9 & 41.0 & 33.7 & 36.6 & 62.4 & 64.9 & 57.7 & 58.8 & 56.4 & 62.7 & 49.3 \\
 & 2 & 32.0 & 47.8 & 35.9 & 41.0 & 33.2 & 36.5 & 62.3 & 64.9 & 58.9 & 59.7 & 56.9 & 63.1 & 49.4 \\
 & 3$_{s=5}$ & 31.4 & 49.9 & 36.0 & 37.0 & 29.1 & 38.1 & 63.5 & 66.1 & 60.5 & 62.1 & 58.9 & 64.3 & 49.7 \\
 & 3$_{s=10}$ & 32.2 & 50.4 & 36.7 & 39.1 & 30.7 & 39.3 & 63.8 & 66.2 & 60.9 & 62.6 & 59.5 & 65.0 & 50.5 \\
 & 3$_{s=20}$ & 32.7 & \textbf{50.7} & 37.3 & 38.8 & 29.9 & 39.7 & 63.8 & 66.1 & 61.0 & 62.6 & 59.6 & 65.1 & 50.6 \\
 & 4 & 32.6 & 48.5 & 36.6 & \textbf{41.4} & \textbf{37.1} & 37.3 & 63.3 & 65.3 & 59.6 & 60.7 & 59.2 & 63.5 & 50.4 \\
 & 5$_{s=5}$ & 32.7 & 50.1 & 37.1 & 38.0 & 34.9 & 39.1 & 63.8 & 66.2 & 60.7 & 62.4 & 59.9 & 64.5 & 50.8 \\
 & \underline{5$_{s=10}$} & 33.7 & 50.4 & 37.5 & 40.4 & 35.1 & 40.0 & 64.1 & \textbf{66.3} & \textbf{61.3} & \textbf{63.0} & 60.4 & \textbf{65.1} & 51.4 \\
 & 5$_{s=20}$ & \textbf{34.1} & 50.4 & \textbf{38.5} & 40.3 & 34.4 & \textbf{40.6} & \textbf{64.2} & 66.2 & 61.1 & 62.9 & \textbf{60.4} & 65.1 & \textbf{51.5} \\
\cmidrule(l){2-15}
 & \textbf{avg} & 32.6 & 49.7 & 36.8 & 39.7 & 33.1 & 38.6 & 63.5 & 65.8 & 60.2 & 61.7 & 59.0 & 64.3 & 50.4 \\
\end{longtable}
\endgroup
\twocolumn

\begin{table*}[t]
    \begin{subtable}{\textwidth}
        \centering
        \resizebox{\textwidth}{!}{
        \begin{tabular}{lrrrrrrr}
        \toprule
            \multirow{2}{*}{\textbf{domain}} & \multirow{2}{*}{\textbf{\#sents}} & \multicolumn{6}{c}{\textbf{domain-level COMET}} \\
            \cmidrule(lr){3-8}
            & & \texttt{\textbf{amh}} & \texttt{\textbf{hau}} & \texttt{\textbf{lug}} & \texttt{\textbf{nso}} & \texttt{\textbf{yor}} & \texttt{\textbf{zul}} \\
        \midrule
            \textit{all-domains} & 5{,}792 & 65.5 (\textcolor{red}{-0.6}) & 62.5 (\textcolor{red}{-0.8}) & 69.5 (\textcolor{red}{-0.4}) & 70.1 (\textcolor{red}{-0.6}) & 66.4 (\textcolor{red}{-0.9}) & 65.4 (\textcolor{red}{-0.6}) \\
        \midrule
            chemistry & 46 & 63.9 (\textcolor{blue}{+0.2}) & 65.7 (\textcolor{blue}{+1.5}) & 69.5 (\textcolor{red}{-0.4}) & 72.8 (\textcolor{red}{-0.4}) & 70.3 (\textcolor{red}{-0.8}) & 66.3 (\textcolor{red}{-1.0}) \\
            Indigenous knowledge & 92 & 66.8 (\textcolor{red}{-1.0}) & 65.7 (\textcolor{red}{-1.1}) & 69.9 (\textcolor{red}{-0.4}) & 71.9 (\textcolor{red}{-0.2}) & 67.5 (\textcolor{red}{-0.3}) & 66.9 (\textcolor{red}{-0.3}) \\
            engineering & 93 & 63.7 (\textcolor{red}{-0.5}) & 61.2 (\textcolor{red}{-1.4}) & 65.9 (\textcolor{red}{-0.7}) & 67.2 (\textcolor{red}{-0.5}) & 65.5 (\textcolor{blue}{+1.6}) & 63.9 (\textcolor{red}{-1.0}) \\
            statistics & 184 & 65.3 (\textcolor{blue}{+0.5}) & 59.6 (\textcolor{red}{-0.6}) & 70.6 (\textcolor{red}{-1.3}) & 70.4 (\textcolor{blue}{+0.1}) & 69.5 (\textcolor{blue}{+0.9}) & 65.8 (\textcolor{blue}{+0.4}) \\
            computer science & 213 & 65.3 (\textcolor{red}{-1.3}) & 67.1 (\textcolor{blue}{+0.2}) & 72.1 (\textcolor{blue}{+0.4}) & 71.9 (\textcolor{red}{-1.2}) & 66.2 (\textcolor{red}{-0.5}) & 64.6 (\textcolor{blue}{+0.4}) \\
            geography & 218 & 69.1 (\textcolor{red}{-0.1}) & 63.9 (\textcolor{red}{-0.7}) & 70.7 (\textcolor{red}{-0.2}) & 71.1 (\textcolor{red}{-0.2}) & 68.4 (\textcolor{red}{-0.1}) & 67.5 (\textcolor{blue}{+0.0}) \\
            agriculture & 388 & 68.0 (\textcolor{red}{-0.9}) & 65.0 (\textcolor{red}{-0.6}) & 69.9 (\textcolor{blue}{+0.4}) & 70.2 (\textcolor{red}{-1.6}) & 65.7 (\textcolor{red}{-0.9}) & 66.8 (\textcolor{red}{-1.4}) \\
            biology & 632 & 64.7 (\textcolor{red}{-0.3}) & 60.7 (\textcolor{red}{-1.7}) & 69.1 (\textcolor{red}{-0.8}) & 69.1 (\textcolor{red}{-0.3}) & 65.5 (\textcolor{red}{-1.5}) & 65.0 (\textcolor{blue}{+0.0}) \\
            biochemistry & 651 & 67.0 (\textcolor{red}{-0.2}) & 63.6 (\textcolor{red}{-0.8}) & 71.1 (\textcolor{red}{-0.0}) & 71.3 (\textcolor{red}{-0.4}) & 67.6 (\textcolor{red}{-0.9}) & 65.4 (\textcolor{red}{-1.0}) \\
            sociology & 670 & 66.3 (\textcolor{red}{-0.4}) & 63.5 (\textcolor{red}{-1.7}) & 70.4 (\textcolor{red}{-0.5}) & 71.1 (\textcolor{red}{-0.6}) & 66.8 (\textcolor{red}{-1.3}) & 65.9 (\textcolor{red}{-0.8}) \\
            health & 2605 & 63.7 (\textcolor{red}{-0.9}) & 60.7 (\textcolor{red}{-0.6}) & 68.5 (\textcolor{red}{-0.6}) & 69.0 (\textcolor{red}{-0.6}) & 64.9 (\textcolor{red}{-1.5}) & 64.3 (\textcolor{red}{-0.7}) \\
        \bottomrule
        \end{tabular}
        }
        \caption{COMET}
    \end{subtable}
    
    \begin{subtable}{\textwidth}
        \centering
        \resizebox{\textwidth}{!}{
        \begin{tabular}{lrrrrrrr}
        \toprule
            \multirow{2}{*}{\textbf{domain}} & \multirow{2}{*}{\textbf{\#sents}} & \multicolumn{6}{c}{\textbf{domain-level ChrF}} \\
            \cmidrule(lr){3-8}
            & & \texttt{\textbf{amh}} & \texttt{\textbf{hau}} & \texttt{\textbf{lug}} & \texttt{\textbf{nso}} & \texttt{\textbf{yor}} & \texttt{\textbf{zul}} \\
        \midrule
            \textit{all-domains} & 5{,}792 & 43.7 (\textcolor{red}{-1.8}) & 66.3 (\textcolor{red}{-1.4}) & 50.5 (\textcolor{red}{-1.0}) & 63.4 (\textcolor{red}{-0.9}) & 17.4 (\textcolor{red}{-0.3}) & 65.0 (\textcolor{red}{-2.2}) \\
        \midrule
            chemistry & 46 & 37.9 (\textcolor{red}{-0.6}) & 71.2 (\textcolor{red}{-0.3}) & 50.7 (\textcolor{red}{-1.5}) & 69.1 (\textcolor{red}{-1.4}) & 18.6 (\textcolor{blue}{+0.4}) & 69.1 (\textcolor{red}{-4.0}) \\
            Indigenous knowledge & 92 & 40.2 (\textcolor{red}{-0.7}) & 65.0 (\textcolor{red}{-2.8}) & 54.5 (\textcolor{blue}{+0.0}) & 63.6 (\textcolor{red}{-0.9}) & 17.7 (\textcolor{blue}{+0.2}) & 67.3 (\textcolor{red}{-0.5}) \\
            engineering & 93 & 43.7 (\textcolor{red}{-1.4}) & 63.3 (\textcolor{red}{-2.1}) & 45.9 (\textcolor{red}{-1.0}) & 61.8 (\textcolor{blue}{+0.1}) & 12.3 (\textcolor{red}{-0.3}) & 71.3 (\textcolor{red}{-2.9}) \\
            statistics & 184 & 44.1 (\textcolor{red}{-2.7}) & 64.9 (\textcolor{red}{-0.2}) & 54.8 (\textcolor{red}{-2.5}) & 66.8 (\textcolor{blue}{+1.0}) & 18.9 (\textcolor{blue}{+0.2}) & 52.0 (\textcolor{blue}{+0.6}) \\
            computer science & 213 & 42.7 (\textcolor{red}{-1.7}) & 72.7 (\textcolor{blue}{+0.3}) & 55.4 (\textcolor{blue}{+0.7}) & 70.5 (\textcolor{blue}{+0.5}) & 16.8 (\textcolor{red}{-0.3}) & 49.5 (\textcolor{red}{-0.9}) \\
            geography & 218 & 51.2 (\textcolor{red}{-2.1}) & 69.8 (\textcolor{red}{-1.3}) & 47.4 (\textcolor{red}{-1.2}) & 63.4 (\textcolor{red}{-1.5}) & 13.1 (\textcolor{red}{-0.1}) & 69.8 (\textcolor{red}{-0.5}) \\
            agriculture & 388 & 48.6 (\textcolor{red}{-2.6}) & 70.8 (\textcolor{red}{-2.5}) & 48.6 (\textcolor{blue}{+0.8}) & 64.4 (\textcolor{red}{-1.5}) & 13.3 (\textcolor{red}{-0.2}) & 68.4 (\textcolor{red}{-3.6}) \\
            biology & 632 & 43.6 (\textcolor{red}{-1.9}) & 65.8 (\textcolor{red}{-1.6}) & 51.0 (\textcolor{red}{-0.9}) & 61.5 (\textcolor{red}{-0.0}) & 26.6 (\textcolor{red}{-0.4}) & 63.5 (\textcolor{red}{-2.8}) \\
            biochemistry & 651 & 45.1 (\textcolor{red}{-2.5}) & 67.4 (\textcolor{red}{-1.9}) & 56.5 (\textcolor{red}{-0.8}) & 66.7 (\textcolor{red}{-2.0}) & 16.4 (\textcolor{red}{-0.1}) & 69.0 (\textcolor{red}{-2.4}) \\
            sociology & 670 & 46.6 (\textcolor{red}{-1.4}) & 68.2 (\textcolor{red}{-1.8}) & 52.6 (\textcolor{red}{-1.1}) & 64.2 (\textcolor{red}{-1.0}) & 14.4 (\textcolor{red}{-0.4}) & 69.6 (\textcolor{red}{-1.6}) \\
            health & 2605 & 41.2 (\textcolor{red}{-1.8}) & 63.3 (\textcolor{red}{-0.9}) & 48.3 (\textcolor{red}{-1.4}) & 61.3 (\textcolor{red}{-0.9}) & 18.0 (\textcolor{red}{-0.6}) & 61.7 (\textcolor{red}{-2.4}) \\
        \bottomrule
        \end{tabular}
        }
        \caption{ChrF}
    \end{subtable}
    \caption{Domain-level performance of NLLB-1.3B on \texttt{eng}$\to$\texttt{xxx} translation under uniform temperature sampling ($\tau{=}\infty$). Numbers in parentheses are the change relative to the natural-distribution baseline ($\tau{=}1$); blue = improvement, red = decrease. Total training volume is held at 5{,}792 sentences across both conditions; only the per-example sampling weight changes.}
    \label{tab:balance_domain_eng_xxx_dual}
\end{table*}


{%
  \setlength{\intextsep}{0pt}%
  \setlength{\textfloatsep}{0pt}%
  \setlength{\abovecaptionskip}{1pt}%
  \setlength{\belowcaptionskip}{0pt}%
  \setlength{\parskip}{0pt}%
  \captionsetup[subfigure]{font=footnotesize, labelfont=footnotesize, skip=0pt, belowskip=0pt, aboveskip=1pt}%
  \captionsetup[figure]{font=small, labelfont=small, skip=2pt, belowskip=0pt}%

  \begin{figure*}[h!]
    \centering

    \begin{subfigure}{\textwidth}
      \centering
      \spGridCOMET
      \caption{COMET}
      \label{fig:spider_per_pair_comet}
    \end{subfigure}

    \vspace{6pt}

    \begin{subfigure}{\textwidth}
      \centering
      \spGridChrF
      \caption{ChrF}
      \label{fig:spider_per_pair_chrf}
    \end{subfigure}

    \par\vspace{4pt}
    {\footnotesize
    \textcolor{spcNLLB}{\textbf{---}} NLLB-1.3B \enspace
    \textcolor{spcGemma}{\textbf{- - -}} Gemma2-9B (LoRA) \enspace
    \textcolor{spcGPT}{\textbf{$\cdot\cdot\cdot$}} GPT-5.4 (10-shot ICL)
    }\par\vspace{6pt}

    \caption{Per-domain performance for the 12 directional language pairs, grouped by language. Each pair compares the best system from each of the three model families: NLLB-1.3B (fine-tuned), Gemma2-9B-IT (LoRA $r{=}64$), and GPT-5.4 (\texttt{template5\textunderscore{}10shot}, $T{=}0$) across the 11 scientific domains.}
    \label{fig:spider_per_pair}
  \end{figure*}
}

\begin{figure*}
    \centering
    \includegraphics[width=\textwidth]{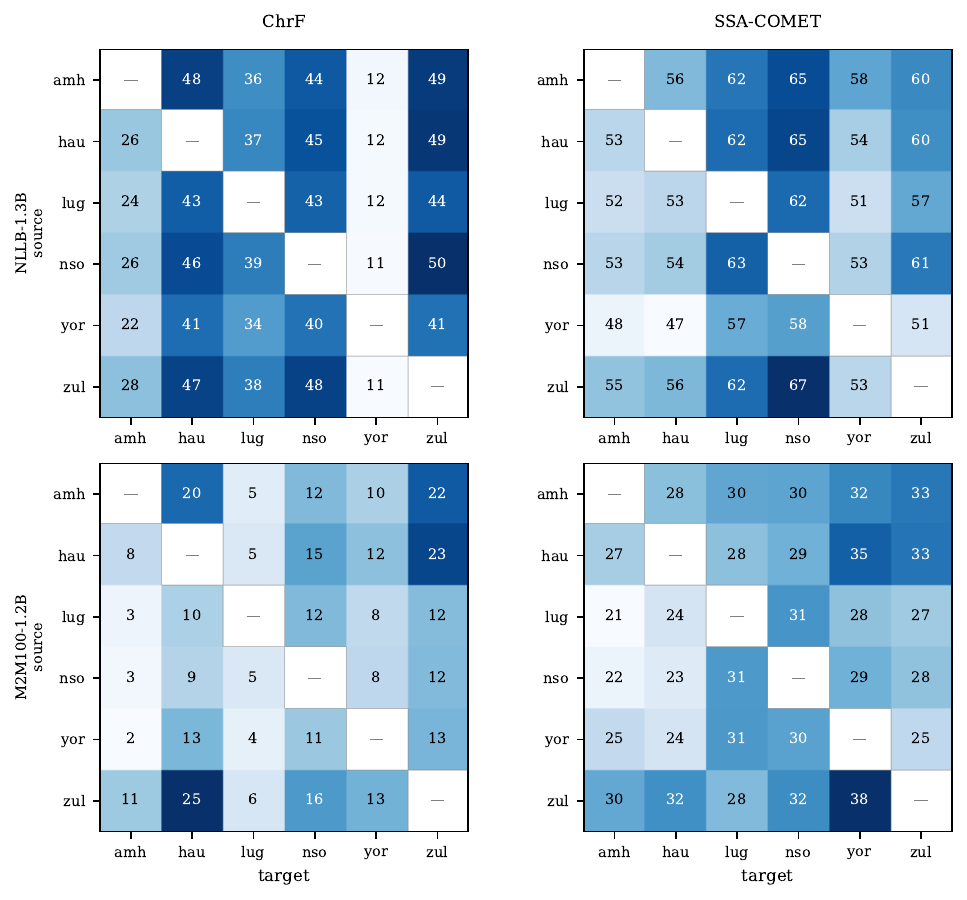}
    \caption{Zero-shot African$\leftrightarrow$African translation. Each cell is one source$\to$target pair (rows = source, columns = target; diagonals omitted). Top row: NLLB-1.3B; bottom row: M2M100-1.2B. Left column: ChrF; right column: COMET.}
    \label{fig:n2n_zero_shot}
\end{figure*}

\begin{table*}[t]
\centering
\small
\renewcommand{\arraystretch}{1.05}
\begin{tabular}{llrrrrrrrrrrrr}
\toprule
\textbf{Setup} & \textbf{Model} & \multicolumn{6}{c}{\textbf{eng$\to$}} & \multicolumn{6}{c}{\textbf{$\to$eng}} \\
\cmidrule(lr){3-8}\cmidrule(lr){9-14}
  &   & \texttt{amh} & \texttt{hau} & \texttt{lug} & \texttt{nso} & \texttt{yor} & \texttt{zul} & \texttt{amh} & \texttt{hau} & \texttt{lug} & \texttt{nso} & \texttt{yor} & \texttt{zul} \\
\midrule
\multirow{6}{*}{\textit{LoRA $r{=}4$}} & AfriqueLlama-8B & \emph{58.3} & \emph{59.9} & 56.3 & \emph{60.1} & \emph{57.2} & 54.5 & \textbf{67.6} & \emph{65.7} & 53.5 & 59.8 & 55.8 & 57.0 \\
 & AfriqueQwen-8B & 43.7 & 40.5 & 53.5 & 57.3 & 49.2 & 34.5 & 51.0 & 53.1 & 32.0 & 36.3 & 39.2 & 12.3 \\
 & Gemma2-9B & 48.9 & 49.8 & \emph{57.5} & 59.1 & 55.7 & 46.9 & 60.2 & 60.7 & \emph{58.0} & \emph{60.8} & \emph{56.3} & 60.4 \\
 & Llama3-8B & 29.1 & 47.4 & 54.6 & 51.0 & 50.6 & 40.3 & 56.2 & 59.7 & 56.9 & 58.7 & 54.7 & 57.4 \\
 & Tiny-Aya-E & 53.8 & 54.8 & 51.6 & 50.5 & 55.1 & \emph{55.5} & 60.8 & 60.9 & 54.7 & 56.6 & 50.4 & \emph{60.7} \\
 & Tiny-Aya-G & 53.4 & 54.5 & 51.8 & 51.2 & 55.3 & 55.3 & 60.5 & 60.8 & 55.0 & 56.4 & 50.1 & 60.6 \\
\midrule
\multirow{6}{*}{\textit{LoRA $r{=}64$}} & AfriqueLlama-8B & 49.6 & 51.5 & 58.7 & 58.7 & 55.5 & 51.6 & 60.9 & 53.5 & 54.7 & 58.7 & 55.7 & 56.4 \\
 & AfriqueQwen-8B & 23.8 & 37.2 & 45.1 & 52.1 & 21.2 & 28.7 & 36.6 & 28.2 & 22.8 & 25.8 & 41.3 & 26.5 \\
 & Gemma2-9B & 54.0 & 50.5 & 62.7 & 62.8 & 58.9 & 48.4 & 60.4 & 60.6 & 58.4 & 61.1 & 56.3 & 60.4 \\
 & Llama3-8B & 42.4 & 51.3 & 61.1 & 60.6 & 56.8 & 50.3 & 58.1 & 60.2 & 57.7 & 60.4 & 55.6 & 59.0 \\
 & Tiny-Aya-E & 54.3 & 54.3 & 59.8 & 59.2 & 57.1 & 55.7 & 57.8 & 61.0 & 56.5 & 58.7 & 50.2 & 60.5 \\
 & Tiny-Aya-G & \textbf{61.9} & \textbf{61.6} & \textbf{65.7} & \textbf{65.3} & \textbf{65.2} & \textbf{63.1} & \emph{67.3} & \textbf{67.2} & \textbf{64.3} & \textbf{66.4} & \textbf{63.8} & \textbf{67.2} \\
\bottomrule
\end{tabular}
\caption{LoRA fine-tuning of six open-source LLMs at two ranks (COMET). Each cell is one (language, direction) pair; each model is fine-tuned bilingually per pair for 10 epochs with early stopping. Comparing $r{=}4$ (low-rank, parameter-efficient) against $r{=}64$ (the headline rank used elsewhere in the paper) isolates the effect of adapter capacity. \textbf{Bold}: best score for that (language, direction) pair across both ranks; \emph{italic}: best score for that pair within a single rank block.}
\label{tab:lora_comet}
\end{table*}

\begin{table*}[t]
\centering
\setlength{\tabcolsep}{4pt}
\renewcommand{\arraystretch}{1.05}
\begin{tabular}{lrrrrrrrrrrrrrr}
\toprule
\textbf{Setup} & \multicolumn{7}{c}{\textbf{chrF}} & \multicolumn{7}{c}{\textbf{COMET}} \\
\cmidrule(lr){2-8}\cmidrule(lr){9-15}
 & \textbf{\texttt{amh}} & \textbf{\texttt{hau}} & \textbf{\texttt{lug}} & \textbf{\texttt{nso}} & \textbf{\texttt{yor}} & \textbf{\texttt{zul}} & \textbf{Avg} & \textbf{\texttt{amh}} & \textbf{\texttt{hau}} & \textbf{\texttt{lug}} & \textbf{\texttt{nso}} & \textbf{\texttt{yor}} & \textbf{\texttt{zul}} & \textbf{Avg} \\
\midrule
\multicolumn{15}{l}{\textbf{NLLB-600M / 1.3B}} \\
\cmidrule(l){1-15}
\multicolumn{15}{l}{\textit{\,Bilingual fine-tune}} \\
\quad NLLB-1.3B & \textbf{45.5} & \textbf{67.7} & \textbf{51.5} & \textbf{64.3} & 17.7 & \textbf{67.2} & \textbf{52.3} & \textbf{66.0} & \textbf{63.3} & \textbf{70.0} & \textbf{70.7} & \textbf{67.3} & \textbf{66.0} & \textbf{67.2} \\
\quad NLLB-600M & 44.4 & 66.9 & 50.6 & 64.2 & 17.6 & 66.3 & 51.7 & 65.1 & 62.9 & 69.5 & 70.3 & 66.8 & 65.7 & 66.7 \\
\cmidrule(l){1-15}
\multicolumn{15}{l}{\textit{\,Multilingual fine-tune (NLLB-600M, training updates)}} \\
\quad @ 40k & 8.1 & 24.6 & 22.1 & 24.0 & 8.9 & 17.6 & 17.6 & 23.4 & 25.7 & 39.0 & 34.9 & 28.2 & 28.5 & 30.0 \\
\quad @ 80k & 10.7 & 30.0 & 24.9 & 30.7 & 10.6 & 22.2 & 21.5 & 24.9 & 28.3 & 40.3 & 37.8 & 33.3 & 30.8 & 32.5 \\
\quad @ 200k & 13.5 & 36.8 & 29.3 & 37.6 & 12.1 & 28.3 & 26.3 & 27.0 & 32.3 & 42.6 & 41.6 & 39.5 & 34.0 & 36.2 \\
\quad @ 400k & 16.4 & 42.4 & 33.9 & 42.7 & 13.1 & 32.9 & 30.2 & 30.3 & 35.6 & 46.2 & 45.1 & 44.6 & 37.0 & 39.8 \\
\midrule
\multicolumn{15}{l}{\textbf{Llama-3.1-8B-Instruct (same model both sides, LoRA rank sweep)}} \\
\cmidrule(l){1-15}
\multicolumn{15}{l}{\textit{\,Bilingual LoRA}} \\
\quad $r{=}8$ & 22.3 & 60.0 & 46.2 & 56.5 & 35.7 & 50.4 & 45.2 & 38.7 & 58.2 & 64.3 & 64.0 & 59.5 & 56.4 & 56.9 \\
\quad $r{=}16$ & 21.9 & 59.4 & 48.3 & 58.3 & 36.0 & 52.2 & 46.0 & 38.0 & 57.9 & 65.9 & 65.2 & 58.7 & 57.9 & 57.3 \\
\quad $r{=}32$ & 22.2 & 59.9 & 47.7 & 58.1 & 35.8 & 52.6 & 46.1 & 38.0 & 58.5 & 65.1 & 65.7 & 58.6 & 58.2 & 57.4 \\
\quad $r{=}64$ & 32.9 & 60.3 & 49.3 & 59.0 & \textbf{39.9} & 52.4 & 49.0 & 42.4 & 51.3 & 61.1 & 60.6 & 56.8 & 50.3 & 53.8 \\
\quad $r{=}128$ & 22.9 & 60.4 & 47.7 & 57.7 & 35.5 & 52.8 & 46.2 & 38.7 & 58.3 & 65.9 & 65.2 & 59.2 & 58.7 & 57.7 \\
\cmidrule(l){1-15}
\multicolumn{15}{l}{\textit{\,Multilingual LoRA}} \\
\quad $r{=}8$ & 22.2 & 58.9 & 46.8 & 57.6 & 35.4 & 49.6 & 45.1 & 38.3 & 57.1 & 65.0 & 64.1 & 60.5 & 56.0 & 56.8 \\
\quad $r{=}16$ & 22.4 & 59.9 & 48.1 & 57.9 & 34.7 & 50.7 & 45.6 & 38.0 & 57.6 & 65.8 & 64.9 & 61.5 & 56.5 & 57.4 \\
\quad $r{=}32$ & 21.7 & 59.2 & 46.7 & 57.9 & 35.6 & 50.3 & 45.2 & 38.1 & 57.1 & 64.2 & 65.5 & 61.3 & 56.9 & 57.2 \\
\quad $r{=}64$ & 21.1 & 59.0 & 47.1 & 57.1 & 35.5 & 50.9 & 45.1 & 37.1 & 57.4 & 64.9 & 64.9 & 60.7 & 57.3 & 57.0 \\
\quad $r{=}128$ & 21.8 & 58.6 & 47.0 & 57.6 & 35.1 & 51.2 & 45.2 & 38.1 & 56.6 & 64.5 & 65.2 & 61.3 & 57.5 & 57.2 \\
\midrule
\multicolumn{15}{l}{\textbf{Gemma-2-9B non-instruct (same base both sides, LoRA rank sweep)}} \\
\cmidrule(l){1-15}
\multicolumn{15}{l}{\textit{\,Bilingual LoRA}} \\
\quad $r{=}8$ & 35.1 & 60.2 & 46.3 & 57.3 & 37.8 & 50.9 & 47.9 & 56.0 & 58.1 & 64.9 & 64.6 & 61.9 & 57.9 & 60.6 \\
\quad $r{=}16$ & 34.8 & 60.5 & 47.2 & 57.8 & 37.1 & 50.8 & 48.0 & 55.8 & 58.3 & 65.5 & 65.1 & 62.0 & 57.6 & 60.7 \\
\quad $r{=}32$ & 35.1 & 60.4 & 47.3 & 57.9 & 37.1 & 51.1 & 48.1 & 56.1 & 58.0 & 65.2 & 65.5 & 62.4 & 57.5 & 60.8 \\
\quad $r{=}64$ & 35.6 & 60.7 & 47.9 & 58.7 & 36.6 & 52.3 & 48.6 & 57.0 & 58.1 & 65.9 & 65.6 & 62.7 & 57.8 & 61.2 \\
\quad $r{=}128$ & 37.8 & 62.8 & 49.4 & 60.5 & 36.8 & 55.1 & 50.4 & 57.9 & 59.7 & 66.7 & 67.0 & 63.0 & 60.5 & 62.5 \\
\cmidrule(l){1-15}
\multicolumn{15}{l}{\textit{\,Multilingual LoRA}} \\
\quad $r{=}8$ & 31.1 & 57.5 & 41.3 & 54.6 & 35.0 & 44.8 & 44.0 & 52.1 & 57.8 & 60.3 & 63.4 & 56.9 & 53.3 & 57.3 \\
\quad $r{=}16$ & 32.0 & 57.7 & 42.8 & 55.6 & 35.9 & 46.0 & 45.0 & 53.3 & 57.8 & 62.3 & 64.2 & 58.0 & 54.9 & 58.4 \\
\quad $r{=}32$ & 32.9 & 58.2 & 42.4 & 56.4 & 35.9 & 47.3 & 45.5 & 54.1 & 57.7 & 61.7 & 64.7 & 59.4 & 55.3 & 58.8 \\
\quad $r{=}64$ & 31.8 & 56.8 & 43.6 & 56.3 & 35.5 & 48.2 & 45.4 & 53.2 & 57.0 & 62.8 & 64.0 & 58.8 & 56.8 & 58.8 \\
\quad $r{=}128$ & 30.2 & 55.7 & 43.1 & 54.0 & 35.0 & 46.5 & 44.1 & 50.3 & 55.8 & 61.6 & 62.2 & 57.1 & 54.6 & 57.0 \\
\bottomrule
\end{tabular}
\caption{Bilingual vs. multilingual fine-tuning on \texttt{eng}$\to$\texttt{xxx} (chrF and COMET side-by-side). Three model families: NLLB-600M (multilingual checkpoint sweep against NLLB-600M and 1.3B bilingual fine-tunes); Llama-3.1-8B-Instruct (same instruct model on both sides); Gemma-2-9B non-instruct base. For NLLB, bilingual fine-tuning is substantially stronger than the multilingual checkpoints. For Gemma LoRA, bilingual leads by $\sim$3.7 COMET / $\sim$5 chrF at the best rank. For Llama LoRA, bilingual and multilingual end up essentially tied at COMET ($\sim$57). \textbf{Bold}: best score for that target language (or Avg) within each metric, across all rows of the table.}
\label{tab:multilingual_vs_bilingual}
\end{table*}

\begin{table*}[t]
\centering
\footnotesize
\setlength{\tabcolsep}{4pt}
\renewcommand{\arraystretch}{1.05}
\resizebox{\textwidth}{!}{
\begin{tabular}{llrrrrrrrrrrrrr}
\toprule
\textbf{Model} & \textbf{Config} & \multicolumn{6}{c}{\textbf{eng$\to$}} & \multicolumn{6}{c}{\textbf{$\to$eng}} & \\
\cmidrule(lr){3-8}\cmidrule(lr){9-14}
  &   & \textbf{\texttt{amh}} & \textbf{\texttt{hau}} & \textbf{\texttt{lug}} & \textbf{\texttt{nso}} & \textbf{\texttt{yor}} & \textbf{\texttt{zul}} & \textbf{\texttt{amh}} & \textbf{\texttt{hau}} & \textbf{\texttt{lug}} & \textbf{\texttt{nso}} & \textbf{\texttt{yor}} & \textbf{\texttt{zul}} & \textbf{Avg} \\
\midrule
\multirow{7}{*}{\rotatebox{90}{Gemini-1.5}} & 1$_{s=0}$ & 38.7 & \textbf{33.8} & 48.6 & 45.5 & 44.5 & 40.1 & 49.7 & 49.7 & 51.1 & \textbf{51.7} & \textbf{52.0} & \textbf{50.0} & 46.3 \\
 & 1$_{s=1}$ & 38.7 & 32.5 & 49.9 & \textbf{47.3} & 44.3 & 39.1 & 48.7 & 48.9 & 50.0 & 49.2 & 50.3 & 49.8 & 45.7 \\
 & 2$_{s=0}$ & 39.1 & 33.8 & 49.3 & 45.0 & 44.8 & 40.7 & 49.7 & \textbf{49.9} & \textbf{51.5} & 50.7 & 51.6 & 49.5 & 46.3 \\
 & 2$_{s=1}$ & \textbf{39.3} & 32.4 & 49.0 & 45.5 & 44.5 & 39.9 & 49.3 & 49.4 & 49.0 & 49.9 & 50.2 & 49.2 & 45.6 \\
 & 3$_{s=0}$ & 39.2 & 33.4 & 49.5 & 47.3 & \textbf{46.2} & \textbf{41.4} & \textbf{50.0} & 49.2 & 51.3 & 50.4 & 51.5 & 49.8 & \textbf{46.6} \\
 & 3$_{s=1}$ & 38.7 & 31.7 & \textbf{50.6} & 47.3 & 43.1 & 40.5 & 49.3 & 49.0 & 49.9 & 49.6 & 50.2 & 49.3 & 45.8 \\
\cmidrule(l){2-15}
 & \textbf{avg} & 38.9 & 32.9 & 49.5 & 46.3 & 44.6 & 40.3 & 49.4 & 49.4 & 50.5 & 50.2 & 51.0 & 49.6 & 46.1 \\
\midrule
\multirow{7}{*}{\rotatebox{90}{GPT-4o}} & 1$_{s=0}$ & 32.1 & \textbf{33.5} & 46.0 & 46.0 & 39.8 & 40.2 & 48.8 & \textbf{48.9} & \textbf{50.2} & 49.3 & 49.2 & 49.4 & 44.4 \\
 & 1$_{s=1}$ & \textbf{33.9} & 33.4 & \textbf{49.6} & 46.8 & 38.7 & \textbf{41.1} & 48.5 & 48.8 & 49.9 & 49.1 & 48.6 & 48.8 & \textbf{44.8} \\
 & 2$_{s=0}$ & 32.8 & 32.6 & 46.0 & 46.3 & 40.4 & 39.8 & \textbf{49.4} & 48.9 & 49.8 & 49.4 & 49.3 & 49.3 & 44.5 \\
 & 2$_{s=1}$ & 33.0 & 33.4 & 49.5 & \textbf{46.9} & \textbf{40.9} & 40.8 & 47.6 & 48.5 & 49.7 & 48.8 & 47.9 & 49.1 & 44.7 \\
 & 3$_{s=0}$ & 32.9 & 32.1 & 44.4 & 43.0 & 39.7 & 39.2 & 49.4 & 48.8 & 50.1 & \textbf{49.7} & \textbf{49.9} & \textbf{49.6} & 44.1 \\
 & 3$_{s=1}$ & 30.9 & 30.4 & 45.9 & 40.6 & 35.7 & 37.7 & 47.9 & 47.9 & 48.1 & 46.7 & 46.0 & 49.2 & 42.3 \\
\cmidrule(l){2-15}
 & \textbf{avg} & 32.6 & 32.6 & 46.9 & 44.9 & 39.2 & 39.8 & 48.6 & 48.6 & 49.6 & 48.8 & 48.5 & 49.2 & 44.1 \\
\midrule
\multirow{7}{*}{\rotatebox{90}{Gemini-3.1}} & 1$_{s=0}$ & 41.5 & 38.9 & 53.4 & 48.1 & 48.2 & \textbf{44.0} & 50.5 & 50.7 & \textbf{52.1} & 51.0 & 51.0 & \textbf{50.6} & 48.3 \\
 & 1$_{s=1}$ & 41.6 & 37.2 & 51.3 & \textbf{48.9} & 44.9 & 43.3 & 50.5 & 50.1 & 51.2 & 50.3 & 50.9 & 50.1 & 47.5 \\
 & 2$_{s=0}$ & \textbf{42.3} & \textbf{39.6} & \textbf{54.0} & 48.2 & \textbf{48.9} & 43.8 & \textbf{50.9} & \textbf{51.2} & 52.0 & \textbf{51.8} & 50.9 & 50.6 & \textbf{48.7} \\
 & 2$_{s=1}$ & 42.1 & 37.2 & 53.2 & 48.1 & 45.4 & 42.9 & 49.6 & 50.3 & 50.9 & 50.2 & 50.7 & 50.1 & 47.6 \\
 & 3$_{s=0}$ & 41.3 & 38.2 & 52.5 & 48.3 & 48.5 & 43.3 & 49.9 & 50.0 & 51.4 & 50.9 & \textbf{51.4} & 50.4 & 48.0 \\
 & 3$_{s=1}$ & 41.5 & 35.6 & 51.6 & 47.9 & 47.2 & 43.6 & 49.7 & 50.1 & 50.4 & 50.4 & 50.5 & 50.5 & 47.4 \\
\cmidrule(l){2-15}
 & \textbf{avg} & 41.7 & 37.8 & 52.7 & 48.2 & 47.2 & 43.5 & 50.2 & 50.4 & 51.3 & 50.8 & 50.9 & 50.4 & 47.9 \\
\midrule
\multirow{7}{*}{\rotatebox{90}{GPT-5.4}} & 1$_{s=0}$ & 39.5 & 36.7 & \textbf{53.2} & 49.3 & 46.5 & 43.9 & 50.8 & 50.1 & 50.6 & \textbf{50.9} & 50.9 & 50.1 & 47.7 \\
 & 1$_{s=1}$ & 39.7 & 37.1 & 52.0 & 48.8 & \textbf{47.7} & 43.3 & 50.5 & \textbf{50.3} & 50.3 & 50.0 & 50.5 & 50.0 & 47.5 \\
 & 2$_{s=0}$ & 39.4 & 36.5 & 53.1 & 49.1 & 47.5 & 42.6 & \textbf{51.8} & 50.1 & \textbf{50.9} & 50.3 & \textbf{51.4} & 49.9 & \textbf{47.7} \\
 & 2$_{s=1}$ & \textbf{40.2} & 37.2 & 52.5 & 49.1 & 46.7 & 42.4 & 50.2 & 49.7 & 50.3 & 50.3 & 50.7 & 49.7 & 47.4 \\
 & 3$_{s=0}$ & 39.1 & 37.1 & 51.8 & \textbf{49.6} & 46.6 & 42.2 & 51.6 & 49.6 & 50.5 & 50.1 & 51.2 & 50.1 & 47.5 \\
 & 3$_{s=1}$ & 38.8 & \textbf{37.9} & 52.8 & 48.4 & 46.9 & \textbf{44.1} & 50.3 & 49.1 & 50.8 & 50.5 & 50.2 & \textbf{50.5} & 47.5 \\
\cmidrule(l){2-15}
 & \textbf{avg} & 39.4 & 37.1 & 52.6 & 49.1 & 47.0 & 43.1 & 50.9 & 49.8 & 50.6 & 50.4 & 50.8 & 50.1 & 47.6 \\
\bottomrule
\end{tabular}
}
\caption{Document-level template/shot ablation for the four closed models (older: GPT-4o, Gemini-1.5-Flash; newer: GPT-5.4, Gemini-3.1-Flash-Lite) across the 3-template $\times$ 2-shot grid (COMET). Each row is a (template, shot-count) configuration for that model (template number first, shot count as subscript $s$). Open LLMs use a single doc-level configuration each (no template variants); see \Cref{tab:doc_comet} for the open-vs-closed doc-level comparison. \textbf{Bold}: per-language column maximum within each model block. The bottom \textbf{avg} row of each block reports the per-language average across configurations.}
\label{tab:doc_template_shot_ablation_comet}
\end{table*}

\begin{table*}[t]
\centering
\resizebox{\textwidth}{!}{
\begin{tabular}{lll|rrrrrr|rrrrrr|rrrrrr|rrrrrr}
\toprule
\multirow{2}{*}{\textbf{dir}} & \multirow{2}{*}{\textbf{model}} & \multirow{2}{*}{\textbf{setting}} & \multicolumn{6}{l|}{\textbf{fluency}} & \multicolumn{6}{l|}{\textbf{accuracy errors}} & \multicolumn{6}{l|}{\textbf{lexical mistakes}} & \multicolumn{6}{l}{\textbf{grammatical mistakes}} \\
\cmidrule(lr){4-9} \cmidrule(lr){10-15} \cmidrule(lr){16-21} \cmidrule(lr){22-27}
& & & \textbf{\texttt{amh}} & \textbf{\texttt{hau}} & \textbf{\texttt{lug}} & \textbf{\texttt{nso}} & \textbf{\texttt{yor}} & \textbf{\texttt{zul}} & \textbf{\texttt{amh}} & \textbf{\texttt{hau}} & \textbf{\texttt{lug}} & \textbf{\texttt{nso}} & \textbf{\texttt{yor}} & \textbf{\texttt{zul}} & \textbf{\texttt{amh}} & \textbf{\texttt{hau}} & \textbf{\texttt{lug}} & \textbf{\texttt{nso}} & \textbf{\texttt{yor}} & \textbf{\texttt{zul}} & \textbf{\texttt{amh}} & \textbf{\texttt{hau}} & \textbf{\texttt{lug}} & \textbf{\texttt{nso}} & \textbf{\texttt{yor}} & \textbf{\texttt{zul}} \\
\midrule
\multirow{28}{*}{eng$\to$} & \multirow{7}{*}{Gemini-1.5} & $1_{0}$ & 2.9 & 3.2 & 2.1 & 2.7 & 2.3 & 2.4 & 1.7 & 2.3 & 4.0 & 3.3 & 3.5 & 3.9 & 1.7 & 1.9 & 2.9 & 2.5 & 2.5 & 2.8 & 1.5 & 1.7 & 2.1 & 2.2 & 1.9 & 2.3 \\
 &  & $1_{1}$ & 3.0 & 3.0 & 2.1 & 2.6 & 2.3 & 2.4 & 2.2 & 2.2 & 3.7 & 3.5 & 3.2 & 4.1 & 1.7 & 1.9 & 2.5 & 2.8 & 2.3 & 2.8 & 1.5 & 1.8 & 2.0 & 2.1 & 1.9 & 2.2 \\
 &  & $2_{0}$ & 2.9 & 3.1 & 2.1 & 2.7 & 2.3 & 2.5 & 1.8 & 2.2 & 3.7 & 3.3 & 3.5 & 3.9 & 1.7 & 1.9 & 2.8 & 2.8 & 2.3 & 2.8 & 1.5 & 1.7 & 2.1 & 2.1 & 1.8 & 2.2 \\
 &  & $2_{1}$ & 2.9 & 3.1 & 2.1 & 2.7 & 2.3 & 2.5 & 1.8 & 2.3 & 3.8 & 3.5 & 3.4 & 3.9 & 1.7 & 1.8 & 2.6 & 2.6 & 2.3 & 3.0 & 1.5 & 1.7 & 1.9 & 2.1 & 2.0 & 2.4 \\
 &  & $3_{0}$ & 2.9 & 3.1 & 2.1 & 2.7 & 2.2 & 2.5 & 1.8 & 2.2 & 3.5 & 3.3 & 3.2 & 3.9 & 1.6 & 2.0 & 2.7 & 2.6 & 2.3 & 2.8 & 1.6 & 1.7 & 2.1 & 1.9 & 1.8 & 2.2 \\
 &  & $3_{1}$ & 2.9 & 3.1 & 2.1 & 2.7 & 2.1 & 2.4 & 1.9 & 2.3 & 3.6 & 3.1 & 3.4 & 3.9 & 1.9 & 1.9 & 2.5 & 2.6 & 2.3 & 2.9 & 1.6 & 1.7 & 2.0 & 2.0 & 1.9 & 2.3 \\
\cmidrule{3-27}
 & & avg & 2.9 & 3.1 & 2.1 & 2.7 & 2.2 & 2.5 & 1.9 & 2.2 & 3.7 & 3.3 & 3.4 & 3.9 & 1.7 & 1.9 & 2.7 & 2.7 & 2.3 & 2.8 & 1.5 & 1.7 & 2.0 & 2.1 & 1.9 & 2.2 \\
\cmidrule{2-27}
 & \multirow{7}{*}{Gemini-3.1} & $1_{0}$ & 4.3 & 4.2 & 3.6 & 3.5 & 4.1 & 4.0 & 0.5 & 0.9 & 2.5 & 2.5 & 1.6 & 0.6 & 0.9 & 1.1 & 1.9 & 1.9 & 1.4 & 1.1 & 0.9 & 1.2 & 1.6 & 1.8 & 1.3 & 1.3 \\
 &  & $1_{1}$ & 3.9 & 4.1 & 3.3 & 3.5 & 3.9 & 3.8 & 0.8 & 0.7 & 2.6 & 2.5 & 1.7 & 0.8 & 1.0 & 1.3 & 2.1 & 2.0 & 1.5 & 1.4 & 1.2 & 1.3 & 2.0 & 1.9 & 1.4 & 1.5 \\
 &  & $2_{0}$ & 4.3 & 4.2 & 3.7 & 3.7 & 4.2 & 4.1 & 0.8 & 0.9 & 2.3 & 2.8 & 1.8 & 0.7 & 0.9 & 1.1 & 1.9 & 1.9 & 1.3 & 1.2 & 0.9 & 1.2 & 1.8 & 1.8 & 1.2 & 1.3 \\
 &  & $2_{1}$ & 4.0 & 4.2 & 3.6 & 3.6 & 3.5 & 3.8 & 0.8 & 0.6 & 2.4 & 2.5 & 1.8 & 1.0 & 1.0 & 1.1 & 1.8 & 2.0 & 1.5 & 1.3 & 1.1 & 1.2 & 1.8 & 1.9 & 1.4 & 1.4 \\
 &  & $3_{0}$ & 3.9 & 4.1 & 3.5 & 3.5 & 3.8 & 3.9 & 0.7 & 0.8 & 2.8 & 2.5 & 1.5 & 0.9 & 1.1 & 1.3 & 2.0 & 1.9 & 1.3 & 1.3 & 1.0 & 1.3 & 1.8 & 1.8 & 1.3 & 1.4 \\
 &  & $3_{1}$ & 3.9 & 3.9 & 3.3 & 3.5 & 3.4 & 3.7 & 0.7 & 0.8 & 2.7 & 2.5 & 1.6 & 0.9 & 1.1 & 1.2 & 1.9 & 1.9 & 1.6 & 1.3 & 1.1 & 1.3 & 1.7 & 1.8 & 1.5 & 1.5 \\
\cmidrule{3-27}
 & & avg & 4.1 & 4.1 & 3.5 & 3.5 & 3.8 & 3.9 & 0.7 & 0.8 & 2.6 & 2.6 & 1.7 & 0.8 & 1.0 & 1.2 & 1.9 & 1.9 & 1.4 & 1.3 & 1.0 & 1.3 & 1.8 & 1.8 & 1.4 & 1.4 \\
\cmidrule{2-27}
 & \multirow{7}{*}{GPT-4o} & $1_{0}$ & 1.4 & 3.3 & 2.1 & 2.8 & 2.8 & 3.3 & 4.5 & 2.3 & 3.8 & 3.7 & 2.9 & 2.3 & 2.7 & 1.8 & 2.6 & 2.5 & 2.1 & 2.0 & 2.4 & 1.7 & 2.2 & 2.1 & 2.1 & 2.1 \\
 &  & $1_{1}$ & 1.3 & 3.3 & 2.5 & 2.9 & 1.9 & 3.3 & 4.9 & 1.9 & 3.8 & 3.6 & 4.5 & 2.2 & 2.5 & 1.7 & 2.5 & 2.4 & 2.6 & 1.8 & 2.2 & 1.8 & 2.2 & 2.1 & 2.5 & 2.0 \\
 &  & $2_{0}$ & 1.3 & 3.3 & 2.0 & 2.7 & 2.8 & 3.4 & 5.5 & 1.9 & 3.7 & 3.8 & 3.2 & 2.6 & 2.5 & 1.6 & 2.3 & 2.6 & 2.1 & 1.9 & 2.2 & 1.7 & 2.2 & 2.1 & 2.1 & 1.9 \\
 &  & $2_{1}$ & 1.3 & 3.4 & 2.4 & 2.9 & 1.9 & 3.3 & 4.5 & 1.9 & 3.7 & 3.6 & 4.0 & 2.3 & 2.5 & 1.6 & 2.7 & 2.5 & 2.1 & 1.9 & 2.3 & 1.8 & 2.5 & 2.1 & 2.1 & 2.0 \\
 &  & $3_{0}$ & 1.4 & 3.3 & 2.1 & 2.6 & 2.8 & 3.3 & 4.7 & 2.3 & 3.6 & 4.0 & 3.1 & 2.6 & 2.4 & 1.9 & 2.7 & 2.6 & 2.1 & 2.0 & 2.4 & 1.9 & 2.5 & 2.4 & 2.0 & 2.1 \\
 &  & $3_{1}$ & 1.2 & 3.3 & 2.6 & 2.9 & 1.6 & 3.3 & 5.4 & 1.9 & 3.9 & 3.5 & 4.7 & 2.5 & 2.7 & 1.8 & 2.5 & 2.4 & 2.2 & 2.0 & 2.2 & 1.8 & 2.2 & 2.1 & 2.2 & 2.0 \\
\cmidrule{3-27}
 & & avg & 1.3 & 3.3 & 2.3 & 2.8 & 2.3 & 3.3 & 4.9 & 2.0 & 3.8 & 3.7 & 3.7 & 2.4 & 2.5 & 1.7 & 2.5 & 2.5 & 2.2 & 2.0 & 2.3 & 1.8 & 2.3 & 2.1 & 2.2 & 2.0 \\
\cmidrule{2-27}
 & \multirow{7}{*}{GPT-5.4} & $1_{0}$ & 3.7 & 4.4 & 3.6 & 3.9 & 4.0 & 4.2 & 1.6 & 0.4 & 2.3 & 2.4 & 2.3 & 0.3 & 1.5 & 1.0 & 1.9 & 1.5 & 1.4 & 1.2 & 1.6 & 1.0 & 1.8 & 1.5 & 1.4 & 1.3 \\
 &  & $1_{1}$ & 3.4 & 4.2 & 3.6 & 3.6 & 3.6 & 4.0 & 1.1 & 0.7 & 2.4 & 2.2 & 2.6 & 0.2 & 1.5 & 1.2 & 2.1 & 1.9 & 1.8 & 1.2 & 1.6 & 1.3 & 1.8 & 1.8 & 1.7 & 1.5 \\
 &  & $2_{0}$ & 3.6 & 4.3 & 3.8 & 3.8 & 3.9 & 4.3 & 1.4 & 0.7 & 2.1 & 1.9 & 1.9 & 0.1 & 1.6 & 1.2 & 1.9 & 1.7 & 1.5 & 1.1 & 1.7 & 1.2 & 1.7 & 1.6 & 1.5 & 1.4 \\
 &  & $2_{1}$ & 3.5 & 4.1 & 3.5 & 3.6 & 3.5 & 3.9 & 1.1 & 0.7 & 2.2 & 2.2 & 2.5 & 0.3 & 1.6 & 1.2 & 2.0 & 1.6 & 1.8 & 1.3 & 1.6 & 1.4 & 1.8 & 1.6 & 1.7 & 1.6 \\
 &  & $3_{0}$ & 3.6 & 4.2 & 3.5 & 3.7 & 3.8 & 4.2 & 1.4 & 0.5 & 2.4 & 2.2 & 2.0 & 0.2 & 1.7 & 1.2 & 2.1 & 1.8 & 1.5 & 1.2 & 1.7 & 1.1 & 1.9 & 1.8 & 1.5 & 1.4 \\
 &  & $3_{1}$ & 3.4 & 3.9 & 3.4 & 3.5 & 3.4 & 3.9 & 1.0 & 0.6 & 2.0 & 1.9 & 2.4 & 0.2 & 1.5 & 1.2 & 1.9 & 1.8 & 1.9 & 1.1 & 1.6 & 1.2 & 1.8 & 1.7 & 1.8 & 1.4 \\
\cmidrule{3-27}
 & & avg & 3.5 & 4.2 & 3.6 & 3.7 & 3.7 & 4.1 & 1.3 & 0.6 & 2.2 & 2.1 & 2.3 & 0.2 & 1.5 & 1.2 & 2.0 & 1.7 & 1.6 & 1.2 & 1.6 & 1.2 & 1.8 & 1.7 & 1.6 & 1.4 \\
\midrule
\multirow{28}{*}{$\to$eng} & \multirow{7}{*}{Gemini-1.5} & $1_{0}$ & 3.9 & 3.9 & 3.8 & 3.9 & 3.8 & 3.9 & 1.9 & 2.8 & 3.9 & 3.6 & 4.5 & 2.9 & 1.4 & 1.3 & 1.5 & 1.9 & 1.8 & 1.3 & 1.2 & 1.0 & 1.2 & 1.5 & 1.5 & 1.2 \\
 &  & $1_{1}$ & 3.9 & 3.9 & 3.7 & 3.9 & 3.7 & 3.9 & 2.6 & 2.9 & 3.4 & 3.4 & 3.5 & 2.9 & 1.5 & 1.2 & 1.6 & 1.5 & 1.4 & 1.5 & 1.3 & 1.1 & 1.3 & 1.2 & 1.1 & 1.2 \\
 &  & $2_{0}$ & 4.0 & 4.0 & 3.9 & 3.9 & 3.7 & 3.9 & 2.4 & 2.9 & 3.6 & 3.4 & 3.5 & 2.3 & 1.6 & 1.2 & 1.7 & 1.4 & 1.5 & 1.1 & 1.4 & 1.0 & 1.2 & 1.0 & 1.4 & 1.1 \\
 &  & $2_{1}$ & 3.9 & 4.0 & 3.9 & 3.9 & 3.8 & 3.9 & 2.4 & 2.7 & 3.1 & 3.2 & 3.3 & 2.8 & 1.0 & 1.4 & 1.3 & 1.5 & 1.5 & 1.5 & 1.2 & 1.2 & 1.1 & 1.1 & 1.3 & 1.4 \\
 &  & $3_{0}$ & 3.8 & 3.8 & 3.8 & 3.7 & 3.6 & 3.9 & 2.3 & 2.9 & 3.3 & 3.4 & 3.9 & 2.8 & 1.4 & 1.8 & 1.4 & 1.5 & 1.6 & 1.2 & 1.2 & 1.5 & 1.2 & 1.3 & 1.4 & 1.2 \\
 &  & $3_{1}$ & 3.9 & 3.9 & 3.8 & 3.8 & 3.6 & 3.9 & 2.1 & 2.6 & 2.9 & 3.3 & 3.7 & 3.2 & 1.2 & 1.2 & 1.4 & 1.5 & 1.3 & 1.3 & 1.2 & 1.2 & 1.4 & 1.3 & 1.2 & 1.4 \\
\cmidrule{3-27}
 & & avg & 3.9 & 3.9 & 3.8 & 3.9 & 3.7 & 3.9 & 2.3 & 2.8 & 3.4 & 3.4 & 3.7 & 2.8 & 1.4 & 1.3 & 1.5 & 1.6 & 1.5 & 1.3 & 1.3 & 1.2 & 1.2 & 1.2 & 1.3 & 1.3 \\
\cmidrule{2-27}
 & \multirow{7}{*}{Gemini-3.1} & $1_{0}$ & 4.2 & 4.3 & 4.1 & 4.1 & 3.7 & 4.3 & 2.2 & 2.4 & 3.3 & 3.0 & 3.9 & 2.4 & 1.8 & 1.3 & 1.5 & 1.2 & 2.1 & 1.4 & 1.4 & 1.1 & 1.2 & 1.2 & 1.6 & 1.3 \\
 &  & $1_{1}$ & 4.1 & 4.1 & 3.9 & 3.8 & 3.6 & 3.9 & 2.4 & 2.4 & 2.8 & 2.7 & 4.4 & 2.1 & 1.5 & 1.4 & 1.6 & 1.4 & 1.9 & 1.3 & 1.1 & 1.1 & 1.3 & 1.3 & 1.6 & 1.2 \\
 &  & $2_{0}$ & 4.2 & 4.3 & 4.2 & 4.2 & 3.8 & 4.3 & 2.4 & 2.4 & 2.5 & 3.8 & 3.7 & 2.4 & 1.6 & 1.2 & 1.7 & 1.3 & 1.7 & 1.4 & 1.2 & 1.1 & 1.3 & 1.2 & 1.4 & 1.2 \\
 &  & $2_{1}$ & 4.1 & 4.2 & 4.0 & 4.0 & 3.6 & 4.2 & 2.0 & 2.5 & 3.1 & 3.2 & 4.1 & 2.4 & 1.6 & 1.4 & 1.5 & 1.3 & 1.8 & 1.5 & 1.3 & 1.2 & 1.1 & 1.2 & 1.6 & 1.2 \\
 &  & $3_{0}$ & 4.0 & 4.1 & 3.9 & 4.0 & 3.6 & 4.1 & 2.3 & 2.4 & 3.5 & 3.0 & 4.6 & 2.9 & 1.2 & 1.4 & 1.4 & 1.4 & 2.0 & 1.4 & 1.1 & 1.2 & 1.3 & 1.3 & 1.6 & 1.2 \\
 &  & $3_{1}$ & 3.9 & 4.0 & 3.8 & 3.9 & 3.5 & 3.9 & 2.4 & 2.3 & 3.0 & 2.6 & 4.2 & 2.5 & 1.2 & 1.3 & 1.6 & 1.4 & 2.0 & 1.2 & 1.1 & 1.1 & 1.4 & 1.2 & 1.8 & 1.2 \\
\cmidrule{3-27}
 & & avg & 4.1 & 4.2 & 4.0 & 4.0 & 3.6 & 4.1 & 2.3 & 2.4 & 3.0 & 3.1 & 4.1 & 2.5 & 1.5 & 1.3 & 1.6 & 1.3 & 1.9 & 1.4 & 1.2 & 1.1 & 1.3 & 1.2 & 1.6 & 1.2 \\
\cmidrule{2-27}
 & \multirow{7}{*}{GPT-4o} & $1_{0}$ & 3.4 & 3.4 & 3.4 & 3.4 & 3.1 & 3.5 & 4.0 & 4.1 & 3.8 & 4.2 & 4.9 & 3.7 & 1.8 & 2.0 & 1.8 & 1.7 & 2.1 & 2.0 & 1.6 & 1.6 & 1.5 & 1.6 & 1.9 & 1.5 \\
 &  & $1_{1}$ & 3.5 & 3.6 & 3.6 & 3.5 & 3.2 & 3.8 & 3.4 & 3.9 & 3.6 & 3.8 & 4.3 & 3.1 & 1.6 & 1.9 & 1.8 & 1.5 & 1.9 & 1.5 & 1.4 & 1.5 & 1.5 & 1.4 & 1.6 & 1.2 \\
 &  & $2_{0}$ & 3.4 & 3.4 & 3.4 & 3.4 & 3.1 & 3.6 & 4.2 & 3.6 & 3.7 & 4.8 & 5.1 & 4.3 & 1.8 & 2.1 & 2.1 & 2.2 & 2.5 & 1.5 & 1.6 & 1.6 & 1.7 & 1.6 & 2.1 & 1.5 \\
 &  & $2_{1}$ & 3.5 & 3.6 & 3.6 & 3.7 & 3.3 & 3.7 & 3.6 & 3.9 & 4.0 & 4.3 & 4.2 & 2.8 & 1.5 & 1.7 & 1.8 & 1.8 & 2.1 & 1.6 & 1.3 & 1.4 & 1.6 & 1.5 & 1.8 & 1.4 \\
 &  & $3_{0}$ & 3.4 & 3.5 & 3.4 & 3.5 & 3.2 & 3.6 & 4.1 & 3.5 & 3.9 & 4.4 & 4.6 & 3.8 & 1.7 & 1.9 & 2.0 & 1.7 & 2.1 & 1.9 & 1.5 & 1.6 & 1.5 & 1.6 & 1.8 & 1.7 \\
 &  & $3_{1}$ & 3.4 & 3.7 & 3.7 & 3.6 & 3.3 & 3.8 & 3.8 & 3.3 & 3.8 & 3.9 & 4.8 & 3.2 & 1.6 & 2.0 & 1.6 & 1.6 & 1.9 & 1.3 & 1.3 & 1.6 & 1.4 & 1.5 & 1.8 & 1.3 \\
\cmidrule{3-27}
 & & avg & 3.5 & 3.5 & 3.5 & 3.5 & 3.2 & 3.7 & 3.8 & 3.7 & 3.8 & 4.2 & 4.6 & 3.5 & 1.7 & 1.9 & 1.8 & 1.8 & 2.1 & 1.6 & 1.4 & 1.6 & 1.5 & 1.5 & 1.8 & 1.4 \\
\cmidrule{2-27}
 & \multirow{7}{*}{GPT-5.4} & $1_{0}$ & 4.4 & 4.4 & 4.4 & 4.4 & 4.2 & 4.5 & 1.6 & 2.1 & 2.6 & 3.0 & 2.8 & 2.6 & 1.4 & 1.5 & 1.2 & 1.3 & 1.5 & 1.2 & 0.9 & 0.9 & 1.1 & 1.1 & 1.1 & 1.1 \\
 &  & $1_{1}$ & 4.2 & 4.2 & 4.2 & 4.1 & 4.1 & 4.2 & 1.8 & 1.9 & 2.0 & 2.9 & 3.1 & 1.9 & 1.5 & 1.4 & 1.6 & 1.1 & 1.4 & 1.1 & 1.3 & 1.2 & 1.1 & 1.1 & 1.2 & 1.1 \\
 &  & $2_{0}$ & 4.4 & 4.4 & 4.3 & 4.3 & 4.3 & 4.4 & 1.8 & 1.7 & 2.3 & 2.7 & 2.7 & 2.2 & 1.5 & 1.1 & 1.2 & 1.6 & 1.4 & 1.3 & 1.1 & 0.9 & 1.1 & 1.2 & 1.2 & 1.2 \\
 &  & $2_{1}$ & 4.2 & 4.2 & 4.2 & 4.2 & 4.2 & 4.1 & 1.8 & 1.7 & 2.1 & 2.9 & 3.0 & 1.5 & 1.3 & 1.2 & 1.3 & 1.0 & 1.2 & 0.8 & 1.2 & 1.2 & 1.2 & 1.0 & 1.2 & 0.9 \\
 &  & $3_{0}$ & 4.3 & 4.3 & 4.3 & 4.3 & 4.1 & 4.2 & 1.5 & 2.5 & 2.6 & 3.2 & 2.7 & 2.3 & 0.9 & 1.6 & 1.6 & 1.4 & 1.1 & 1.4 & 0.9 & 1.2 & 1.3 & 1.1 & 1.1 & 1.2 \\
 &  & $3_{1}$ & 4.0 & 4.0 & 4.1 & 4.1 & 3.9 & 4.1 & 1.4 & 2.5 & 2.1 & 2.7 & 3.2 & 1.8 & 1.5 & 1.0 & 1.4 & 1.2 & 1.5 & 1.1 & 1.3 & 0.9 & 1.2 & 1.1 & 1.2 & 1.1 \\
\cmidrule{3-27}
 & & avg & 4.2 & 4.2 & 4.2 & 4.2 & 4.2 & 4.3 & 1.6 & 2.1 & 2.3 & 2.9 & 2.9 & 2.0 & 1.3 & 1.3 & 1.4 & 1.3 & 1.4 & 1.2 & 1.1 & 1.1 & 1.2 & 1.1 & 1.2 & 1.1 \\
\bottomrule
\end{tabular}
}
\caption{LLM-judge evaluation of the four closed models' document-level translations on the six template configurations common to the older and newer models. Scores are the per-pair mean of two judges (GPT-5.5 and Gemini-3.1-Flash-Lite). Higher fluency is better; lower accuracy / lexical / grammatical mistake counts are better.}
\label{tab:doc_evaluation}
\end{table*}

\begin{table*}
\centering
\small
\renewcommand{\arraystretch}{1.05}
\begin{tabular}{lrrrrr}
\toprule
\textbf{metric} & \textbf{Pearson} & \textbf{Spearman} & \textbf{GPT-5.5} & \textbf{Gemini-3.1} & \textbf{mean$|\Delta|$} \\
\midrule
fluency               &  0.84 &  0.70 & 3.32 & 3.75 & 0.47 \\
accuracy errors       &  0.12 &  0.10 & 0.71 & 4.71 & 4.01 \\
lexical mistakes      & -0.59 & -0.63 & 0.38 & 3.03 & 2.67 \\
grammatical mistakes  & -0.60 & -0.66 & 0.27 & 2.79 & 2.53 \\
\bottomrule
\end{tabular}
\caption{Inter-judge agreement on the four document-level evaluation dimensions. Each row pools 288 paired observations (four closed models $\times$ six template configurations $\times$ six target languages $\times$ two directions). The two judges agree closely on fluency but interpret the absolute error count scale differently, with the Gemini-3.1-Flash-Lite judge reporting 7--11 times more errors than the GPT-5.5 judge, and the count-based metrics showing near-zero or negative correlation.}
\label{tab:judge_agreement}
\end{table*}

\begin{prompttemplate*}
\centering
\begin{minipage}{\textwidth}
\begin{tcolorbox}[enhanced, breakable, colback=metricsbg,
  colframe=white, arc=2mm, boxrule=0pt, width=\textwidth]
\begin{lstlisting}[basicstyle=\ttfamily\footnotesize, breaklines=true]
Translate {src_lang} scientific texts to {tgt_lang}.

Examples:
{examples}

Now translate:
{src_lang}: {src_text}
{tgt_lang}:
\end{lstlisting}
\end{tcolorbox}
\subcaption{Template 0: Vanilla Template}
\label{prompt:sent_template_vanilla}
\end{minipage}

\vspace{0.2cm}

\begin{minipage}{\textwidth}
\begin{tcolorbox}[enhanced, breakable, colback=metricsbg,
  colframe=white, arc=2mm, boxrule=0pt, width=\textwidth]
\begin{lstlisting}[basicstyle=\ttfamily\footnotesize, breaklines=true]
Instruction:
Translate the following text from {src_lang} to {tgt_lang}. Ensure that the meaning is preserved, and the translation sounds natural. Do not send the {src_lang} text back in the response, generate only the translation and nothing more.

{src_lang}: {text}

{tgt_lang}:
\end{lstlisting}
\end{tcolorbox}
\subcaption{Template 1: Basic translation}
\label{prompt:sent_template1}
\end{minipage}

\vspace{0.2cm}

\begin{minipage}{\textwidth}
\begin{tcolorbox}[enhanced, breakable, colback=metricsbg,
  colframe=white, arc=2mm, boxrule=0pt, width=\textwidth]
\begin{lstlisting}[basicstyle=\ttfamily\footnotesize, breaklines=true]
Instruction:
Translate the following scientific text from {src_lang} to {tgt_lang}, ensuring that all technical terms are accurately translated with no additional explanations. Do not send the {src_lang} text back in the response, generate only the translation and nothing more.

{src_lang}: {text}

{tgt_lang}:
\end{lstlisting}
\end{tcolorbox}
\subcaption{Template 2: Scientific translation}
\label{prompt:sent_template2}
\end{minipage}

\vspace{0.2cm}

\begin{minipage}{\textwidth}
\begin{tcolorbox}[enhanced, breakable, colback=metricsbg,
  colframe=white, arc=2mm, boxrule=0pt, width=\textwidth]
\begin{lstlisting}[basicstyle=\ttfamily\footnotesize, breaklines=true]
Instruction:
Translate the following scientific text from {src_lang} to {tgt_lang}, using the provided translation examples for consistency. Do not send the {src_lang} text back in the response, generate only the translation and nothing more.

Translation examples:
{examples}

Now generate the {tgt_lang} translation of the following {src_lang} text:

{src_lang}: {text}

{tgt_lang}:
\end{lstlisting}
\end{tcolorbox}
\subcaption{Template 3: Translation with examples}
\label{prompt:sent_template3}
\end{minipage}

\caption{Sentence-level prompt templates for translation evaluation (continued on next page)}
\label{prompt:sent_templates}
\end{prompttemplate*}

\begin{prompttemplate*}[p]
\ContinuedFloat
\centering
\begin{minipage}{\textwidth}
\begin{tcolorbox}[enhanced, breakable, colback=metricsbg,
  colframe=white, arc=2mm, boxrule=0pt, width=\textwidth]
\begin{lstlisting}[basicstyle=\ttfamily\footnotesize, breaklines=true]
Instruction:
Translate the following scientific text from {src_lang} to {tgt_lang}, using the following glossary (if provided) to ensure accurate technical term translation. Do not send the {src_lang} text back in the response, generate only the translation and nothing more.

Glossary terms:
{glossary}

Now generate the {tgt_lang} translation of the following {src_lang} text:

{src_lang}: {text}

{tgt_lang}:
\end{lstlisting}
\end{tcolorbox}
\subcaption{Template 4: Translation with glossary}
\label{prompt:sent_template4}
\end{minipage}

\vspace{0.2cm}

\begin{minipage}{\textwidth}
\begin{tcolorbox}[enhanced, breakable, colback=metricsbg,
  colframe=white, arc=2mm, boxrule=0pt, width=\textwidth]
\begin{lstlisting}[basicstyle=\ttfamily\footnotesize, breaklines=true]
Instruction:
Translate the following scientific text from {src_lang} to {tgt_lang}, using both the provided examples and glossary (if provided) for consistency. Do not send the {src_lang} text back in the response, generate only the translation and nothing more.

Glossary terms:
{glossary}

Translation examples:
{examples}

Now generate the {tgt_lang} translation of the following 
{src_lang} text:

{src_lang}: {text}

{tgt_lang}:
\end{lstlisting}
\end{tcolorbox}
\subcaption{Template 5: Translation with examples and glossary}
\label{prompt:sent_template5}
\end{minipage}

\caption{Sentence-level prompt templates for translation evaluation. Templates vary in the amount of context provided: basic instruction (1), scientific focus (2), with examples (3), with glossary (4), and with both examples and glossary (5).}
\end{prompttemplate*}

\begin{prompttemplate*}[p]
\centering
\begin{minipage}{\textwidth}
\begin{tcolorbox}[enhanced, breakable, colback=metricsbg,
  colframe=white, arc=2mm, boxrule=0pt, width=\textwidth]
\begin{lstlisting}[basicstyle=\ttfamily\footnotesize, breaklines=true]
Instruction:
Translate the following document from {src_lang} to {tgt_lang}. Do not send the {src_lang} document back in the response; generate only the translation and nothing more.<<shots>>

{src_lang} document: {src_document}

{tgt_lang} document:
\end{lstlisting}
\end{tcolorbox}
\subcaption{Template 1: Basic document translation}
\label{prompt:doc_template1}
\end{minipage}

\vspace{0.2cm}

\begin{minipage}{\textwidth}
\begin{tcolorbox}[enhanced, breakable, colback=metricsbg,
  colframe=white, arc=2mm, boxrule=0pt, width=\textwidth]
\begin{lstlisting}[basicstyle=\ttfamily\footnotesize, breaklines=true]
Instruction:
Translate the following {domain} document from {src_lang} to {tgt_lang}. Do not send the {src_lang} document back in the response; generate only the translation and nothing more.<<shots>>

{src_lang} document: {src_document}

{tgt_lang} document:
\end{lstlisting}
\end{tcolorbox}
\subcaption{Template 2: Domain-specific document translation}
\label{prompt:doc_template2}
\end{minipage}

\vspace{0.2cm}

\begin{minipage}{\textwidth}
\begin{tcolorbox}[enhanced, breakable, colback=metricsbg,
  colframe=white, arc=2mm, boxrule=0pt, width=\textwidth]
\begin{lstlisting}[basicstyle=\ttfamily\footnotesize, breaklines=true]
Instruction:
Please provide the {tgt_lang} translation for the following. Provide only the translation.<<shots>>

{src_lang} document: {src_document}

{tgt_lang} document:
\end{lstlisting}
\end{tcolorbox}
\subcaption{Template 3: Simplified document translation}
\label{prompt:doc_template3}
\end{minipage}

\vspace{0.2cm}

\begin{minipage}{\textwidth}
\begin{tcolorbox}[enhanced, breakable, colback=metricsbg,
  colframe=white, arc=2mm, boxrule=0pt, width=\textwidth]
\begin{lstlisting}[basicstyle=\ttfamily\footnotesize, breaklines=true]
Use the example below for consistency.

Example:
{src_lang} document: {example_src_document}
{tgt_lang} document: {example_tgt_document}

[Additional examples if n>1...]

Now translate the following.
\end{lstlisting}
\end{tcolorbox}
\subcaption{Few-shot example format}
\label{prompt:fewshot_format}
\end{minipage}

\caption{Document-level prompt templates for translation generation. The \texttt{<<shots>>} placeholder in templates 1-3 is replaced with the few-shot format (d) when $n>0$. Few-shot examples are randomly sampled from the training set, matching the target document's domain.}
\label{prompt:doc_templates}
\end{prompttemplate*}
\begin{prompttemplate*}
\centering
\begin{tcolorbox}[enhanced, breakable, colback=metricsbg,
  colframe=white, arc=2mm, boxrule=0pt, width=\textwidth]
\begin{lstlisting}[basicstyle=\ttfamily\footnotesize, breaklines=true]
Please evaluate the fluency of the following text in {tgt_lang}.
------
### **Instructions:**
- **Task**: Evaluate the fluency of the text.
- Scoring: Provide a score from 1 to 5, where:
  - **5**: The text is **highly fluent**, with no grammatical errors, unnatural wording, or stiff syntax.
  - **4**: The text is **mostly fluent**, with minor errors that do not impede understanding.
  - **3**: The text is **moderately fluent**, with noticeable errors that may slightly affect comprehension.
  - **2**: The text has **low fluency**, with frequent errors that hinder understanding.
  - **1**: The text is **not fluent**, with severe errors that make it difficult to understand.
- **Explanation**: Support your score with specific examples to justify your evaluation.
------
### **Output Format:**
Provide your evaluation in the following JSON format:
{
  "Fluency": {
    "Score": "<the score>",
    "Explanation": "<your explanation on how you made the decision>"
  }
}
------
**Text to Evaluate:**
\end{lstlisting}
\end{tcolorbox}
\caption{Fluency evaluation for document-level translation evaluation}
\label{prompt:fluency_eval}
\end{prompttemplate*}

\begin{prompttemplate*}
\centering
\begin{tcolorbox}[enhanced, breakable, colback=metricsbg,
  colframe=white, arc=2mm, boxrule=0pt, width=\textwidth]
\begin{lstlisting}[basicstyle=\ttfamily\footnotesize, breaklines=true]
Please evaluate the accuracy of the following translated text in {tgt_lang} by comparing it to the provided reference text.
------
### **Instructions:**
- **Task**: Compare the text to the reference text.
- Identify Mistakes: List all mistakes related to accuracy.
- Mistake Types:
  - **Wrong Translation**: Incorrect meaning or misinterpretation leading to wrong information.
  - **Omission**: Missing words, phrases, or information present in the reference text.
  - **Addition**: Extra words, phrases, or information not present in the reference text.
  - **Others**: Mistakes that are hard to define or categorize.
- **Note**: If the text expresses the same information as the reference text but uses different words or phrasing, it is **not** considered a mistake.
- **Provide a List**: Summarize all mistakes without repeating the exact sentences. Provide an empty list if there are no mistakes.
------
### **Output Format:**
Provide your evaluation in the following JSON format:
{
  "Accuracy": {
    "Mistakes": [
      "<list of all mistakes in the text with format 'Mistake Type: summarize the mistake', provide an empty list if there are no mistakes>"
    ]
  }
}
------
**Reference Text:**
{ref}
------
**Text to Evaluate:**
\end{lstlisting}
\end{tcolorbox}
\caption{Accuracy evaluation for document-level translation evaluation}
\label{prompt:accuracy_eval}
\end{prompttemplate*}

\vspace{0.2cm}

\begin{prompttemplate*}
\centering
\begin{tcolorbox}[enhanced, breakable, colback=metricsbg,
  colframe=white, arc=2mm, boxrule=0pt, width=\textwidth]
\begin{lstlisting}[basicstyle=\ttfamily\footnotesize, breaklines=true]
Please evaluate the cohesion of the following translated text in {tgt_lang} by comparing it to the provided reference text.
------
### **Instructions:**
- **Task**: Evaluate the cohesion of the text.
- **Definition**: Cohesion refers to how different parts of a text are connected using language structures like grammar and vocabulary. It ensures that sentences flow smoothly and the text makes sense as a whole.
- Identify Mistakes: List all mistakes related to cohesion.
- Separate the mistakes into:
  - **Lexical Cohesion Mistakes**: Issues with vocabulary usage, incorrect or missing synonyms, or overuse of certain words that disrupt the flow.
  - **Grammatical Cohesion Mistakes**: Problems with pronouns, conjunctions, or grammatical structures that link sentences and clauses.
- **Provide Lists**: Provide separate lists for lexical cohesion mistakes and grammatical cohesion mistakes. Provide empty lists if there are no mistakes.
------
### **Output Format:**
Provide your evaluation in the following JSON format:
{
  "Cohesion": {
    "Lexical Cohesion Mistakes": [
      "<list of all mistakes in the text one by one, provide an empty list if there are no mistakes>"
    ],
    "Grammatical Cohesion Mistakes": [
      "<list of all mistakes in the text one by one, provide an empty list if there are no mistakes>"
    ]
  }
}
------
**Reference Text:**
{ref}
------
**Text to Evaluate:**
\end{lstlisting}
\end{tcolorbox}
\caption{Cohesion evaluation for document-level translation evaluation}
\label{prompt:cohesion_eval}
\end{prompttemplate*}
\begin{promptexample*}
\centering
\begin{tcolorbox}[enhanced, breakable, colback=metricsbg,
  colframe=white, arc=2mm, boxrule=0pt, width=\textwidth]
\begin{lstlisting}[basicstyle=\ttfamily\footnotesize, breaklines=true, columns=flexible, literate={ƴ}{{\'{y}}}1 {ɗ}{{\d{d}}}1 {ƙ}{{\'{k}}}1 {ɓ}{{\d{b}}}1 {ɓ}{{\d{b}}}1 {ɗ}{{\d{d}}}1]
System input:

Instruction:
Translate the following scientific text from English to Hausa, using the provided glossary terms and translation examples for consistency. Do not send the English text back in the response.

Glossary terms:
1. English: covid; Hausa: covid
2. English: covid-19; Hausa: covid-19
3. English: infection; Hausa: cuta
4. English: syndrome; Hausa: ciwo

Translation examples:
1. English: About 75% of women who had been pregnant when they were between ages 15 and 19 used contraceptives at the time of the study.
   Hausa: Kimanin kashi 75% na matan da suke da juna biyu lokacin da suke tsakanin shekaru 15 zuwa 19 sun yi amfani da rigakafin hana haihuwa a lokacin binciken.

2. English: Moreover, 36 non-synonymous mutations were identified in the receptor-binding domain (RBD) of the spike protein with a low prevalence (<1%) across all genomes, of which only four could potentially enhance the binding of the SARS-CoV-2 spike protein to the human ACE2 receptor.
   Hausa: Haka kuma, an gano sauye guda 36 da ba a san su ba a cikin yankin muhimmin bangaren kwaƴar cuta (RBD) na furotin mai kauri tare da ƙarancin yaɗuwa (<1%) a duk nau'o'in kwayoyin gadon, waɗanda huɗu ne kawai za su iya haɓaka daurin SARS-CoV-2 furotin mai kauri ga mai karɓar ACE2 na ɗan'adam.

3. English: The F1 and the F2 measurement values, in hertz (Hz) were taken at three different points of the vowel spectrogram.
   Hausa: An ɗauki darajojin ma'aunin F1 da F2 a cikin hertz (Hz), a wurare daban-daban guda uku na spectrogram na wasali.

4. English: On a regional scale, we observed 91 unique and 8 shared haplotypes amongst the Sudan samples.
   Hausa: A kan sikelin yanki, mun lura da haɗe-haɗe na musamman guda 91 da kuma gadaddun kwayoyin gado guda 8 na iyaye ɗaya da aka raba a tsakanin samfuran Sudan.

5. English: Findings indicate that the general public does not have faith in government authorities, due to a lack of communication.
   Hausa: Bincike ya nuna cewa jama'a ba su da amana ga hukumomin gwamnati, saboda rashin sadarwa.

Now generate the Hausa translation of the following English 
text:
\end{lstlisting}
\end{tcolorbox}

\begin{tcolorbox}[enhanced, breakable, colback=metricsbg,
  colframe=white, arc=2mm, boxrule=0pt, width=\textwidth]
\begin{lstlisting}[basicstyle=\ttfamily\footnotesize, breaklines=true]
Human input:

The novel Coronavirus disease (COVID-19), caused by the severe acute respiratory syndrome coronavirus-2 (SARS-CoV-2), in Africa is characterised by a more substantial proportion of asymptomatic (or mildly symptomatic) individuals thought to be playing a role in the spread of the infection.
\end{lstlisting}
\end{tcolorbox}

\caption{Example of a generated prompt for English→Hausa translation using sentence-level template 5 (Prompt Template~\ref{prompt:sent_template5}) with glossary terms and 5-shot examples. The system input combines the template instruction with domain-matched examples and terminology, followed by the source text to be translated.}
\label{prompt:prompt_example}
\end{promptexample*}

\begin{table*}[t]
\centering
\small
\renewcommand{\arraystretch}{1.05}
\begin{tabular}{lrrrrrrrrrrrrr}
\toprule
\textbf{Model} & \multicolumn{6}{c}{\textbf{eng$\to$}} & \multicolumn{6}{c}{\textbf{$\to$eng}} & \\
\cmidrule(lr){2-7}\cmidrule(lr){8-13}
   & \texttt{amh} & \texttt{hau} & \texttt{lug} & \texttt{nso} & \texttt{yor} & \texttt{zul} & \texttt{amh} & \texttt{hau} & \texttt{lug} & \texttt{nso} & \texttt{yor} & \texttt{zul} & \textbf{Avg} \\
\midrule
\multicolumn{14}{l}{\textit{Zero-shot}} \\
\midrule
M2M100\textunderscore{}1.2B & 11.9 & 30.6 & 6.1 & 15.1 & \emph{13.3} & 30.3 & 36.6 & 39.7 & 20.3 & 18.9 & 15.7 & 45.8 & 23.7 \\
M2M100\textunderscore{}418M & 6.2 & 14.8 & 5.6 & 7.6 & 8.6 & 16.9 & 27.7 & 27.5 & 16.4 & 17.6 & 15.2 & 33.7 & 16.5 \\
NLLB\textunderscore{}1.3B & \textbf{37.4} & \textbf{59.4} & \textbf{43.2} & \textbf{51.4} & 13.0 & \textbf{60.1} & \textbf{57.6} & \textbf{57.5} & \textbf{49.5} & \textbf{58.5} & \textbf{46.0} & \textbf{63.2} & \textbf{49.7} \\
NLLB\textunderscore{}600M & 37.1 & 56.2 & 40.4 & 50.5 & 12.3 & 60.1 & 55.1 & 56.2 & 47.2 & 56.1 & 43.6 & 61.2 & 48.0 \\
\midrule
AfriqueLlama-8B & 23.6 & 31.0 & \emph{17.7} & 23.2 & 13.0 & 31.5 & 39.4 & 27.5 & 20.0 & 21.8 & 22.3 & 28.9 & 25.0 \\
AfriqueQwen-8B & 5.8 & 11.2 & 8.2 & 10.4 & 8.5 & 10.2 & 6.2 & 14.3 & 12.9 & 12.4 & 13.6 & 21.4 & 11.3 \\
Gemma2-9B & 8.6 & 30.6 & 12.1 & 18.5 & 10.2 & 20.5 & 39.3 & 45.9 & \emph{29.7} & 32.0 & 23.6 & 43.8 & 26.2 \\
Llama3-8B & 7.9 & 29.0 & 13.0 & 12.4 & 8.5 & 15.5 & 25.5 & 43.3 & 22.7 & 19.5 & 21.9 & 31.5 & 20.9 \\
Tiny-Aya-E & 27.4 & \emph{52.5} & 10.7 & 11.5 & \textbf{14.5} & \emph{44.8} & 47.9 & 45.7 & 20.1 & 17.3 & 28.5 & 49.6 & 30.9 \\
Tiny-Aya-G & \emph{28.1} & 51.6 & 10.2 & 11.7 & 14.5 & 44.5 & 47.8 & 46.2 & 19.9 & 17.2 & 28.0 & \emph{49.9} & 30.8 \\
TranslateGemma-12B & 18.8 & 38.3 & 17.3 & \emph{23.7} & 13.5 & 39.4 & \emph{51.0} & \emph{53.5} & 27.5 & \emph{41.4} & \emph{34.3} & 35.7 & \emph{32.9} \\
\midrule\midrule
\multicolumn{14}{l}{\textit{Fine-tuned}} \\
\midrule
M2M100\textunderscore{}1.2B & 39.0 & 63.1 & 49.5 & 62.4 & 16.6 & 58.4 & 46.9 & 55.2 & 48.0 & 57.0 & 43.3 & 52.2 & 49.3 \\
M2M100\textunderscore{}418M & 38.8 & 62.9 & 48.6 & 61.1 & 16.6 & 55.8 & 50.7 & 54.3 & 46.8 & 53.9 & 45.1 & 55.7 & 49.2 \\
NLLB\textunderscore{}1.3B & \textbf{45.5} & \textbf{67.7} & \textbf{51.5} & \textbf{64.3} & \emph{17.7} & \textbf{67.2} & \textbf{63.1} & \textbf{64.3} & \textbf{55.4} & \textbf{65.8} & \textbf{53.3} & \textbf{68.6} & \textbf{57.0} \\
NLLB\textunderscore{}600M & 44.4 & 66.9 & 50.6 & 64.2 & 17.6 & 66.3 & 60.1 & 62.1 & 53.2 & 63.2 & 50.7 & 66.6 & 55.5 \\
\midrule
AfriqueLlama-8B & 36.1 & 59.8 & 45.5 & 54.9 & 20.9 & 53.3 & \emph{59.3} & 56.6 & 47.0 & 58.6 & \emph{50.7} & 62.4 & 50.4 \\
AfriqueQwen-8B & 27.7 & 59.2 & 39.1 & 59.2 & 13.4 & 35.5 & 40.4 & 34.6 & 24.0 & 29.9 & 35.2 & 35.7 & 36.1 \\
Gemma2-9B & \emph{39.7} & 58.8 & \emph{50.1} & \emph{61.6} & \textbf{41.6} & 50.2 & 58.6 & \emph{62.7} & \emph{54.3} & \emph{62.5} & 50.7 & \emph{65.3} & \emph{54.7} \\
Llama3-8B & 32.9 & 60.3 & 49.3 & 59.0 & 39.9 & 52.4 & 53.0 & 60.3 & 52.6 & 60.2 & 48.8 & 61.2 & 52.5 \\
Tiny-Aya-E & 37.8 & \emph{62.0} & 46.9 & 57.7 & 39.5 & 56.1 & 58.5 & 61.6 & 50.4 & 57.1 & 46.2 & 64.7 & 53.2 \\
Tiny-Aya-G & 38.1 & 60.9 & 47.3 & 57.6 & 38.3 & \emph{56.2} & 56.9 & 60.1 & 49.2 & 55.9 & 48.2 & 63.4 & 52.7 \\
\bottomrule
\end{tabular}
\caption{Sentence-level zero-shot vs. fine-tuned performance (chrF). The four seq2seq models (M2M100-418M, NLLB-600M, M2M100-1.2B, NLLB-1.3B) and six open-source LLMs appear in both major groups; the \textit{Zero-shot} group additionally includes TranslateGemma-12B (evaluated but not fine-tuned). LLMs in the \textit{Fine-tuned} group use per-pair LoRA $r{=}64$. Models are alphabetically sorted within each sub-block. \textbf{Bold}: column maximum within the major group (\textit{Zero-shot} or \textit{Fine-tuned}); \emph{italic}: column maximum within the sub-block (seq2seq or LLM).}
\label{tab:main_zs_ft_chrf}
\end{table*}

\begin{table*}[t]
\centering
\small
\renewcommand{\arraystretch}{1.05}
\begin{tabular}{lrrrrrrrrrrrrr}
\toprule
\textbf{Model} & \multicolumn{6}{c}{\textbf{eng$\to$}} & \multicolumn{6}{c}{\textbf{$\to$eng}} & \\
\cmidrule(lr){2-7}\cmidrule(lr){8-13}
   & \texttt{amh} & \texttt{hau} & \texttt{lug} & \texttt{nso} & \texttt{yor} & \texttt{zul} & \texttt{amh} & \texttt{hau} & \texttt{lug} & \texttt{nso} & \texttt{yor} & \texttt{zul} & \textbf{Avg} \\
\midrule
\multicolumn{14}{l}{\textit{Zero-shot}} \\
\midrule
Best open & \emph{28.1} & \emph{52.5} & \emph{17.7} & \emph{23.7} & \emph{14.5} & \emph{44.8} & \emph{51.0} & \emph{53.5} & \emph{29.7} & \emph{41.4} & \emph{34.3} & \emph{49.9} & \emph{36.7} \\
\midrule
Gemini-1.5 & 36.4 & 53.9 & 37.3 & 46.9 & 15.2 & 49.5 & 58.5 & 56.7 & 47.6 & 52.1 & 42.0 & 60.3 & 46.4 \\
Gemini-3.1 & 35.7 & 50.3 & 36.4 & 45.9 & 15.6 & 45.4 & 47.5 & 54.1 & 44.7 & 51.1 & 42.2 & 60.3 & 44.1 \\
GPT-4o & 21.1 & 58.0 & 41.9 & 47.3 & 16.3 & 58.5 & 53.0 & 60.7 & 51.1 & 58.5 & 49.0 & 65.8 & 48.4 \\
GPT-5.4 & \textbf{37.6} & \textbf{58.4} & \textbf{48.9} & \textbf{53.9} & \textbf{16.3} & \textbf{62.8} & \textbf{63.5} & \textbf{64.9} & \textbf{55.4} & \textbf{63.0} & \textbf{53.8} & \textbf{69.4} & \textbf{54.0} \\
\midrule\midrule
\multicolumn{14}{l}{\textit{template5, 10-shot ICL}} \\
\midrule
Gemini-1.5 & 39.5 & 57.7 & 41.0 & 50.9 & 22.2 & 49.5 & 59.7 & 59.1 & 49.1 & 55.6 & 46.7 & 62.7 & 49.5 \\
Gemini-3.1 & 41.6 & 61.1 & 49.5 & 55.9 & 22.9 & 61.8 & 63.1 & 64.1 & 54.4 & 63.0 & 54.6 & 68.3 & 55.0 \\
GPT-4o & 31.1 & 60.7 & 47.0 & 53.4 & 21.2 & 60.5 & 53.4 & 61.1 & 51.8 & 59.4 & 50.5 & 66.0 & 51.3 \\
GPT-5.4 & \textbf{43.8} & \textbf{63.0} & \textbf{56.4} & \textbf{59.5} & \textbf{28.1} & \textbf{68.9} & \textbf{65.8} & \textbf{66.3} & \textbf{58.0} & \textbf{65.8} & \textbf{57.0} & \textbf{70.3} & \textbf{58.6} \\
\midrule
AfriqueLlama-8B & 2.7 & 9.4 & 8.9 & 10.6 & 6.3 & 9.8 & 1.1 & 7.8 & 6.8 & 7.4 & 7.1 & 9.6 & 7.3 \\
AfriqueQwen-8B & 6.7 & 16.4 & 11.3 & 12.5 & 7.5 & 14.7 & 15.6 & 13.7 & 10.5 & 9.7 & 11.5 & 14.0 & 12.0 \\
Gemma2-9B & 16.0 & 44.6 & 16.7 & 28.1 & 17.4 & 30.4 & 45.5 & \emph{52.7} & \emph{38.3} & \emph{42.1} & 35.2 & 52.3 & \emph{34.9} \\
Llama3-8B & 8.8 & 34.6 & \emph{25.5} & \emph{28.9} & \emph{21.5} & 25.8 & 30.3 & 43.6 & 30.9 & 22.0 & 27.1 & 33.5 & 27.7 \\
Tiny-Aya-E & 27.3 & \emph{51.6} & 8.5 & 8.0 & 14.7 & \emph{47.1} & \emph{51.0} & 51.9 & 26.2 & 24.2 & \emph{42.5} & \emph{55.3} & 34.0 \\
Tiny-Aya-G & \emph{27.6} & 51.3 & 8.2 & 8.0 & 14.8 & 46.9 & 50.7 & 51.2 & 25.6 & 23.9 & 42.0 & 55.0 & 33.8 \\
TranslateGemma-12B & 6.6 & 25.9 & 15.5 & 19.0 & 11.9 & 26.9 & 0.5 & 10.5 & 7.8 & 7.5 & 3.6 & 11.1 & 12.2 \\
\bottomrule
\end{tabular}
\caption{Open vs. closed-source LLMs at sentence-level in-context learning (chrF), all rows at the uniform \texttt{template5\textunderscore{}10shot} configuration (see \Cref{tab:template_shot_ablation_chrf} for the template/shot ablation that selected this config). \textit{Best open} in the zero-shot block is the per-pair maximum across the seven open LLMs in \Cref{tab:main_zs_ft_chrf}. AfriqueLlama-8B, AfriqueQwen-8B, and TranslateGemma-12B degenerate at this prompt (visible as COMET~$\gg$~chrF in their rows); see \Cref{tab:prompt_engineering_open} for their vanilla-prompt baseline. \textbf{Bold}: best performance across \textit{zero-shot} and \textit{ICL}; \emph{italic}: best performance within open or closed model group.}
\label{tab:open_vs_closed_icl_chrf}
\end{table*}

\begin{table*}[t]
\centering
\small
\renewcommand{\arraystretch}{1.05}
\begin{tabular}{lrrrrrrrrrrrrr}
\toprule
\textbf{Model} & \multicolumn{6}{c}{\textbf{eng$\to$}} & \multicolumn{6}{c}{\textbf{$\to$eng}} & \\
\cmidrule(lr){2-7}\cmidrule(lr){8-13}
   & \texttt{amh} & \texttt{hau} & \texttt{lug} & \texttt{nso} & \texttt{yor} & \texttt{zul} & \texttt{amh} & \texttt{hau} & \texttt{lug} & \texttt{nso} & \texttt{yor} & \texttt{zul} & \textbf{Avg} \\
\midrule
\multicolumn{14}{l}{\textit{Zero-shot}} \\
\midrule
Gemini-1.5 & 17.1 & 41.3 & 28.3 & 32.0 & 9.0 & 35.1 & 69.4 & 68.9 & 60.6 & 66.5 & 59.8 & 73.3 & 46.8 \\
Gemini-3.1 & \textbf{43.5} & 61.2 & 48.6 & 50.7 & 14.8 & 63.8 & 68.1 & 66.6 & 64.4 & 68.2 & 63.9 & 75.0 & 57.4 \\
GPT-4o & 12.0 & 66.0 & 33.3 & 54.3 & 18.4 & 61.2 & 62.3 & 71.5 & 64.5 & 70.7 & 63.0 & \textbf{76.0} & 54.4 \\
GPT-5.4 & 26.0 & \textbf{66.1} & \textbf{59.0} & \textbf{63.3} & \textbf{19.8} & \textbf{66.1} & \textbf{70.6} & \textbf{72.7} & \textbf{66.9} & \textbf{73.0} & \textbf{65.8} & 75.8 & \textbf{60.4} \\
\midrule
AfriqueLlama-8B & 6.5 & 36.8 & 10.6 & \emph{24.1} & \emph{9.7} & 37.3 & 21.9 & 56.6 & 29.8 & 40.0 & 29.9 & 60.6 & 30.3 \\
AfriqueQwen-8B & 3.8 & 10.3 & 7.7 & 8.3 & 6.9 & 12.7 & 1.0 & 15.2 & 9.4 & 15.8 & 9.1 & 16.7 & 9.8 \\
Gemma2-9B & 4.4 & 29.9 & 7.6 & 11.0 & 2.7 & 13.2 & 25.7 & 42.4 & 27.6 & 28.1 & 22.8 & 34.6 & 20.8 \\
Llama3-8B & 2.6 & 11.0 & 9.0 & 7.9 & 2.2 & 9.6 & 26.7 & 49.9 & 32.3 & 31.0 & 29.0 & 34.8 & 20.5 \\
Tiny-Aya-E & \emph{20.3} & 39.4 & 7.2 & 8.3 & 8.1 & 39.0 & 57.1 & 59.9 & 22.3 & 18.7 & 33.8 & 65.3 & 31.6 \\
Tiny-Aya-G & 19.0 & \emph{40.8} & 7.7 & 8.0 & 8.0 & \emph{41.0} & 56.3 & 60.5 & 21.0 & 17.6 & 35.3 & 65.4 & 31.7 \\
TranslateGemma-12B & 20.1 & 34.2 & \emph{12.8} & 18.5 & 8.3 & 33.9 & \emph{58.6} & \emph{61.3} & \emph{48.9} & \emph{52.4} & \emph{44.0} & \emph{65.6} & \emph{38.2} \\
\midrule\midrule
\multicolumn{14}{l}{\textit{1-shot}} \\
\midrule
Gemini-1.5 & 17.1 & 41.9 & 29.0 & 32.3 & 9.6 & 35.5 & 70.1 & 70.3 & 63.3 & 68.4 & 61.4 & 73.7 & 47.7 \\
Gemini-3.1 & \textbf{45.1} & 64.3 & 50.8 & 55.8 & 17.5 & 62.9 & 70.9 & 65.5 & 65.3 & 69.6 & 64.9 & 75.7 & 59.0 \\
GPT-4o & 11.1 & \textbf{66.9} & 43.0 & 61.0 & 12.8 & 63.1 & 64.1 & 72.1 & 65.3 & 71.6 & 64.2 & 76.0 & 55.9 \\
GPT-5.4 & 42.8 & 66.8 & \textbf{59.5} & \textbf{64.9} & \textbf{28.7} & \textbf{68.6} & \textbf{72.3} & \textbf{74.0} & \textbf{68.1} & \textbf{74.3} & \textbf{66.7} & \textbf{77.0} & \textbf{63.6} \\
\midrule
AfriqueLlama-8B & 6.6 & \emph{39.2} & 11.3 & 19.6 & 6.8 & 36.8 & 16.6 & 56.5 & 35.6 & 35.5 & 34.4 & 62.3 & 30.1 \\
AfriqueQwen-8B & 5.5 & 11.4 & 10.2 & 7.3 & 4.7 & 6.6 & 5.6 & 27.3 & 15.3 & 25.3 & 26.5 & 35.1 & 15.1 \\
Gemma2-9B & 3.8 & 30.9 & 8.2 & 10.4 & 2.7 & 12.6 & 14.0 & 45.7 & 27.8 & 24.3 & 18.9 & 40.6 & 20.0 \\
Llama3-8B & 2.4 & 13.0 & 8.0 & 7.3 & 3.4 & 10.4 & 22.7 & 44.9 & 32.3 & 31.0 & 28.4 & 35.5 & 20.0 \\
Tiny-Aya-E & 12.8 & 34.8 & 6.7 & 7.8 & 5.5 & 35.8 & 24.5 & 37.2 & 16.2 & 15.8 & 12.1 & 36.1 & 20.4 \\
Tiny-Aya-G & 14.2 & 33.8 & 7.3 & 7.6 & 6.3 & \emph{38.4} & 13.6 & 43.1 & 13.8 & 14.4 & 11.8 & 32.3 & 19.7 \\
TranslateGemma-12B & \emph{20.4} & 33.3 & \emph{15.6} & \emph{21.5} & \emph{7.0} & 32.7 & \emph{58.7} & \emph{63.1} & \emph{50.1} & \emph{52.2} & \emph{40.4} & \emph{66.3} & \emph{38.4} \\
\bottomrule
\end{tabular}
\caption{Document-level zero-shot vs. 1-shot in-context learning (chrF). Two major groups; each has closed-model and open-LLM sub-blocks. Models alphabetically sorted within each sub-block. \textbf{Bold}: column maximum within the major group (\textit{Zero-shot} or \textit{1-shot}); \emph{italic}: column maximum within the sub-block (closed or open).}
\label{tab:doc_chrf}
\end{table*}

\begin{table*}[t]
\centering
\small
\renewcommand{\arraystretch}{1.05}
\begin{tabular}{llrrrrrrrrrrrr}
\toprule
\textbf{Setup} & \textbf{Model} & \multicolumn{6}{c}{\textbf{Eng$\to$}} & \multicolumn{6}{c}{\textbf{$\to$Eng}} \\
\cmidrule(lr){3-8}\cmidrule(lr){9-14}
  &   & \texttt{amh} & \texttt{hau} & \texttt{lug} & \texttt{nso} & \texttt{yor} & \texttt{zul} & \texttt{amh} & \texttt{hau} & \texttt{lug} & \texttt{nso} & \texttt{yor} & \texttt{zul} \\
\midrule
\multirow{6}{*}{\textit{LoRA $r{=}4$}} & AfriqueLlama-8B & 36.2 & 60.9 & 41.8 & \emph{57.4} & 25.5 & \textbf{56.5} & \textbf{60.4} & 61.3 & 42.5 & 60.5 & \emph{50.4} & 64.0 \\
 & AfriqueQwen-8B & 32.9 & 51.0 & 41.0 & 53.8 & 13.5 & 40.0 & 47.6 & 56.2 & 35.3 & 35.8 & 48.7 & 29.7 \\
 & Gemma2-9B & 36.0 & 57.2 & \emph{44.9} & 57.3 & \emph{39.3} & 48.9 & 57.8 & \textbf{62.7} & \emph{52.7} & \emph{61.1} & 49.7 & 64.6 \\
 & Llama3-8B & 25.3 & 56.6 & 42.6 & 51.3 & 36.7 & 43.5 & 48.5 & 58.7 & 49.6 & 55.3 & 45.4 & 56.5 \\
 & Tiny-Aya-E & \emph{37.3} & 62.2 & 39.9 & 51.0 & 37.5 & 56.4 & 58.0 & 61.8 & 45.6 & 51.4 & 44.0 & \emph{64.9} \\
 & Tiny-Aya-G & 36.8 & \textbf{62.2} & 40.3 & 51.1 & 37.4 & 56.4 & 57.6 & 61.5 & 45.8 & 51.1 & 44.3 & 64.8 \\
\midrule
\multirow{6}{*}{\textit{LoRA $r{=}64$}} & AfriqueLlama-8B & 36.1 & 59.8 & 45.5 & 54.9 & 20.9 & 53.3 & \emph{59.3} & 56.6 & 47.0 & 58.6 & \textbf{50.7} & 62.4 \\
 & AfriqueQwen-8B & 27.7 & 59.2 & 39.1 & 59.2 & 13.4 & 35.5 & 40.4 & 34.6 & 24.0 & 29.9 & 35.2 & 35.7 \\
 & Gemma2-9B & \textbf{39.7} & 58.8 & \textbf{50.1} & \textbf{61.6} & \textbf{41.6} & 50.2 & 58.6 & \emph{62.7} & \textbf{54.3} & \textbf{62.5} & 50.7 & \textbf{65.3} \\
 & Llama3-8B & 32.9 & 60.3 & 49.3 & 59.0 & 39.9 & 52.4 & 53.0 & 60.3 & 52.6 & 60.2 & 48.8 & 61.2 \\
 & Tiny-Aya-E & 37.8 & \emph{62.0} & 46.9 & 57.7 & 39.5 & 56.1 & 58.5 & 61.6 & 50.4 & 57.1 & 46.2 & 64.7 \\
 & Tiny-Aya-G & 38.1 & 60.9 & 47.3 & 57.6 & 38.3 & \emph{56.2} & 56.9 & 60.1 & 49.2 & 55.9 & 48.2 & 63.4 \\
\bottomrule
\end{tabular}
\caption{LoRA fine-tuning of six open-source LLMs at two ranks (chrF). Each cell is one (language, direction) pair; each model is fine-tuned bilingually per pair for 10 epochs with early stopping. Comparing $r{=}4$ (low-rank, parameter-efficient) against $r{=}64$ (the main rank used elsewhere in the paper) isolates the effect of adapter capacity. \textbf{Bold}: best score for that (language, direction) pair across both ranks; \emph{italic}: best score for that pair within a single rank block.}
\label{tab:lora_chrf}
\end{table*}

\begin{table*}[t]
\centering
\small
\renewcommand{\arraystretch}{1.05}
\begin{tabular}{lllrrrrrrr}
\toprule
\textbf{Model} & \textbf{Version} & \textbf{Setting} & \multicolumn{6}{c}{\textbf{eng$\to$}} & \\
\cmidrule(lr){4-9}
   &    &    & \texttt{amh} & \texttt{hau} & \texttt{lug} & \texttt{nso} & \texttt{yor} & \texttt{zul} & \textbf{Avg} \\
\midrule
\multicolumn{10}{l}{\textit{Reported, vanilla prompts ($T{=}0$): ZS vs ICL (5-shot)}} \\
\midrule
\multirow{3}{*}{Gemini} & \multirow{1}{*}{1.5} & ZS & 36.4 & 53.9 & 37.3 & 46.9 & 15.2 & 49.5 & 39.9 \\
\cmidrule(lr){2-10}
 & \multirow{2}{*}{3.1} & ZS & 35.7 & 50.3 & 36.4 & 45.9 & 15.6 & 45.4 & 38.2 \\
 &  & ICL (5-shot) & \textbf{\emph{39.9}} & \textbf{\emph{60.0}} & \emph{47.5} & \emph{53.5} & \textbf{\emph{23.7}} & \emph{60.8} & \emph{47.5} \\
\cmidrule(lr){1-10}
\multirow{3}{*}{GPT} & \multirow{1}{*}{4o} & ZS & 21.1 & 58.0 & 41.9 & 47.3 & 16.3 & 58.5 & 40.5 \\
\cmidrule(lr){2-10}
 & \multirow{2}{*}{5.4} & ZS & 37.6 & 58.4 & 48.9 & 53.9 & 16.3 & 62.8 & 46.3 \\
 &  & ICL (5-shot) & \emph{39.5} & \emph{59.6} & \textbf{\emph{50.2}} & \textbf{\emph{55.5}} & \emph{23.6} & \textbf{\emph{63.4}} & \textbf{\emph{48.6}} \\
\midrule\midrule
\multicolumn{10}{l}{\textit{Reported, templated prompts ($T{=}0$): ZS (avg \texttt{template\{1,2,4\}\textunderscore{}0shot}) vs ICL (best 10-shot)}} \\
\midrule
\multirow{4}{*}{Gemini} & \multirow{2}{*}{1.5} & ZS (avg \texttt{template\{1,2,4\}}) & 36.9 & 54.8 & 38.7 & 47.7 & 16.3 & 49.2 & 40.6 \\
 &  & ICL (10-shot, template5) & \emph{39.5} & \emph{57.7} & \emph{41.0} & \emph{50.9} & \emph{22.2} & \emph{49.5} & \emph{43.5} \\
\cmidrule(lr){2-10}
 & \multirow{2}{*}{3.1} & ZS (avg \texttt{template\{1,2,4\}}) & 38.9 & 58.1 & 47.0 & 53.1 & 18.1 & 60.8 & 46.0 \\
 &  & ICL (10-shot, template5) & \emph{41.6} & \emph{61.1} & \emph{49.5} & \emph{55.9} & \emph{22.9} & \emph{61.8} & \emph{48.8} \\
\cmidrule(lr){1-10}
\multirow{4}{*}{GPT} & \multirow{2}{*}{4o} & ZS (avg \texttt{template\{1,2,4\}}) & 27.5 & 59.2 & 43.1 & 50.5 & 17.2 & 59.8 & 42.9 \\
 &  & ICL (10-shot, template3) & \emph{30.0} & \emph{60.8} & \emph{45.2} & \emph{53.1} & \emph{19.9} & \emph{60.9} & \emph{45.0} \\
\cmidrule(lr){2-10}
 & \multirow{2}{*}{5.4} & ZS (avg \texttt{template\{1,2,4\}}) & 41.8 & 60.8 & 54.7 & 57.2 & 19.5 & 62.6 & 49.4 \\
 &  & ICL (10-shot, template5) & \textbf{\emph{43.8}} & \textbf{\emph{63.0}} & \textbf{\emph{56.4}} & \textbf{\emph{59.5}} & \textbf{\emph{28.1}} & \textbf{\emph{68.9}} & \textbf{\emph{53.3}} \\
\midrule\midrule
\multicolumn{10}{l}{\textit{Retrieval ablation: templated vs. semantic-similarity, $T{=}0$}} \\
\midrule
\multirow{4}{*}{Gemini} & \multirow{2}{*}{1.5} & 10-shot (templated, template5) & 39.5 & 57.7 & 41.0 & 50.9 & 22.2 & 49.5 & 43.5 \\
 &  & 10-shot (semantic similarity) & \emph{40.9} & \emph{58.1} & \emph{46.0} & \emph{54.9} & \emph{24.6} & \emph{52.0} & \emph{46.1} \\
\cmidrule(lr){2-10}
 & \multirow{2}{*}{3.1} & 10-shot (templated, template5) & 41.6 & 61.1 & 49.5 & 55.9 & 22.9 & 61.8 & 48.8 \\
 &  & 10-shot (semantic similarity) & \emph{43.6} & \emph{62.9} & \emph{52.5} & \emph{58.9} & \textbf{\emph{30.5}} & \emph{62.7} & \emph{51.9} \\
\cmidrule(lr){1-10}
\multirow{4}{*}{GPT} & \multirow{2}{*}{4o} & 10-shot (templated, template3) & \emph{30.0} & 60.8 & 45.2 & 53.1 & 19.9 & 60.9 & 45.0 \\
 &  & 10-shot (semantic similarity) & 28.7 & \emph{62.8} & \emph{48.2} & \emph{57.5} & \emph{20.4} & \emph{61.9} & \emph{46.6} \\
\cmidrule(lr){2-10}
 & \multirow{2}{*}{5.4} & 10-shot (templated, template5) & \textbf{\emph{43.8}} & \textbf{\emph{63.0}} & \textbf{\emph{56.4}} & 59.5 & 28.1 & \textbf{\emph{68.9}} & \textbf{\emph{53.3}} \\
 &  & 10-shot (semantic similarity) & 42.2 & 62.1 & 53.5 & \textbf{\emph{59.7}} & \emph{30.5} & 64.2 & 52.0 \\
\midrule\midrule
\multicolumn{10}{l}{\textit{Temperature ablation ($T{=}0$ vs $T{=}0.6$, all templated)}} \\
\midrule
\multirow{8}{*}{Gemini} & \multirow{4}{*}{1.5} & ZS ($T{=}0$) & 36.9 & 54.8 & 38.7 & 47.7 & 16.3 & 49.2 & 40.6 \\
 &  & ZS ($T{=}0.6$) & 36.6 & 54.6 & 39.1 & 47.6 & 16.6 & 49.5 & 40.7 \\
 &  & 10-shot (template5) $T{=}0$ & \emph{39.5} & \emph{57.7} & 41.0 & 50.9 & 22.2 & 49.5 & 43.5 \\
 &  & 10-shot (template5) $T{=}0.6$ & 39.1 & 57.2 & \emph{42.9} & \emph{51.8} & \emph{22.7} & \emph{50.1} & \emph{44.0} \\
\cmidrule(lr){2-10}
 & \multirow{4}{*}{3.1} & ZS ($T{=}0$) & 38.9 & 58.1 & 47.0 & 53.1 & 18.1 & 60.8 & 46.0 \\
 &  & ZS ($T{=}0.6$) & 38.5 & 57.4 & 45.8 & 52.5 & 18.0 & 59.8 & 45.3 \\
 &  & 10-shot (template5) $T{=}0$ & \emph{41.6} & \emph{61.1} & \emph{49.5} & \emph{55.9} & \emph{22.9} & \emph{61.8} & \emph{48.8} \\
 &  & 10-shot (template5) $T{=}0.6$ & 41.0 & 60.5 & 49.0 & 55.2 & 22.8 & 61.0 & 48.3 \\
\cmidrule(lr){1-10}
\multirow{8}{*}{GPT} & \multirow{4}{*}{4o} & ZS ($T{=}0$) & 27.5 & 59.2 & 43.1 & 50.5 & 17.2 & 59.8 & 42.9 \\
 &  & ZS ($T{=}0.6$) & 22.0 & 58.5 & 41.8 & 47.8 & 17.4 & 58.7 & 41.0 \\
 &  & 10-shot (template3) $T{=}0$ & \emph{30.0} & \emph{60.8} & \emph{45.2} & \emph{53.1} & \emph{19.9} & \emph{60.9} & \emph{45.0} \\
 &  & 10-shot (template3) $T{=}0.6$ & 25.0 & 60.2 & 44.6 & 49.3 & 18.7 & 60.0 & 43.0 \\
\cmidrule(lr){2-10}
 & \multirow{4}{*}{5.4} & ZS ($T{=}0$) & 41.8 & 60.8 & 54.7 & 57.2 & 19.5 & 62.6 & 49.4 \\
 &  & ZS ($T{=}0.6$) & 41.7 & 59.9 & 54.6 & 57.3 & 19.0 & 64.6 & 49.5 \\
 &  & 10-shot (template5) $T{=}0$ & \textbf{\emph{43.8}} & \textbf{\emph{63.0}} & 56.4 & 59.5 & \textbf{\emph{28.1}} & 68.9 & \textbf{\emph{53.3}} \\
 &  & 10-shot (template5) $T{=}0.6$ & 43.7 & 62.7 & \textbf{\emph{56.5}} & \textbf{\emph{59.5}} & 27.8 & \textbf{\emph{68.9}} & 53.2 \\
\bottomrule
\end{tabular}
\caption{Closed-API ablations on \texttt{eng}$\to$\texttt{xxx} (chrF) for the four closed models (older: GPT-4o, Gemini-1.5-Flash; newer: GPT-5.4, Gemini-3.1-Flash-Lite). Four blocks: vanilla ZS vs ICL; templated ZS vs ICL; retrieval (templated vs semantic 10-shot); and temperature ($T{=}0$ vs $T{=}0.6$, all templated). \textbf{Bold}: per-block column maximum. \emph{Italic}: per-version best across that model's rows. See \Cref{app:closed_ablations_notes} for the full setup and the rationale behind each block.}
\label{tab:closed_ablations_chrf}
\end{table*}

\onecolumn
\begingroup
\footnotesize
\setlength{\tabcolsep}{4pt}
\renewcommand{\arraystretch}{1.05}
\begin{longtable}{llrrrrrrrrrrrrr}
\toprule
\textbf{Model} & \textbf{Config} & \multicolumn{6}{c}{\textbf{eng$\to$}} & \multicolumn{6}{c}{\textbf{$\to$eng}} & \\
\cmidrule(lr){3-8}\cmidrule(lr){9-14}
  &   & \textbf{\texttt{amh}} & \textbf{\texttt{hau}} & \textbf{\texttt{lug}} & \textbf{\texttt{nso}} & \textbf{\texttt{yor}} & \textbf{\texttt{zul}} & \textbf{\texttt{amh}} & \textbf{\texttt{hau}} & \textbf{\texttt{lug}} & \textbf{\texttt{nso}} & \textbf{\texttt{yor}} & \textbf{\texttt{zul}} & \textbf{Avg} \\
\midrule
\endfirsthead
\multicolumn{15}{l}{\textit{(continued)}} \\
\toprule
\textbf{Model} & \textbf{Config} & \multicolumn{6}{c}{\textbf{eng$\to$}} & \multicolumn{6}{c}{\textbf{$\to$eng}} & \\
\cmidrule(lr){3-8}\cmidrule(lr){9-14}
  &   & \textbf{\texttt{amh}} & \textbf{\texttt{hau}} & \textbf{\texttt{lug}} & \textbf{\texttt{nso}} & \textbf{\texttt{yor}} & \textbf{\texttt{zul}} & \textbf{\texttt{amh}} & \textbf{\texttt{hau}} & \textbf{\texttt{lug}} & \textbf{\texttt{nso}} & \textbf{\texttt{yor}} & \textbf{\texttt{zul}} & \textbf{Avg} \\
\midrule
\endhead
\midrule
\multicolumn{15}{r}{\textit{(continued on next page)}} \\
\endfoot
\bottomrule
\caption{Template/shot ablation at $T{=}0$ for the six reference models -- two open (Llama3-8B, Gemma2-9B-IT) and four closed models (older: GPT-4o, Gemini-1.5-Flash; newer: GPT-5.4, Gemini-3.1-Flash-Lite) (chrF). Each row is a (template, shot-count) configuration (template number first, shot count as subscript $s$); the underlined row \texttt{5}$_{s=10}$ is the configuration used throughout the paper. \textbf{Bold}: per-language column maximum within each model block. The bottom \textbf{avg} row of each block reports the per-language average across configurations. Shot counts are capped at 20; the 10$\to$20-shot gain is negligible.} \label{tab:template_shot_ablation_chrf} \\
\endlastfoot
\multirow{10}{*}{\rotatebox{90}{Gemini-1.5}} & 1 & 36.4 & 53.9 & 37.3 & 46.9 & 15.2 & 49.5 & 58.5 & 56.7 & 47.6 & 52.1 & 42.0 & 60.3 & 46.4 \\
 & 2 & 37.0 & 54.8 & 37.8 & 47.6 & 15.2 & 50.1 & 57.5 & 55.9 & 46.9 & 52.6 & 41.9 & 59.9 & 46.4 \\
 & 3$_{s=5}$ & 38.8 & 56.9 & 40.0 & 47.9 & 17.5 & 51.7 & 60.1 & 58.9 & 49.1 & 55.5 & 45.2 & 63.2 & 48.7 \\
 & 3$_{s=10}$ & 38.9 & 57.3 & 41.2 & 50.0 & 18.3 & 50.1 & 60.2 & 59.1 & 48.6 & 55.4 & 45.6 & 63.3 & 49.0 \\
 & 3$_{s=20}$ & 39.1 & 57.5 & 40.1 & 50.2 & 18.9 & \textbf{52.0} & \textbf{60.8} & 60.2 & \textbf{49.9} & 56.4 & 46.0 & \textbf{64.3} & 49.6 \\
 & 4 & 37.2 & 55.6 & 41.1 & 48.7 & 18.4 & 47.9 & 58.4 & 57.4 & 48.3 & 53.0 & 44.7 & 60.8 & 47.6 \\
 & 5$_{s=5}$ & 39.3 & 57.0 & \textbf{41.5} & 48.9 & 22.7 & 51.7 & 59.4 & 59.1 & 49.4 & 55.5 & 46.5 & 62.3 & 49.4 \\
 & \underline{5$_{s=10}$} & 39.5 & 57.7 & 41.0 & \textbf{50.9} & 22.2 & 49.5 & 59.7 & 59.1 & 49.1 & 55.6 & 46.7 & 62.7 & 49.5 \\
 & 5$_{s=20}$ & \textbf{39.5} & \textbf{57.8} & 41.1 & 50.8 & \textbf{23.8} & 49.5 & 60.2 & \textbf{60.2} & 49.8 & \textbf{56.5} & \textbf{47.4} & 63.6 & \textbf{50.0} \\
\cmidrule(l){2-15}
 & \textbf{avg} & 38.4 & 56.5 & 40.1 & 49.1 & 19.1 & 50.2 & 59.4 & 58.5 & 48.7 & 54.7 & 45.1 & 62.3 & 48.5 \\
\midrule
\multirow{10}{*}{\rotatebox{90}{GPT-4o}} & 1 & 27.1 & 58.5 & 42.5 & 50.3 & 16.1 & 59.6 & 53.0 & 60.7 & 51.1 & 58.5 & 49.0 & 65.8 & 49.3 \\
 & 2 & 26.9 & 59.5 & 42.3 & 50.2 & 16.1 & 60.2 & 53.5 & 61.1 & 51.5 & 59.2 & 49.3 & 66.6 & 49.7 \\
 & 3$_{s=5}$ & 29.4 & 60.7 & 44.8 & 52.0 & 16.6 & 60.9 & 54.0 & 62.1 & 52.5 & 60.1 & 49.8 & 67.5 & 50.9 \\
 & 3$_{s=10}$ & 30.0 & 60.8 & 45.2 & 53.1 & 19.9 & \textbf{60.9} & 54.1 & 62.2 & 52.5 & 60.3 & 50.3 & 67.5 & 51.4 \\
 & 3$_{s=20}$ & 30.1 & 61.1 & 45.5 & 53.5 & 20.0 & 60.8 & \textbf{55.2} & \textbf{63.2} & \textbf{53.1} & \textbf{60.7} & 50.9 & \textbf{68.2} & \textbf{51.9} \\
 & 4 & 28.6 & 59.7 & 44.6 & 50.9 & 19.5 & 59.6 & 52.6 & 59.9 & 50.7 & 58.1 & 49.6 & 64.8 & 49.9 \\
 & 5$_{s=5}$ & 30.3 & 60.5 & 46.3 & 52.4 & 19.9 & 60.4 & 53.2 & 60.9 & 51.5 & 59.3 & 50.2 & 66.0 & 50.9 \\
 & \underline{5$_{s=10}$} & \textbf{31.1} & 60.7 & 47.0 & 53.4 & 21.2 & 60.5 & 53.4 & 61.1 & 51.8 & 59.4 & 50.5 & 66.0 & 51.3 \\
 & 5$_{s=20}$ & 31.1 & \textbf{61.1} & \textbf{47.0} & \textbf{53.8} & \textbf{21.5} & 60.5 & 54.1 & 61.7 & 52.2 & 59.8 & \textbf{51.0} & 66.5 & 51.7 \\
\cmidrule(l){2-15}
 & \textbf{avg} & 29.4 & 60.3 & 45.0 & 52.2 & 19.0 & 60.4 & 53.7 & 61.4 & 51.9 & 59.5 & 50.1 & 66.5 & 50.8 \\
\midrule
\multirow{10}{*}{\rotatebox{90}{Gemini-3.1}} & 1 & 37.2 & 56.0 & 45.8 & 52.3 & 16.2 & 59.2 & 62.2 & 62.5 & 53.3 & 61.3 & 52.8 & 68.0 & 52.2 \\
 & 2 & 39.9 & 60.0 & 47.0 & 53.2 & 16.4 & 62.2 & 63.7 & 64.0 & 54.0 & 62.5 & 53.3 & 68.7 & 53.7 \\
 & 3$_{s=5}$ & 40.6 & 61.5 & 48.3 & 55.1 & 19.4 & \textbf{63.1} & 63.8 & 64.7 & 54.6 & 63.5 & 54.4 & 69.4 & 54.9 \\
 & 3$_{s=10}$ & 41.3 & 61.4 & 48.8 & 55.1 & 21.3 & 62.9 & 64.1 & 64.7 & 54.5 & 63.8 & 54.7 & 69.3 & 55.1 \\
 & 3$_{s=20}$ & 41.7 & \textbf{62.0} & 48.8 & 55.4 & 20.0 & 62.9 & \textbf{64.9} & \textbf{65.6} & \textbf{55.9} & \textbf{64.4} & \textbf{55.7} & \textbf{70.3} & \textbf{55.6} \\
 & 4 & 39.6 & 58.4 & 48.2 & 53.7 & 21.7 & 61.0 & 62.6 & 62.5 & 53.6 & 61.5 & 53.3 & 67.6 & 53.6 \\
 & 5$_{s=5}$ & 41.4 & 61.0 & 49.5 & 55.9 & 21.1 & 62.3 & 62.8 & 63.8 & 54.6 & 62.6 & 54.1 & 68.3 & 54.8 \\
 & \underline{5$_{s=10}$} & 41.6 & 61.1 & 49.5 & \textbf{55.9} & \textbf{22.9} & 61.8 & 63.1 & 64.1 & 54.4 & 63.0 & 54.6 & 68.3 & 55.0 \\
 & 5$_{s=20}$ & \textbf{42.1} & 61.3 & \textbf{50.0} & 55.6 & 21.6 & 61.8 & 63.9 & 64.8 & 55.4 & 63.7 & 55.5 & 68.9 & 55.4 \\
\cmidrule(l){2-15}
 & \textbf{avg} & 40.6 & 60.3 & 48.4 & 54.7 & 20.1 & 61.9 & 63.4 & 64.1 & 54.5 & 62.9 & 54.3 & 68.8 & 54.5 \\
\midrule
\multirow{10}{*}{\rotatebox{90}{GPT-5.4}} & 1 & 40.8 & 57.9 & 53.8 & 56.7 & 17.8 & 61.5 & 66.0 & 65.8 & 57.1 & 64.8 & 55.6 & 70.1 & 55.7 \\
 & 2 & 42.7 & 62.7 & 54.4 & 57.3 & 17.8 & 63.9 & 66.5 & 67.0 & 57.7 & 65.6 & 56.0 & 71.4 & 56.9 \\
 & 3$_{s=5}$ & 43.6 & 63.1 & 54.7 & 58.7 & 25.5 & 70.0 & 66.4 & 66.7 & 57.9 & 65.8 & 55.9 & 71.5 & 58.3 \\
 & 3$_{s=10}$ & 44.0 & \textbf{63.5} & 55.1 & 59.5 & \textbf{31.4} & \textbf{70.1} & 66.8 & 66.6 & 58.1 & 66.2 & 56.5 & 71.3 & 59.1 \\
 & 3$_{s=20}$ & 44.0 & 63.3 & 55.1 & 59.8 & 30.1 & 70.0 & \textbf{67.8} & \textbf{67.5} & 58.6 & \textbf{67.0} & 57.3 & \textbf{72.4} & \textbf{59.4} \\
 & 4 & 42.0 & 61.7 & 55.8 & 57.8 & 22.8 & 62.3 & 65.1 & 65.8 & 57.5 & 64.6 & 56.1 & 69.8 & 56.8 \\
 & 5$_{s=5}$ & 44.0 & 62.4 & 56.2 & 58.9 & 24.9 & 69.3 & 65.4 & 65.8 & 57.7 & 65.4 & 56.9 & 70.2 & 58.1 \\
 & \underline{5$_{s=10}$} & 43.8 & 63.0 & \textbf{56.4} & 59.5 & 28.1 & 68.9 & 65.8 & 66.3 & 58.0 & 65.8 & 57.0 & 70.3 & 58.6 \\
 & 5$_{s=20}$ & \textbf{44.2} & 62.8 & 56.2 & \textbf{59.9} & 27.3 & 68.6 & 66.4 & 66.7 & \textbf{58.6} & 66.4 & \textbf{57.8} & 70.6 & 58.8 \\
\cmidrule(l){2-15}
 & \textbf{avg} & 43.2 & 62.3 & 55.3 & 58.7 & 25.1 & 67.2 & 66.2 & 66.5 & 57.9 & 65.7 & 56.6 & 70.9 & 58.0 \\
\midrule
\multirow{10}{*}{\rotatebox{90}{Llama3-8B}} & 1 & 6.4 & 31.3 & 19.1 & 20.0 & 10.0 & 21.3 & 29.9 & 45.6 & 28.9 & 27.5 & 25.2 & \textbf{33.5} & 24.9 \\
 & 2 & 6.2 & 30.9 & 19.0 & 19.6 & 10.2 & 21.0 & 30.6 & 46.1 & 30.8 & \textbf{28.3} & 26.5 & 33.3 & 25.2 \\
 & 3$_{s=5}$ & 6.7 & 32.1 & 21.5 & 25.3 & 15.2 & 23.4 & 31.0 & 41.3 & 26.2 & 23.0 & 25.1 & 29.0 & 25.0 \\
 & 3$_{s=10}$ & 7.5 & 33.4 & 24.1 & 28.2 & 19.9 & 24.7 & 29.8 & 45.6 & 30.5 & 25.1 & 28.0 & 31.8 & 27.4 \\
 & 3$_{s=20}$ & 7.8 & 34.0 & 25.0 & 28.8 & 18.9 & 25.1 & 32.3 & 45.7 & \textbf{31.1} & 22.0 & 28.2 & 31.3 & 27.5 \\
 & 4 & 7.7 & 32.2 & 21.2 & 21.0 & 13.5 & 22.5 & \textbf{33.2} & 45.4 & 29.5 & 27.5 & 27.2 & 31.3 & 26.0 \\
 & 5$_{s=5}$ & 8.0 & 33.0 & 23.1 & 26.0 & 17.9 & 24.2 & 31.3 & 40.9 & 25.5 & 22.1 & 24.1 & 27.8 & 25.3 \\
 & \underline{5$_{s=10}$} & 8.8 & 34.6 & 25.5 & 28.9 & \textbf{21.5} & 25.8 & 30.3 & 43.6 & 30.9 & 22.0 & 27.1 & 33.5 & 27.7 \\
 & 5$_{s=20}$ & \textbf{8.9} & \textbf{34.8} & \textbf{26.3} & \textbf{29.6} & 20.6 & \textbf{26.0} & 31.6 & \textbf{47.0} & 30.7 & 22.8 & \textbf{28.4} & 30.7 & \textbf{28.1} \\
\cmidrule(l){2-15}
 & \textbf{avg} & 7.6 & 32.9 & 22.8 & 25.3 & 16.4 & 23.8 & 31.1 & 44.6 & 29.3 & 24.5 & 26.6 & 31.3 & 26.4 \\
\midrule
\multirow{10}{*}{\rotatebox{90}{Gemma2-9B}} & 1 & 12.6 & 40.3 & 15.0 & 21.8 & 9.4 & 26.4 & 42.4 & 49.3 & 33.7 & 35.9 & 29.2 & 47.3 & 30.3 \\
 & 2 & 12.4 & 39.4 & 14.4 & 21.6 & 9.3 & 26.1 & 42.4 & 48.9 & 35.0 & 36.4 & 29.9 & 47.4 & 30.3 \\
 & 3$_{s=5}$ & 13.8 & 42.8 & 13.9 & 23.2 & 9.3 & 27.9 & 44.8 & 52.8 & 37.2 & 41.1 & 32.4 & 51.1 & 32.5 \\
 & 3$_{s=10}$ & 14.6 & 44.1 & 15.0 & 26.4 & 14.4 & 29.1 & 44.8 & 52.4 & 37.4 & 41.5 & 33.0 & 52.2 & 33.7 \\
 & 3$_{s=20}$ & 14.9 & 44.4 & 15.9 & 27.0 & 10.3 & 29.4 & 45.4 & 53.2 & 38.0 & 42.2 & 33.6 & 53.0 & 33.9 \\
 & 4 & 14.0 & 41.1 & 16.1 & 22.9 & 12.6 & 27.3 & 44.4 & 50.4 & 36.4 & 38.0 & 33.5 & 48.8 & 32.1 \\
 & 5$_{s=5}$ & 15.4 & 43.7 & 15.1 & 24.7 & 13.7 & 29.2 & 45.3 & 53.2 & 38.3 & 41.7 & 34.6 & 51.1 & 33.8 \\
 & \underline{5$_{s=10}$} & 16.0 & 44.6 & 16.7 & 28.1 & \textbf{17.4} & 30.4 & 45.5 & 52.7 & 38.3 & 42.1 & 35.2 & 52.3 & 34.9 \\
 & 5$_{s=20}$ & \textbf{16.6} & \textbf{45.1} & \textbf{17.4} & \textbf{28.8} & 13.9 & \textbf{30.7} & \textbf{46.0} & \textbf{53.7} & \textbf{38.8} & \textbf{42.7} & \textbf{35.6} & \textbf{53.0} & \textbf{35.2} \\
\cmidrule(l){2-15}
 & \textbf{avg} & 14.5 & 42.8 & 15.5 & 25.0 & 12.3 & 28.5 & 44.6 & 51.9 & 37.0 & 40.2 & 33.0 & 50.7 & 33.0 \\
\end{longtable}
\endgroup

\twocolumn
\begin{table*}[t]
\centering
\footnotesize
\setlength{\tabcolsep}{4pt}
\renewcommand{\arraystretch}{1.05}
\resizebox{\textwidth}{!}{
\begin{tabular}{llrrrrrrrrrrrrr}
\toprule
\textbf{Model} & \textbf{Config} & \multicolumn{6}{c}{\textbf{eng$\to$}} & \multicolumn{6}{c}{\textbf{$\to$eng}} & \\
\cmidrule(lr){3-8}\cmidrule(lr){9-14}
  &   & \textbf{\texttt{amh}} & \textbf{\texttt{hau}} & \textbf{\texttt{lug}} & \textbf{\texttt{nso}} & \textbf{\texttt{yor}} & \textbf{\texttt{zul}} & \textbf{\texttt{amh}} & \textbf{\texttt{hau}} & \textbf{\texttt{lug}} & \textbf{\texttt{nso}} & \textbf{\texttt{yor}} & \textbf{\texttt{zul}} & \textbf{Avg} \\
\midrule
\multirow{7}{*}{\rotatebox{90}{Gemini-1.5}} & 1$_{s=0}$ & 17.4 & 42.0 & 28.8 & 32.6 & 9.1 & 35.7 & 70.0 & 68.9 & 60.5 & 67.5 & 59.8 & 74.5 & 47.2 \\
 & 1$_{s=1}$ & 17.4 & \textbf{42.6} & \textbf{29.4} & 32.9 & \textbf{9.7} & 36.2 & 70.7 & 71.4 & 63.1 & 69.4 & 61.4 & 74.9 & 48.3 \\
 & 2$_{s=0}$ & 17.2 & 41.5 & 29.1 & 32.3 & 8.8 & 36.0 & 69.6 & 69.4 & 60.9 & 66.1 & 60.0 & 74.2 & 47.1 \\
 & 2$_{s=1}$ & 17.4 & 42.4 & 29.4 & 33.1 & 9.5 & \textbf{36.4} & 70.8 & 71.6 & 63.3 & 69.1 & 61.3 & 75.0 & 48.3 \\
 & 3$_{s=0}$ & 17.4 & 41.4 & 28.5 & 32.2 & 8.5 & 36.0 & 70.9 & 71.6 & 63.2 & 68.6 & 59.9 & 75.4 & 47.8 \\
 & 3$_{s=1}$ & \textbf{17.6} & 42.5 & 29.2 & \textbf{33.1} & 9.0 & 36.1 & \textbf{71.5} & \textbf{72.4} & \textbf{65.0} & \textbf{70.8} & \textbf{61.8} & \textbf{75.9} & \textbf{48.8} \\
\cmidrule(l){2-15}
 & \textbf{avg} & 17.4 & 42.1 & 29.1 & 32.7 & 9.1 & 36.1 & 70.6 & 70.9 & 62.7 & 68.6 & 60.7 & 75.0 & 47.9 \\
\midrule
\multirow{7}{*}{\rotatebox{90}{GPT-4o}} & 1$_{s=0}$ & 12.1 & 65.7 & 33.6 & 54.2 & 18.5 & 60.7 & 63.2 & 71.3 & 64.3 & 70.5 & 62.8 & 75.6 & 54.4 \\
 & 1$_{s=1}$ & 11.0 & 66.6 & 43.4 & \textbf{60.8} & 12.9 & 62.6 & 64.0 & \textbf{71.9} & 65.0 & 71.3 & 64.0 & 75.6 & 55.8 \\
 & 2$_{s=0}$ & 12.1 & 65.8 & 27.3 & 51.2 & \textbf{18.7} & 64.6 & 63.4 & 71.3 & 64.2 & 70.2 & 62.5 & 75.3 & 53.9 \\
 & 2$_{s=1}$ & \textbf{12.9} & \textbf{66.7} & 46.2 & 57.9 & 10.8 & \textbf{67.4} & 63.8 & 71.6 & 64.9 & 71.1 & 63.3 & 75.8 & \textbf{56.0} \\
 & 3$_{s=0}$ & 11.1 & 64.4 & 27.7 & 48.7 & 17.6 & 61.8 & 60.8 & 71.3 & 64.6 & 70.3 & 61.8 & 75.7 & 53.0 \\
 & 3$_{s=1}$ & 12.6 & 66.4 & \textbf{50.0} & 50.1 & 9.7 & 67.0 & \textbf{64.1} & 71.7 & \textbf{65.3} & \textbf{71.3} & \textbf{64.1} & \textbf{76.1} & 55.7 \\
\cmidrule(l){2-15}
 & \textbf{avg} & 12.0 & 66.0 & 38.0 & 53.8 & 14.7 & 64.0 & 63.2 & 71.5 & 64.7 & 70.8 & 63.1 & 75.7 & 54.8 \\
\midrule
\multirow{7}{*}{\rotatebox{90}{Gemini-3.1}} & 1$_{s=0}$ & 45.7 & 64.7 & 52.4 & 58.1 & 17.6 & 67.2 & \textbf{70.9} & \textbf{70.2} & 65.7 & 70.6 & 65.8 & 74.2 & 60.3 \\
 & 1$_{s=1}$ & 46.4 & 67.3 & 54.4 & 62.2 & 27.5 & 68.0 & 70.0 & 68.7 & \textbf{66.5} & 66.2 & \textbf{66.4} & 70.0 & 61.1 \\
 & 2$_{s=0}$ & 43.9 & 63.6 & 52.2 & 60.0 & 18.2 & 67.0 & 70.5 & 67.4 & 65.5 & \textbf{72.1} & 65.3 & 71.4 & 59.8 \\
 & 2$_{s=1}$ & \textbf{46.6} & 67.6 & \textbf{57.0} & 62.8 & 29.8 & \textbf{68.4} & 69.5 & 69.2 & 66.3 & 71.2 & 65.9 & 72.1 & 62.2 \\
 & 3$_{s=0}$ & 45.0 & 65.7 & 52.6 & 60.3 & 18.8 & 67.6 & 69.5 & 66.6 & 66.0 & 68.2 & 66.0 & 74.8 & 60.1 \\
 & 3$_{s=1}$ & 46.5 & \textbf{68.0} & 55.4 & \textbf{63.1} & \textbf{35.3} & 68.3 & 70.3 & 67.3 & 66.5 & 71.4 & 66.1 & \textbf{75.2} & \textbf{62.8} \\
\cmidrule(l){2-15}
 & \textbf{avg} & 45.7 & 66.2 & 54.0 & 61.1 & 24.5 & 67.7 & 70.1 & 68.2 & 66.1 & 70.0 & 65.9 & 73.0 & 61.0 \\
\midrule
\multirow{7}{*}{\rotatebox{90}{GPT-5.4}} & 1$_{s=0}$ & 42.9 & 65.8 & 58.9 & 63.8 & 19.5 & 68.3 & 72.7 & 73.9 & 68.3 & 74.0 & 67.3 & 77.1 & 62.7 \\
 & 1$_{s=1}$ & 44.0 & 67.0 & \textbf{60.3} & 65.6 & 32.4 & 69.2 & 73.4 & 75.1 & 69.1 & 75.4 & 68.4 & 78.2 & 64.8 \\
 & 2$_{s=0}$ & 42.6 & 65.8 & 59.5 & 63.9 & 19.8 & 68.6 & 72.6 & 74.2 & 67.6 & 74.0 & 67.4 & 77.2 & 62.8 \\
 & 2$_{s=1}$ & 43.6 & 67.3 & 59.8 & 65.0 & 35.1 & 69.5 & 73.4 & 75.1 & 69.0 & 75.2 & 68.3 & 78.4 & 65.0 \\
 & 3$_{s=0}$ & 42.8 & 66.5 & 59.4 & 63.7 & 19.8 & 69.1 & 73.1 & 74.6 & 68.8 & 74.8 & 68.2 & 78.1 & 63.2 \\
 & 3$_{s=1}$ & \textbf{44.3} & \textbf{68.0} & 60.2 & \textbf{65.8} & \textbf{37.0} & \textbf{69.5} & \textbf{74.1} & \textbf{75.9} & \textbf{69.5} & \textbf{76.1} & \textbf{69.0} & \textbf{78.5} & \textbf{65.7} \\
\cmidrule(l){2-15}
 & \textbf{avg} & 43.3 & 66.7 & 59.7 & 64.6 & 27.3 & 69.0 & 73.2 & 74.8 & 68.7 & 74.9 & 68.1 & 77.9 & 64.0 \\
\bottomrule
\end{tabular}
}
\caption{Document-level template/shot ablation for the four closed models (older: GPT-4o, Gemini-1.5-Flash; newer: GPT-5.4, Gemini-3.1-Flash-Lite) across the 3-template $\times$ 2-shot grid (chrF). Each row is a (template, shot-count) configuration for that model (template number first, shot count as subscript $s$). Open LLMs use a single doc-level configuration each (no template variants); see \Cref{tab:doc_chrf} for the open-vs-closed doc-level comparison. \textbf{Bold}: per-language column maximum within each model block. The bottom \textbf{avg} row of each block reports the per-language average across configurations.}
\label{tab:doc_template_shot_ablation_chrf}
\end{table*}

\end{document}